\documentclass[Afour,sageh,times]{sagej}
\usepackage{moreverb}

\newcommand\BibTeX{{\rmfamily B\kern-.05em \textsc{i\kern-.025em b}\kern-.08em
T\kern-.1667em\lower.7ex\hbox{E}\kern-.125emX}}

\usepackage{graphics} 
\usepackage{epsfig} 
\usepackage{times} 
\usepackage{amsmath} 
\usepackage{amssymb}  
\usepackage{bm}
\usepackage{multirow}
\usepackage{float}
\usepackage{subfig}
\usepackage{url}
\usepackage{enumerate,mdwlist,asciilist}
\usepackage[breakable]{tcolorbox}
\usepackage{hyperref}
\hypersetup{hypertex=true,
urlcolor=blue,
colorlinks=true,
linkcolor=blue,
anchorcolor=blue,
citecolor=blue}
\usepackage{flushend}
\usepackage{booktabs}
\usepackage{bbding}
\usepackage{xcolor} 
\usepackage{ifthen}

\usepackage{soul} 

\definecolor{Second}{rgb}{0.8745, 0.6896, 0.6824}
\definecolor{First}{rgb}{0.5451, 0.7216, 0.7412}
\newcommand{\bestresult}[1]{\textcolor{First}{\textbf{#1}}}
\newcommand{\secondresult}[1]{\textcolor{Second}{\underline{#1}}}
\newcommand{\boldcolor}[2]{\ifthenelse{\equal{#1}{First}}{\bestresult{#2}}{\secondresult{#2}}}
\newcommand{\realgsr}[4]{#1 (#2) {\scriptsize[#3, #4]}}
\newcommand{\realdr}[4]{#1 (#2/75) {\scriptsize[#3, #4]}}
\usepackage{newtxmath}   

\usepackage{tabularx}

\usepackage{tikz}     

\AsciiListSetAutochars{-,*,+,=}

\newcommand{\pinktriangle}[1][1]{%
    \begin{tikzpicture}[baseline=0.1ex, scale=#1]
        \filldraw[fill=pink, draw=black] (0,0) -- (0.4,0) -- (0.2,0.35) -- cycle;
    \end{tikzpicture}%
}

\def\volumeyear{2026}

\begin{document}

\runninghead{Song et al.}

\title{Equivariant Volumetric Grasping}

\author{Pinhao Song$^{1,3}$, Yutong Hu$^{1,3}$, Pengteng Li$^{4}$, Renaud Detry$^{1,2,3}$}

\affiliation{$^1$KU Leuven, Dept. Mechanical Engineering, Research unit Robotics, Automation and Mechatronics. \\
$^2$KU Leuven, Dept. Electrical Engineering, Research unit Processing Speech and Images.\\
$^3$Flanders Make@KU Leuven.\\
$^4$AI Thrust, HKUST(GZ)
}

\corrauth{{Pinhao Song, \textit{pinhao.song@kuleuven.be}}}


\begin{abstract}
We propose a new volumetric grasp model that is equivariant to $90^\circ$ rotations around the vertical axis, leading to a significant improvement in sample efficiency. Our model employs a tri-plane volumetric feature representation---i.e., the projection of 3D features onto three canonical planes. We introduce a novel tri-plane feature design in which features on the horizontal plane are \emph{equivariant} to $90^\circ$ rotations, while the paired features from the other two planes transform under the corresponding induced action. Building on this representation, we further develop equivariant adaptations of two state-of-the-art volumetric grasp planners, GIGA and IGD, to demonstrate its applicability across discriminative and generative grasp-planning pipelines.
We provide a detailed analytical justification of the proposed equivariance properties and validate our approach through extensive simulated and real-world experiments. Our results demonstrate that the proposed projection-based design reduces both computational and memory costs. Moreover, the equivariant grasp models built on top of our tri-plane features outperform their non-equivariant counterparts, achieving higher performance within a real-time cost constraint. Video and code are available at: \url{https://mousecpn.github.io/evg-page/}

\end{abstract}


\maketitle

\section{Introduction} \label{sec: introduction}
Grasping in cluttered environments remains a grand challenge in robotics. Determining the optimal gripper pose for densely packed, partially occluded objects is complex, particularly when items vary in shape, material, and orientation. Unlike predictable, well-structured scenes, cluttered settings necessitate perceptual systems finely attuned to visual detail. Effective grasp planning must therefore integrate multiple sensory cues, such as shape, appearance, and shading, a capability crucial for progress in industrial automation, service robotics, and healthcare.

\begin{figure}[!t]
  \centering
  \includegraphics[width=\linewidth]{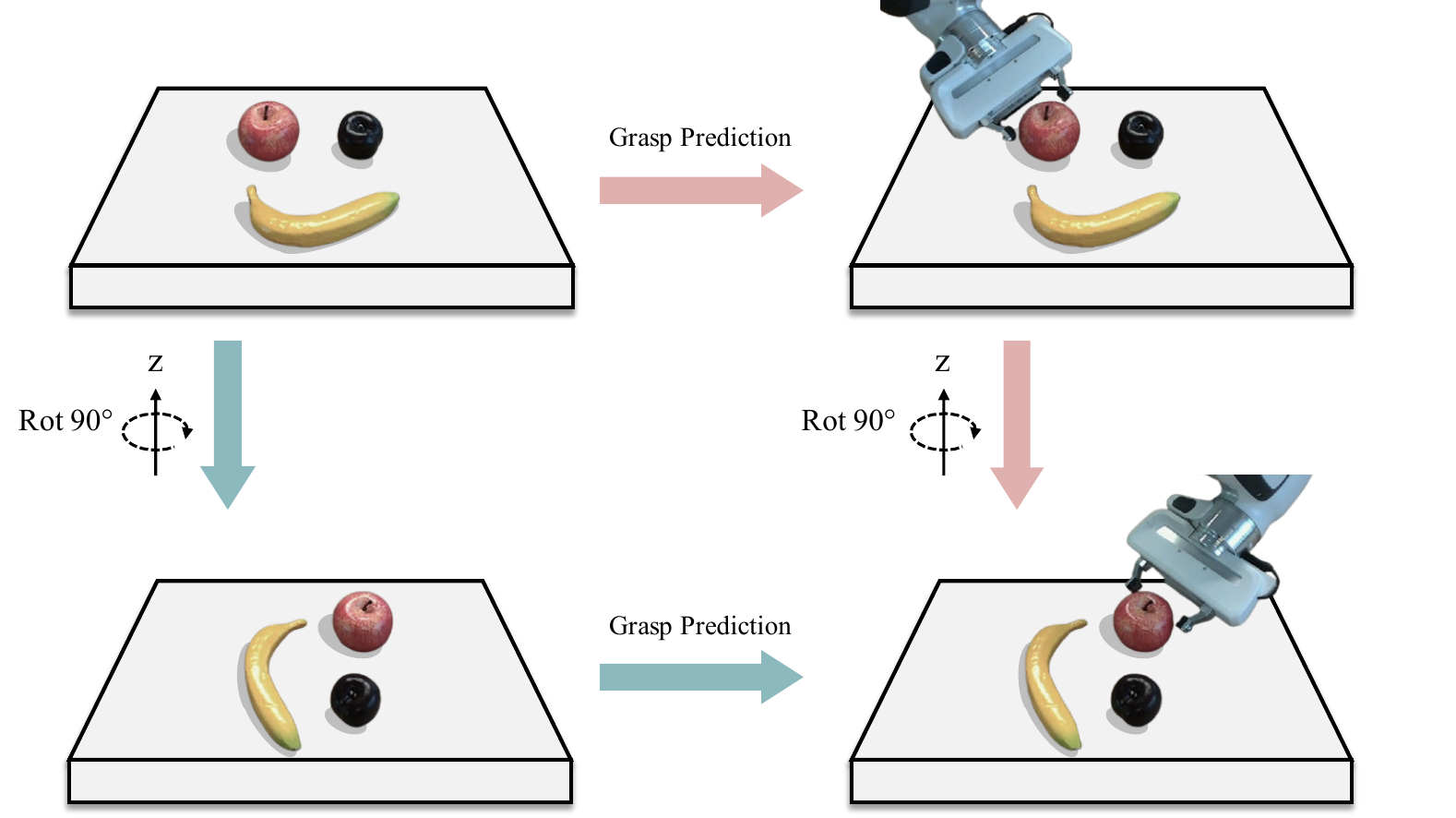}
  \caption{An illustration of rotational equivariance in robotic grasp prediction. When the scene (\textcolor{First}{left}) is rotated by $90^\circ$ (\textcolor{First}{bottom}), the predicted grasps (\textcolor{Second}{right}) also rotate by the same amount, ensuring consistent grasp orientations for the transformed scene.}
  \label{fig: equi}
\end{figure}

Recent advances in volumetric grasping approaches~\citep{vgn, igd, giga, liu2024efficient, neugraspnet} have demonstrated significant promise in handling the complexity of cluttered scenes. These methods first employ one convolutional layer to lift a volumetric scene representation into \emph{volumetric primitive features}; subsequently, these features are refined into \emph{volumetric grasp features} via a stack of CNNs; and finally, feasible grasps are predicted at the voxel level using grasp features.
The local receptive fields of CNNs are essential for capturing the fine geometric details critical to grasp synthesis, while their inherent parallelism facilitates real-time inference~\citep{vgn}. By applying a single kernel across all spatial locations in the feature maps, CNNs are inherently translation equivariant, enabling generalization across varying object translations. Unfortunately, CNNs' \emph{architectural} equivariance stops with translation. No mechanism in CNNs endows them with architectural rotation equivariance, and rotation generalization only comes from exposure to large datasets that sample the rotation space exhaustively---i.e., \emph{data-driven} equivariance.


Instead of attempting to address orientation generalizability through data, \emph{designing the model} for \emph{architectural} rotation equivariance has the potential to improve sample efficiency. An equivariant model ensures that its outputs transform predictably and coherently with input transformations (e.g., rotation, translation, reflection). In robotic grasping, this property allows a policy to generalize across different object orientations without requiring additional training data. As illustrated in Fig.~\ref{fig: equi}, a rotationally equivariant policy can apply learned strategies regardless of an object's pose, thereby improving both grasp success and stability when encountering novel orientations. However, while recent $\textnormal{SO}(3)$-equivariant models have shown strong performance~\citep{hu2024orbitgrasp, diffusion-edfs, edf, freiberg2025diffusion}, they come at a steep computational cost: for instance, \textit{OrbitGrasp} infers only 17 grasps per second, and \textit{Multi-embodiment Grasping} achieves a mere 0.6 grasps per second. In stark contrast, non-equivariant models like \textit{VGN}~\citep{vgn} can predict over $100{,}000$ grasps per second. The computational and memory burdens associated with current high-fidelity equivariant methods restrict their practical applicability to real-world, time-critical applications.

In this paper, we propose a new \textit{volumetric grasping model} that achieves strong sample efficiency through rotational equivariance. We specifically focus on means of achieving equivariance that strike a more favorable trade-off between computational cost and sample efficiency than previous works. Specifically, instead of imposing computationally expensive $\textnormal{SO}(3)$ equivariance, we strategically focus on equivariance to vertical-axis rotations. This design choice is particularly effective in tabletop scenarios. While objects may physically rotate around any axis, gravity commonly restricts their posture to standing or lying down, making the vertical axis the overwhelmingly dominant axis of rotation. Thus, equivariance to rotations around the vertical axis is poised to have a significant impact on sample efficiency in tabletop scenarios. By contrast, generalizing the prediction of feasible grasping poses to less frequent rotations inefficiently uses representational capacity. Instead, any necessary generalizations to these less frequent poses are more effectively learned implicitly from data.



Our main contribution focuses on preserving equivariance between volumetric primitive features and volumetric grasp features (per the definitions provided above). Because of the computational cost of 3D steerable convolution, we opt \emph{not} to represent them with a stack of 3D steerable filters. Instead, we represent volumetric grasp features via a tri-plane representation \citep{giga,peng2020convolutional}, which projects volumetric features onto the three canonical planes, allowing us to process features within three 2D planes instead of a 3D volume. The key observation of our work is that a tri-plane feature representation does preserve rotation equivariance, albeit not as trivially as 3D steerable convolutions. Instead, we observe that 90$^\circ$ rotations of the workspace around the vertical axis have a distinct, predictable effect on the tri-plane representation, which consists of a combination of plane rotations, reflections, and permutations (see Sec.~\ref{sec: encoding}).
Our work capitalizes on this observation by designing the convolutional layers that enrich the features of the three canonical planes in a way that \emph{preserves this form of rotational equivariance}. The cost reduction from the planar computations owed to the tri-plane representation, combined with the sample efficiency endowed by rotational equivariance, yields expressive features at a fraction of the computational cost of 3D equivariant CNNs.

To demonstrate that the proposed representation can be used beyond a single planner architecture, we instantiate it in two state-of-the-art volumetric grasp planners, \textit{GIGA} \citep{giga} and \textit{IGD} \citep{igd}, yielding EquiGIGA and EquiIGD. These instantiations preserve the representation's equivariance through grasp orientation regression in GIGA and grasp sampling in IGD. For EquiIGD, we further derive an equivariant formulation of deformable attention and an equivariant flow-matching model for grasp orientations. We provide a formal analysis of the proposed equivariance properties and validate our design through comprehensive simulated and real-world experiments. Our results show that the projection-based design reduces memory and computational costs. Moreover, models built on our tri-plane features generally outperform non-equivariant baselines, achieving superior performance at a cost within a real-time constraint.


In summary, we propose a \textbf{volumetric grasp model} that is equivariant to 90$^\circ$ vertical-axis rotations: \begin{itemize}
\item We introduce a novel \textbf{tri-plane representation} that maintains equivariance, at a computation cost far smaller than that of 3D steerable CNNs.
\item We demonstrate the representation's generality by integrating it into GIGA and IGD, yielding \textbf{EquiGIGA and EquiIGD} that preserve equivariance in grasp orientation regression (GIGA) or sampling (IGD).
\item We introduce vertical-to-tableplane feature augmentation (V2TP), a secondary component that transfers complementary vertical-plane information to the table-plane branch while preserving the induced $C_4$ action.
\item Our work demonstrates improved sample efficiency, as well as improved performance over baseline GIGA and IGD, both in simulation and the real world, at a tractable memory and runtime cost.
\end{itemize} 

\section{Related Works}
\subsection{6-DoF Grasp Planning}
Recent work on 6-DoF grasping can be grouped into two families: volumetric grasping \citep{vgn,igd,giga,liu2024efficient,neugraspnet,ko2025simultaneous} and point-cloud grasping \citep{wang2021graspness, fang2020graspnet, huang2023edge, hu2024orbitgrasp}. Volumetric grasping uses a volumetric scene representation as input, e.g., a truncated signed distance function (\emph{TSDF}), and it predicts grasps for each voxel of this representation \citep{vgn, giga}. Owing to their highly structured and dense representation, these methods directly inherit the fast inference of CNNs (e.g., 9 ms for a forward pass of VGN \citep{vgn}). In contrast, point-cloud grasping methods operate directly on point clouds and infer grasps anchored at points \citep{fang2020graspnet,wang2021graspness,s4g}, surface patches \citep{hu2024orbitgrasp, huang2023edge}, or anchored in a representation of the entire scene \citep{se3dif, 6dofgraspnet, freiberg2025diffusion}. These methods employ point cloud processing architectures like PointNet \citep{qi2017pointnet} or Vector Neuron \citep{vn} to extract features. They commonly utilize farthest point sampling (FPS) for efficient downsampling and K-nearest neighbors (KNN) for aggregating local features, serving a similar purpose to convolutional operations in grid-based data. Point-cloud methods can preserve fine geometric details by directly handling sparse data. However, these methods suffer from slower inference times due to the more complex processing of unstructured point clouds.

There is a long-standing debate between researchers supporting volumetric representation and those advocating for point cloud representation. On one hand, some studies \citep{hu2024orbitgrasp} argue that volumetric representations suffer from resolution limitations, hindering their ability to capture the fine-grained structure of objects. Moreover, the computational cost and memory footprint of voxel-based models increase cubically with the input resolution, rendering them impractical for memory-constrained scaling. On the other hand, point clouds can offer higher resolution but are inherently less structured, making them more challenging to process. Additionally, as Wu et al. \citep{wu2024point} note, operations such as KNN and FPS introduce significant extra computational overhead. To avoid the use of KNN and FPS, Minkowski Convolutional Network \citep{minkunet} can extract features efficiently through voxelization and sparse convolution, which is adopted by GSNet \citep{wang2021graspness}, ICGNet \citep{zurbrugg2024icgnet}, and DexGraspNet2 \citep{zhang2024dexgraspnet}. Similarly, PVCNN \citep{liu2019point} also employs voxelization to structure point features, thereby reducing memory consumption and enhancing computational efficiency. In summary, the substantial efficiency afforded by voxelization could outweigh the resulting loss of high resolution.

Given this crucial trade-off, this work leverages a variant of volumetric representation---the tri-plane representation \citep{peng2020convolutional, giga, igd}, which projects 3D features onto three orthogonal planes. This approach enables the processing of 3D features via 2D convolution, reducing the cubic memory cost to a quadratic one. In addition, we introduce a {vertical-to-tableplane} feature augmentation module, which facilitates communication between {vertical features} and tableplane features, thereby enhancing the model's overall 3D understanding with a tractable memory and runtime cost.

\subsection{Equivariant Deep Learning in Robot Manipulation}
If tools or objects are translated or rotated, many of their properties, such as the grasp poses that they offer, generally transform accordingly. A model designed to encode this equivariance is highly desirable. Equivariant deep learning improves sample efficiency and generalization in various robot learning tasks \citep{hu2024orbitgrasp, edf, huang2023edge}. There are three main equivariant methodologies leveraged in robot learning. First, steerable CNNs \citep{cohen2016steerable} incorporate symmetries directly into the structure of convolutional layers, enabling them to generalize automatically across a symmetry group. They achieve equivariance by parameterizing kernels in terms of a band-limited basis (e.g., circular or spherical harmonics), ensuring that the learned features transform predictably when input data are transformed. Previous works \citep{wang2022equivariant,zhu2023robot} have introduced steerable CNNs for learning $\textnormal{SE}(2)$ grasps. Steerable CNNs can operate over both finite and continuous transformation groups; however, some studies \citep{cesa2022program} have observed that models defined on continuous groups underperform those defined on finite groups, and the latter tend to have lower computational cost. The second equivariant methodology used in robot learning is the Vector Neuron \citep{vn}, designed to achieve exact $\textnormal{SO}(3)$ equivariance by extending neurons from 1D scalars to steerable 3D vectors. Several works \citep{huang2023edge, lim2024equigraspflow} have employed vector-neuron-based encoders to attain equivariance. EquiGraspFlow \citep{lim2024equigraspflow} is an $SE(3)$-equivariant, object-level flow-based generator trained on complete point clouds from four objects; it is therefore methodologically related but not designed for the multi-object decluttering setting considered here. Nevertheless, the performance of Vector Neuron is constrained by its limited degree of equivariant representation: only vector features are communicated in the network. In addition, the non-linearity in Vector Neurons introduces new parameters and therefore computational overhead. Equivariant transformers \citep{fuchs2020se, liao2022equiformer, liao2023equiformerv2} are the third equivariant methodology in robotics. Several studies leveraging equivariant transformers \citep{hu2024orbitgrasp, edf, diffusion-edfs} have achieved promising performance in robotic manipulation. However, these models incur a high computational and memory cost. For example, EDF \citep{edf} and OrbitGrasp \citep{hu2024orbitgrasp} can only be trained with a batch size of one, and OrbitGrasp \citep{hu2024orbitgrasp} infers only 17 grasps per second. Similarly, Multi-embodiment Grasping \citep{freiberg2025diffusion} requires 35 seconds to generate 20 grasps. This cost hinders their practical application, particularly when deploying on edge devices.

In this work, we propose an equivariant feature encoder for volumetric grasping models. Our model is defined on the cyclic group $\textnormal{C}_4 \subset \textnormal{SO}(2)$ because it efficiently captures most of the varieties of objects' transformations in tabletop scenarios. The proposed approach first projects volumetric features onto three orthogonal planes ($XY$, $XZ$, and $YZ$).
By exploiting the induced transformation structure of $\textnormal{C}_4$, we model the $XY$-plane features as $\textnormal{C}_4$ regular representations, while processing the paired $XZ$- and $YZ$-plane features as $\textnormal{C}_2$ reflection-equivariant fields. Under a vertical-axis rotation, the latter jointly follow an induced $\textnormal{C}_4$ action consisting of plane permutation and reflection.
By projecting 3D features into 2D and employing a finite symmetry group, our model achieves efficient training and inference.

\section{Preliminaries} \label{sec: preliminaries}
This paper leverages the concept of symmetry---specifically, \emph{equivariance}---to improve representation learning for robotic grasping. This section introduces elements of group theory and representation theory upon which our methodology builds. A function is said to be equivariant if transformations of its input (e.g., a translation, rotation, reflection, etc.) have a \emph{predictable effect} on its output. The most intuitive example is that rotating a model’s input causes its output to rotate by the same angle. An example of a model that exhibits this type of equivariance is a trivial segmentation model where the input is an image and the output is a pixel-wise mask with a value based purely on thresholding the color of each corresponding input pixel. For such a model, $S(R_a (I)) = R_a(S(I))$, where $S$ is the segmentation model, $I$ is any input image, and $R_a$ is a rotation by an angle $a$ that is a multiple of $90^\circ$. The model is equivariant to $90^\circ$ rotations, and the predictable effect of a rotated input is a rotated output.

Predictable effects, such as the one above, where the output transformation is identical to the input transformation, are common, but other types of relationships exist. Consider a model that tracks motion between two video frames to create an optical flow map (a map of arrows showing where objects are moving). If you rotate the input video frames, two things happen to the output: (a) The positions of the arrows move to new locations; (b) The direction of the arrows must also rotate to stay accurate to the new orientation. In this case, the \emph{predictable effect} is more complex because the rotation affects both where the data is and what the data says. The formulation of the predictable effect requires us to formalize two key concepts: (1) the \emph{predictability} of the functions, and (2) the \emph{effect} of applying a transformation to the data. To formalize the first concept, we write a mathematical definition of equivariant functions, which communicates the \emph{effect} of the group from input to output (Sec.\ \ref{sec: equi funcs}). To formalize the second concept, we explain how a group's \emph{representations} transform features in a way that respects symmetry (Sec.\ \ref{sec: group act}). Finally, we will also introduce some common building blocks of equivariant neural networks (Sec.\ \ref{sec: enn}).

Today's grasp models are already \emph{translation}-equivariant thanks to their use of convolutional layers. In our work, we primarily focus on models that, in addition to translation equivariance, provide equivariance to $90^\circ$ rotations around the vertical axis. Thus, we will introduce the cyclic group of $\textnormal{SO}(2)$ (Sec.\ \ref{sec:so2}), which represents the planar rotations upon which our methodology (Sec.\ \ref{sec: encoding}) builds.

Readers who are already familiar with equivariant deep learning theory may skip this section. A comprehensive theoretical overview can be found in the book of Weiler et\ al.\ \citep{weiler2024equivariant}.

\begin{figure*}[!t]
  \centering
  \includegraphics[width=\linewidth]{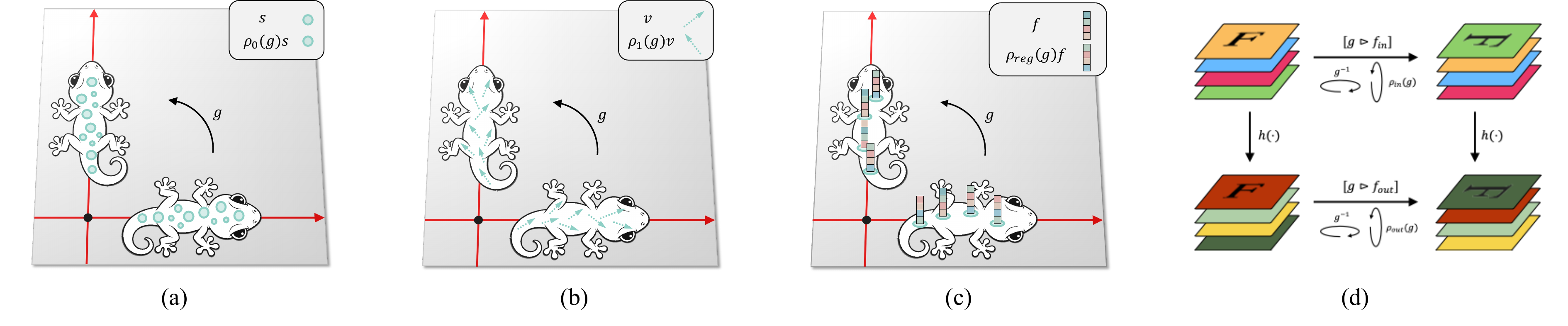}
  \caption{(a) Trivial representation and its transformation. (b) Irreducible representation and its transformation. (c) Regular representation and its transformation under $\textnormal{C}_4$. For (a-c), see text for details. (d) The diagram illustrates that $h$, an equivariant function, commutes with the group action. Applying a group transformation $g$ to the input feature $f_{\text{in}}$ results in transformed output $f_{\text{out}}$, where the transformation is carried through $h$ via consistent actions defined by $\rho_{\text{in}}$ and $\rho_{\text{out}}$. Thus, applying $g$ before $h$ (top-right path) is equivalent to applying $g$ after (bottom-left path).}
  \label{fig: background}
\end{figure*}

\subsection{The Cyclic Group of $\textnormal{SO}(2)$} \label{sec:so2}
In this paper, we primarily focus on encoding equivariance to $90^\circ$ rotations around the vertical axis. To mathematically formalize $90^\circ$ rotations, it is natural to first consider generic planar rotations in 2D space. A planar rotation can be represented by an element in the special orthogonal group $\textnormal{SO}(2)$, a group of all $2\times2$ real orthogonal matrices with determinant $1$, which represents all possible rotations in a 2D plane:
\begin{equation}
    \textnormal{SO}(2) = \left\{\left.R(\theta)=\begin{bmatrix}
        \cos \theta & -\sin \theta  \\
        \sin \theta & \cos \theta
        \end{bmatrix} \right| \theta \in \mathbb{R} \right\}.
\end{equation}
In the context of $\textnormal{SO}(2)$, a cyclic subgroup is a finite subgroup of $\textnormal{SO}(2)$ that consists of rotations by fixed angles that are integer multiples of a fundamental rotation. Mathematically, the cyclic group $\textnormal{C}_n$ of $\textnormal{SO}(2)$ is defined as:
\begin{equation}
    \textnormal{C}_n = \left\{ \left. R \left(\frac{2\pi k}{n}\right) \right| k\in\{0,\ldots,n-1\} \right\}.
\end{equation}
Thus, a $90^\circ$ rotation can correspond to an element in the cyclic group $\textnormal{C}_4 = \{e, r, r^2, r^3\}$, where $e$ is the identity element ($0^\circ$ rotation), and $r^k$ represents a rotation by $90^\circ \times k$. In this paper, we primarily use the $\textnormal{C}_4$ group due to its computational efficiency, acceptable level of $\textnormal{SO}(2)$ signal reconstruction, and the distinct predictable effect of $90^\circ$ rotations on a tri-plane representation that we introduced in Sec.\ \ref{sec: introduction} and discuss in detail in Sec.\ \ref{sec: encoding}.

\subsection{Equivariant Functions} \label{sec: equi funcs}
We are interested in designing an equivariant model, where transforming its input has a predictable effect on its output. To build such a model, we need to mathematically formalize \emph{equivariant functions}. Specifically, given a feature map $f$, a mapping $h$ is said to be equivariant with respect to $G$ if it commutes with the group action, as:
\begin{equation}
    h[g \triangleright_{i} f] = g \triangleright_{o} h[f], \quad g \in G, \label{eq: equi func}
\end{equation}
where $g \triangleright_{i} f$ and $g \triangleright_{o} h[f]$ denote the group action on the input feature $f$ and output feature $h[f]$, respectively. They describe the effect of applying the group element $g$ to $f$ and $h[f]$. The effect of the group action on $f$ and $h[f]$ can be different because the input and output features may have different structures. An important special case of equivariant functions is \emph{invariant functions}, where the output remains unchanged under group actions, as:
\begin{equation}
    h[g \triangleright_{i} f] = h[f], \label{eq: invariant func}
\end{equation}
which can be seen as $g \triangleright_{o}$ being the identity action for all $g \in G$ in Eq.~\ref{eq: equi func}.

\subsection{Group Action on Feature Fields} \label{sec: group act}
Equivariant functions can communicate the effects of a group action from input to output. To further understand equivariant functions, we need to clarify the \emph{effect} of applying a group element to feature fields. We aim to encode equivariance for 2D or 3D feature maps. To align with the convention in group theory, we view feature maps as \emph{feature vector fields} $f: \mathbb{R}^2 \rightarrow \mathbb{R}^C$, which assign a $C$-dimensional feature vector $ f(\mathbf{p}) \in \mathbb{R}^C$ to each position $\mathbf{p} \in \mathbb{R}^2$. There are two kinds of effects when applying a group element to the feature vector field: (i) transforming the coordinates of the feature vectors and (ii) transforming the feature vectors themselves. 

The most intuitive effect is directly rotating the coordinates of the feature vectors. Let us take Fig.~\ref{fig: background}~(a) as an example. Applying a group element $g \in G$ to a lizard results in a rotation of its position. Mathematically, the group element $g$ acts on the feature field $f$ by transforming the input coordinates as follows:
\begin{equation}
    [g \triangleright f](\mathbf{p}) = f(g^{-1} \cdot \mathbf{p}), \label{eq: trivial repr}
\end{equation}
where $g \triangleright f$ denotes group action on the feature field $f$. This means that the feature vector at position $\mathbf{p}$ in the transformed field $g \triangleright f$ is equal to the feature vector at the position $g^{-1} \cdot \mathbf{p}$ in the original field $f$. 

The second kind of effect occurs when the feature vectors themselves change. Consider a feature field where each position contains a 2D vector indicating a direction, such as an optical flow field. When a group element $g$ acts on this feature field, the positions of the vectors move to new locations, and the direction of the vectors must also rotate to stay accurate to the new orientation. As shown in Fig.~\ref{fig: background}~(b), the vectors' origins move with the lizard, and the vectors' orientations also rotate by $g$. This can be expressed mathematically as:
\begin{equation}
    [g \triangleright f](\mathbf{p}) = \rho_1(g) f(g^{-1} \cdot \mathbf{p}).
\end{equation}
Compared to Eq.~\ref{eq: trivial repr}, where only the coordinate space changed, here the feature vectors themselves also change according to the rotation applied. For $g=R(\theta)$, this case uses the type-1 representation $\rho_1(R(\theta))=R(\theta)$. More generally, a type-$\omega$ feature transforms under the irreducible representation $\rho_\omega(R(\theta))=R(\omega\theta)$, where $\omega$ is the angular frequency. The feature that transforms under this representation is called the type-$\omega$ feature. Specifically, $\rho_0(g) = 1$ is \emph{the trivial representation}. Features that transform this way are called \textit{invariant features}, since they do not respond to rotations at all, which corresponds to the previous case we discussed. It should be noted that \textit{invariant features} do not necessarily come from invariant functions. We can compare Eq.~\ref{eq: trivial repr} and Eq.~\ref{eq: invariant func}. Eq.~\ref{eq: trivial repr} describes a feature field consisting of invariant features, but the feature coordinates still change under group actions due to coordinate transformations. In contrast, in Eq.~\ref{eq: invariant func}, even the feature coordinates remain unchanged under group actions. Irreducible representations $\rho_\omega(g)$ and trivial representation $\rho_0(g)$ are \emph{group representations}, which describe how feature vectors transform by group actions.

Besides the two aforementioned group representations, there is also a widely used group representation called \emph{regular representation} (denoted as $\rho_{\text{reg}}$). This representation applies to a vector $x \in \mathbb{R}^n$, where $n=|G|$ is the number of elements in the group $G$. For the group $\textnormal{C}_4 = \{e, r, r^2, r^3\}$, each group element $g = r^m$ corresponds to a cyclic permutation of the entries of $x$. For example, a $90^\circ$ rotation applied to $x = (x_1, x_2, x_3, x_4)$ would yield $\rho_{\text{reg}}(r)x = (x_4, x_1, x_2, x_3)$, shifting all elements one position to the right. As shown in Fig.~\ref{fig: background}~(c), the feature positions move with the lizard, and the feature channels also shift. Features that follow this pattern are called \textit{regular features}.

To summarize, we have introduced three types of feature transformations under group actions. Then we can write a general formula for group action on feature fields as:
\begin{equation}
    [g \triangleright_\rho f](\mathbf{p}) = \underbrace{\rho(g)}_{\text{Feature transform}} f(\underbrace{g^{-1} \cdot \mathbf{p}}_{\text{Coordinate transform}}),
\end{equation}
where $\rho(g)$ can be any group representation, such as trivial, irreducible, or regular representations.

Equivariance often takes a form that cannot be expressed with a single feature representation, but needs a combination of representations instead. For instance, a grasp model may predict grasp quality alongside gripper pose, two variables whose predictability needs to be modeled with different group representations. Feature combination is often achieved via concatenation $f = \bigoplus_{k=1}^K f^k_{\phi(k)}$, where $f^k_{\phi(k)}$ is the $k$-th feature component, $\phi(k) \in \{0, 1, \text{reg}\}$ denotes its feature type, and $f$ is the concatenated feature. The group representation of these concatenated features can be built as a block-diagonal concatenation of the group representation of all components (\emph{direct sum}) $\rho(g) = \bigoplus_{k=1}^K \rho_{\phi(k)}(g)$, where $\rho_{\phi(k)}(g)$ is the representation acting on feature type $\phi(k)$. Under the rotation of $g$, the combined features $f$ will transform in a way that their components transform independently, as $\rho(g)f = \bigoplus_{k=1}^K \rho_{\phi(k)}(g) f^k_{\phi(k)}$. This structure allows us to control how features transform under symmetry operations. By determining the feature types, we can ensure the model’s outputs have the desired geometric properties---for example, remaining invariant to rotation or changing predictably under it.

\subsection{Equivariant Neural Networks} \label{sec: enn}
Finally, we assemble the concepts of equivariant functions and group actions to construct equivariant neural networks. An equivariant neural network is built by stacking multiple equivariant functions with learnable parameters. Since each function is equivariant, the entire network is also equivariant, ensuring that the predictable effect of transformations is preserved throughout the model. A common parametrized equivariant function for equivariant neural networks is \emph{the equivariant linear layer}, which is analogous to conventional linear layers in neural networks. Besides, analogous to how convolutional layers generalize linear layers for spatially structured data, \emph{steerable convolutional layers} (steerable CNNs) extend equivariant linear layers to 2D or 3D domains. A steerable convolutional layer can not only achieve equivariance to translations (like a standard convolutional layer) but also to other transformations such as rotations, reflections, and more complex symmetries, depending on the group representations used in the layer. Steerable convolutional layers share the same weight parameters across translation and rotation transformations, which improves sample efficiency. The model does not need to see an object in every possible orientation during training to recognize it (or grasp it) in a new orientation during testing. The generalization is mathematically guaranteed by the group structure, achieving \emph{architectural} equivariance.

\section{Overview}
Our work focuses on improving the sample efficiency of grasp planning by incorporating equivariance constraints into the model. Setting aside the robot's kinematic constraints, grasp planning tasks inherently exhibit equivariance to translations and rotations: If a scene is transformed by an element of $ \textnormal{SE}(3)$, the predicted grasps transform accordingly. Grasp \emph{models} typically preserve translation equivariance \citep{vgn}, but they lack \emph{rotation} equivariance, which hurts their sample efficiency. Our work focuses on the preservation of a form of rotation equivariance.

As discussed above, the preservation of equivariance in a DNN comes at an increase of the compute cost of the filters that form the network. From a theoretical perspective, preserving $\textnormal{SO}(3)$ equivariance would be ideal, but previous attempts have shown that the additional compute cost that it incurs is excessive compared to gains in sample efficiency.
Here, we focus instead on equivariance to rotations around the $Z$ (vertical) axis. The choice of limiting our model to equivariance to rotations around $Z$ is motivated by the following observations in tabletop grasping tasks: (a) While objects can physically rotate arbitrarily, gravity predominantly restricts their stable postures (standing or lying) such that rotations around the vertical axis become overwhelmingly dominant.
(b) Feasible, collision-free grasps in tabletop environments are primarily constrained to the upper hemisphere. The restricted search space offers significantly less geometric variety and fewer opportunities for generalization across non-vertical axes compared to the vertical axis.

Consequently, preserving equivariance to rotations around the vertical axis yields a greater relative improvement in sample efficiency than equivariance around other axes, where physical constraints already limit the reachable configuration space. It allows us to effectively generalize within the most relevant degrees of freedom of the task, while mitigating computational costs associated with the enforcement of symmetries.

The vertical-axis rotations we aim to preserve are the $90^\circ$ rotations that form the $\textnormal{C}_4$ subgroup of $\textnormal{SO}(2)$. This choice is motivated by the design of the volumetric grasp planners we build upon \citep{vgn,giga}. These planners typically consist of two parts: a feature encoder and a set of prediction heads. The feature encoder transforms a volumetric input (3D grid or feature field) into an implicit representation that can be queried at arbitrary 3D coordinates. In turn, the prediction heads predict the rotation, graspness, and occupancy given a certain position. Equivariance must be preserved in both the feature encoder (Sec.~\ref{sec: encoding}) and the prediction heads (Sec.~\ref{sec: decoding}).

Our feature encoder builds upon the tri-plane UNet (Tri-UNet) previously used in GIGA \citep{giga} and ConvONets \citep{peng2020convolutional}. The tri-plane UNet reduces computation and memory consumption from cubic to quadratic compared to volumetric representations by replacing 3D CNNs with 2D ones. We propose a new form of tri-plane UNet that preserves equivariance to vertical-axis rotations in addition to translations. The orthogonality of the three planes is the part of the model that is conducive to preserving equivariance to rotations by $90^\circ$: the effect of these rotations on the two vertical planes consists of a simple permutation and reflection of the two planes. This observation makes $\textnormal{C}_4$ equivariance far simpler to preserve than generic $\textnormal{SO}(2)$ equivariance. This point is further discussed in the next section.

We build the prediction heads with equivariant linear layers, which map the queried features to grasp rotation, graspness, and occupancy. These variables transform equivariantly in different ways: graspness and occupancy are invariant to the input transformation, and rotation transforms consistently with the input transformation. Our proposed equivariant volumetric grasping model achieves these consistencies by identifying the feature types of the output variables, and improves sample efficiency and robustness thanks to their intrinsic equivariance constraints.

\section{Rotation-equivariant Tri-plane UNet for Volumetric Feature Extraction} \label{sec: encoding}

We aim to design an efficient equivariant feature encoder for volumetric data. The feature encoder generates an equivariant feature field, allowing us to query equivariant features using 3D coordinates. We implement this feature field with a tri-plane representation, which projects the volumetric primitive features onto three canonical planes. 
Our key finding is that the tri-plane representation preserves equivariance to rotations by multiples of 90$^\circ$ about the vertical axis: the horizontal-plane features are $\textnormal{C}_4$-equivariant under in-plane rotations, while the two vertical-plane feature maps are jointly $\textnormal{C}_4$-equivariant under the induced plane permutations and in-plane coordinate reflections.
Our primary contribution is the design of a tri-plane UNet that processes tri-plane features while preserving their equivariance.


This section first discusses the generation of tri-plane features from the projection of 3D data to three canonical planes (Sec.~\ref{sec: data projection}). Next, it discusses how tri-plane features retain a form of equivariance to $90^\circ$ Z-rotations of the workspace (Sec.~\ref{sec: equi tri plane}). Finally, Sec.~\ref{sec: equi feat proc} details the design of filters that enrich the tri-plane features while preserving the equivariance described in Sec.~\ref{sec: equi tri plane}.

\subsection{Data Projection} \label{sec: data projection}

The tri-plane representation is obtained from our volumetric input $\mathcal{V} \in \mathbb{R}^{S \times S \times S}$, a truncated signed distance function (TSDF). We first use a single-layer 3D CNN (lifting convolution) to transform TSDF into volumetric primitive features. Next, we project the features into the canonical planes $XY$, $XZ$, and $YZ$ via average pooling. Following our earlier convention, to simplify mathematical expressions, we view feature maps as vector fields. We denote the volumetric primitive features by $f: \mathbb R^3 \rightarrow \mathbb R^{C}$, and the three projections of $f$ yield the \emph{tri-plane primitive feature fields} $f_{\text{xy}} : \mathbb R^2 \rightarrow \mathbb R^{C}$, $f_{\text{xz}} : \mathbb R^2 \rightarrow \mathbb R^{C}$, $f_{\text{yz}} : \mathbb R^2 \rightarrow \mathbb R^{C}$, where $C$ is the number of feature channels.

\subsection{Equivariant Tri-plane Representation} \label{sec: equi tri plane}
To impose $\textnormal{C}_4$ equivariance on the tri-plane representation, we identify how each projected feature transforms under the group action. Once we identify these transformations, we can apply steerable CNNs to extract features from the patterns and transform them into the desired output. Specifically, the $XY$ projection preserves $Z$-rotation equivariance trivially, while the $XZ$ and $YZ$ projections transform jointly under an induced $\textnormal{C}_4$ action. The following paragraphs provide an intuitive explanation of these findings.

\begin{figure}[!t]
  \centering
  \includegraphics[width=\linewidth]{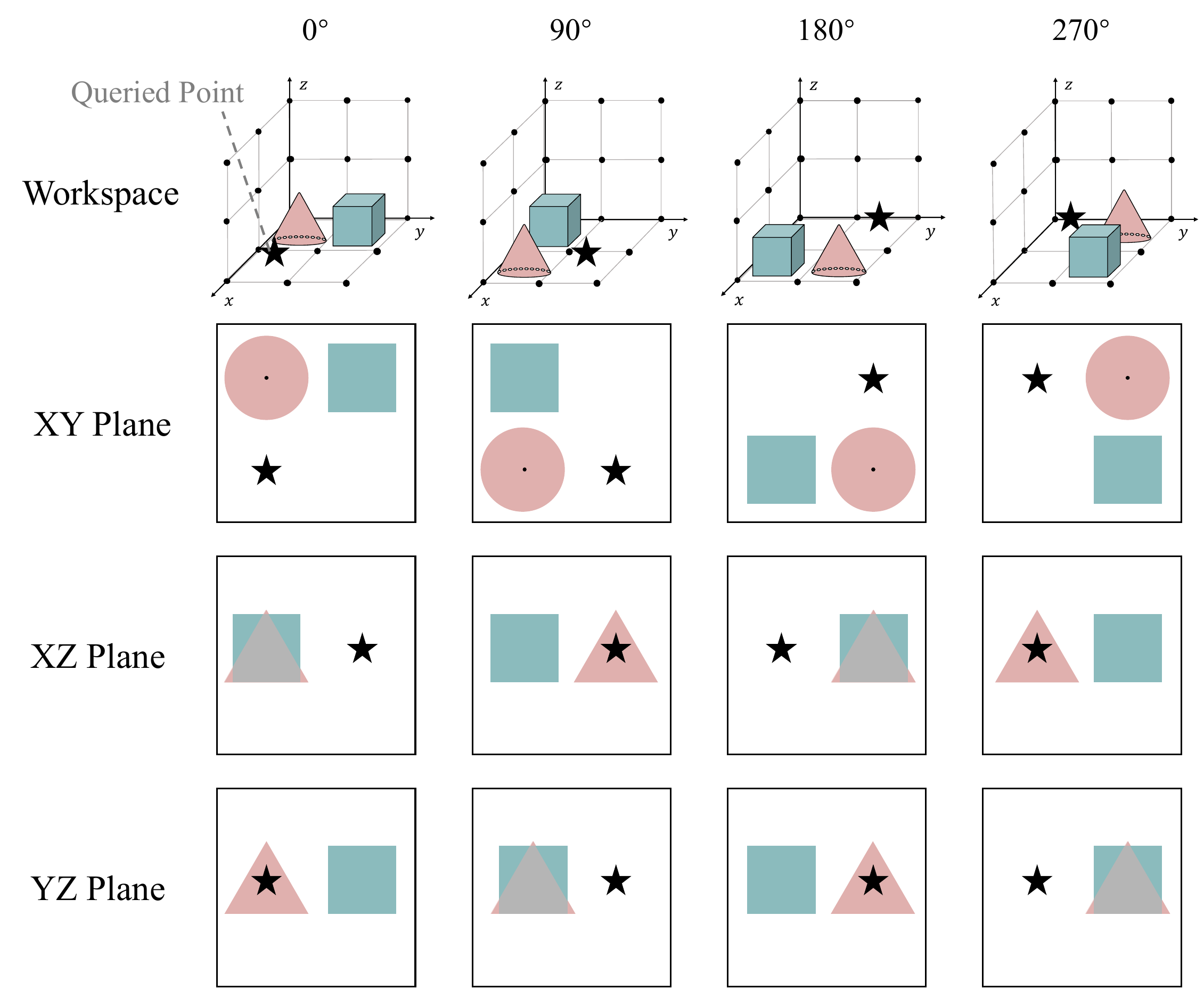}
  \caption{Illustration of equivariant manifestations of $90^\circ$ $Z$-axis rotations of a 3D scene composed of a cube and a cone. Rows labeled $XY$, $XZ$, $YZ$ show projections of the scene on the three planes. Each column corresponds to a $90^\circ$ $Z$-axis rotation of the one that precedes it. The $XY$ plane transforms equivariantly under the $\textnormal{C}_4$ group. $XZ$ and $YZ$ do not exhibit trivial equivariance, but they do exhibit a predictable response to $90^\circ$ $Z$-axis rotations in the form of $XZ$-$YZ$ permutations occasionally accompanied by an in-plane reflection. The black star is the queried position.}
  \label{fig: triequi}
\end{figure}

To support the discussion, we consider the example illustrated in Fig.~\ref{fig: triequi}. We assume a workspace that contains a pink cone and a blue cube, and we visualize three $90^\circ Z$-rotations of the workspace and its projections onto the three canonical planes. We observe that the $XY$ plane transforms consistently with the workspace: a $Z$ rotation of the workspace induces an in-plane rotation by the same angle. Rotation equivariance is therefore preserved trivially in this plane. The response of the two vertical planes is more complex, but regularity exists: Each time the workspace rotates by $90^\circ$, the two planes permute, and the $XZ$ plane undergoes a reflection.
Importantly, the ordered pair of features on the two vertical planes follows a predictable transformation under $\textnormal{C}_4$, consisting of a permutation of the two planes combined with an in-plane reflection. To illustrate this structure, we consider the 2D features corresponding to an arbitrary 3D query point, marked by the black star in Fig.~\ref{fig: triequi}. As the workspace rotates, the query point rotates accordingly and is projected onto both vertical planes. At $0^\circ$, the corresponding feature pair is $(\emptyset,\pinktriangle)$. Successive $90^\circ$ rotations transform this ordered pair as \[ (\emptyset,\pinktriangle) \rightarrow (\tilde{\pinktriangle},\emptyset) \rightarrow (\tilde{\emptyset},\tilde{\pinktriangle}) \rightarrow (\pinktriangle,\tilde{\emptyset}) \rightarrow (\emptyset,\pinktriangle), \] where $\tilde{\cdot}$ denotes the corresponding in-plane reflection. The pair therefore undergoes a cyclic transformation consistent with the action of $\textnormal{C}_4$. Let us denote by $g$ the group element representing the 90$^\circ$ rotation in $\textnormal{C}_4$, and by $[g \triangleright f]_{\text{xy}}$ the $XY$ projection of $g \triangleright f$ (similar for the $XZ$ and $YZ$ planes). Mathematically, the equivariance of the $XY$ plane described above can be written as:
\begin{equation}
     [g \triangleright f]_{\text{xy}}(\mathbf{p}) = f_{\text{xy}}(g^{-1} \cdot \mathbf{p}). \label{eq: xy_equi}
\end{equation}
The equivariance of the $XZ$ and $YZ$ planes can be written as:
\begin{equation}
    \begin{aligned}
    \begin{bmatrix}
    [g \triangleright f]_{\text{xz}}(\mathbf{p}) \\
    [g \triangleright f]_{\text{yz}}(\mathbf{p})   \\
    \end{bmatrix}
    =
    \begin{bmatrix}
     f_{\text{yz}}(g^{-1} \cdot \mathbf{p})\\ f_{\text{xz}}(g^{-1} \cdot \mathbf{p})
    \end{bmatrix}
    =
    \bar \rho_{\mathrm{s}}(g) \begin{bmatrix}
     f_{\text{xz}}(g^{-1} \cdot \mathbf{p})\\ f_{\text{yz}}(g^{-1} \cdot \mathbf{p})
    \end{bmatrix},
    \label{eq: side_equi}
    \end{aligned}
\end{equation}
where $\bar \rho_{\mathrm{s}}(g)
=
\begin{bmatrix}
0 & 1 \\
1 & 0
\end{bmatrix}
$ denotes the permutation of the $XZ$- and $YZ$-plane features induced by a $90^\circ$ rotation.
We note that, in the equations above, the query point $\mathbf p$ (the black star in Fig.~\ref{fig: triequi}) belongs to $\mathbb R^3$, whereas the domain of the plane-projected fields is $\mathbb R^2$. For compactness, we assume that the evaluation of a 2D field at a 3D point implicitly ignores the coordinate along which the field has been projected. For instance, $f_\mathrm{xy}(\mathbf p) = f_\mathrm{xy}(\mathbf p_\mathrm{x}, \mathbf p_\mathrm{y})$. To be consistent with the discussion in this subsection, we set $f$ as the volumetric feature field constructed by trivial features. Readers can refer to \emph{Lemmas 1.1 and 1.2} in the Appendix for more complicated cases and their proofs.

In summary, tableplane features transform \emph{equivariantly} to $\textnormal{SO}(2) \ Z$ rotations, and the \emph{combination} of matching features in the $XZ$ and $YZ$ planes is \emph{equivariant} to $90^\circ \ Z$ rotations. Together, the three tri-plane features are $\textnormal{C}_4$ $Z$-axis rotation equivariant. Therefore, if we implement the lifting convolution with a 3D steerable CNN, we can obtain a $\textnormal{C}_4$-equivariant tri-plane representation after projection.


\subsection{Equivariant Feature Processing} \label{sec: equi feat proc}
\begin{figure*}[!t]
  \centering
  \includegraphics[width=\linewidth]{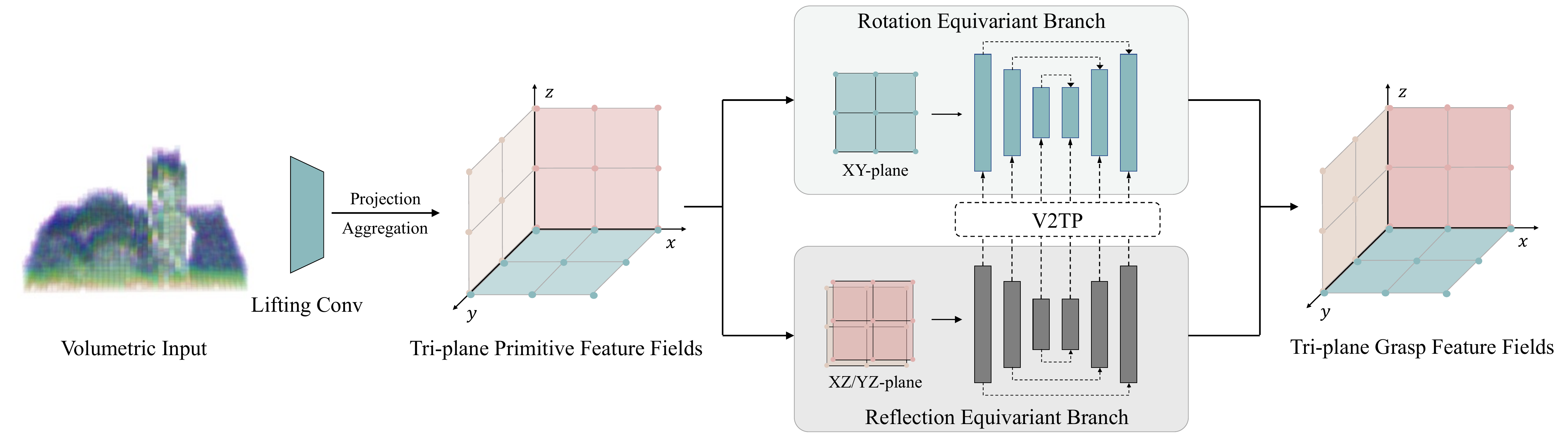}
  \caption{The architecture of the Equivariant Tri-plane UNet. Given a TSDF input, the model applies a single 3D steerable convolution to lift the volumetric data into a 3D feature field, which is then projected onto three canonical planes. The $XY$-plane features are processed by a $\textnormal{C}_4$-equivariant UNet, while the $XZ$- and $YZ$-plane features are handled by a reflection-equivariant UNet. The outputs from both branches are fused into a unified tri-plane feature field. A vertical-to-tableplane feature augmentation (V2TP) module enables communication between the two equivariant branches, enhancing feature representation through structured interactions.}
  \label{fig: triunet}
\end{figure*}

To further process the tri-plane features and maintain their equivariance, we propose a dual-branch network (Fig.~\ref{fig: triunet}). The first branch, $h_{\text{tp}}(\cdot)$, processes the $XY$ plane. It consists of a steerable CNN-based UNet designed to be equivariant under $g \in \textnormal{C}_4$, yielding a refined tableplane feature field $\hat{f}_{\text{xy}}$. The second branch, $h_{\text{s}}(\cdot)$, processes the $XZ$ and $YZ$ planes. It consists of a reflection-equivariant UNet designed to be equivariant under $s \in \textnormal{C}_2$, and yields $\hat{f}_{\text{xz}}$ and $\hat{f}_{\text{yz}}$. Let us denote by $\rho_{\text{tp}}(g)$ a group representation acting on features $\hat{f}_{\text{xy}}$, which is a direct sum of multiple $\rho_\textnormal{reg}(g)$. Building upon Eq.~\ref{eq: xy_equi}, the preservation of equivariance through the $XY$ projection \emph{and} the $h_{\text{tp}}(\cdot)$ model can be written as:
\begin{equation}
\begin{aligned}
    & \left[ h_{\text{tp}} \cdot [g \triangleright f]_{\text{xy}}\right](\mathbf{p}) \\
    &\quad = \rho_{\text{tp}}(g)[h_{\text{tp}} \cdot f_{\text{xy}}](g^{-1} \cdot \mathbf{p}) = \rho_{\text{tp}}(g)\hat{f}_{\text{xy}}(g^{-1} \cdot \mathbf{p}). \label{eq: xyunet}
    \end{aligned}
\end{equation}
In turn, building upon Eq.~\ref{eq: side_equi}, the preservation of equivariance of the vertical-plane pair through the vertical projection \emph{and} the $h_{\text{s}}(\cdot)$ model can be written as:
\begin{equation}
    \begin{aligned}
    \begin{bmatrix}
    \left[ h_{\mathrm{s}}\cdot  [g\triangleright f]_{\mathrm{xz}}\right](\mathbf p)
    \\
    \left[ h_{\mathrm{s}}\cdot  [g\triangleright f]_{\mathrm{yz}}\right](\mathbf p)
    \end{bmatrix}
     & =
    \rho_{\mathrm{s}}(g)
    \begin{bmatrix}
    \hat f_{\mathrm{xz}}(g^{-1}\cdot\mathbf p)
    \\
    \hat f_{\mathrm{yz}} (g^{-1}\cdot\mathbf p)
    \end{bmatrix} \\  & = \rho_{\mathrm{s}}(g) \hat f_{{\textnormal{vt}}}(g^{-1} \cdot \mathbf p),
    \label{eq: sideunet}
    \end{aligned}
\end{equation}
where $\rho_{\mathrm{s}}(g)$ is a group representation acting on {vertical-plane} features $\hat f_{{\textnormal{vt}}}$.
Readers can refer to \emph{Proposition 1} in the Appendix for the proof of Eq.~\ref{eq: xyunet} and Eq.~\ref{eq: sideunet}.

Similarly to the \emph{tri-plane primitive feature fields} ${f}_{\text{xy}}$, ${f}_{\text{xz}}$ and ${f}_{\text{yz}}$, the \emph{tri-plane grasp feature fields} $\hat{f}_{\text{xy}}$, $\hat{f}_{\text{xz}}$ and $\hat{f}_{\text{yz}}$ form a new implicit tri-plane feature space. We define the feature queried at position $\mathbf{p}$ in tri-plane grasp feature fields as: 
\begin{equation}
    \mathbf{c}(\mathbf{p}) = [\hat{f}_{\text{xy}}(\mathbf{p}), \hat f_{{\textnormal{vt}}}(\mathbf p)] = [\hat{f}_{\text{xy}}(\mathbf{p}),\hat{f}_{\text{xz}} (\mathbf{p}), \hat{f}_{\text{yz}}(\mathbf{p})]
\end{equation}
where $\hat{f}_{{ij}} (\mathbf{p})$ ($i,j \in \{\textnormal{x},\textnormal{y},\textnormal{z}\}$) is the queried feature at position $\mathbf{p}$ on the projected plane via bilinear interpolation. 
The tri-plane feature representation $\mathbf{c}$ is equivariant as:
\begin{equation}
\begin{aligned}
    & [g\triangleright \mathbf{c}](\mathbf{p}) \\
    & = [g\triangleright_{\rho_{\text{tp}}} \hat{f}_{\text{xy}}](\mathbf{p}) \oplus  [g\triangleright_{\rho_{{\textnormal{s}}}} \hat{f}_{\text{vt}}](\mathbf{p}) \\
    & = (\rho_\text{tp} \oplus \rho_s)(g) [\hat{f}_{\text{xy}}(g^{-1}\mathbf{p}) \oplus \hat{f}_{\text{vt}}(g^{-1}\mathbf{p})] .
\end{aligned}
\end{equation}

The dual-branch design is necessary to preserve the equivariance of the tri-plane representation during feature processing. Although the projected features obey a predictable transformation rule, a standard tri-plane UNet based on ordinary 2D convolutions does not, in general, preserve this rule after successive convolutions and feature fusion. Under a $\textnormal{C}_4$ rotation about the $Z$-axis, the $XY$ plane undergoes an in-plane rotation, whereas the $XZ$/$YZ$ pair undergoes a joint permutation-reflection action. We therefore process the $XY$ plane with a $\textnormal{C}_4$-equivariant branch and the two vertical planes with a reflection-equivariant branch. Assigning these branches in the opposite way would not match the respective induced group actions and would break the equivariance of the resulting tri-plane feature fields.

\begin{figure}[!t]
  \centering
  \includegraphics[width=\linewidth]{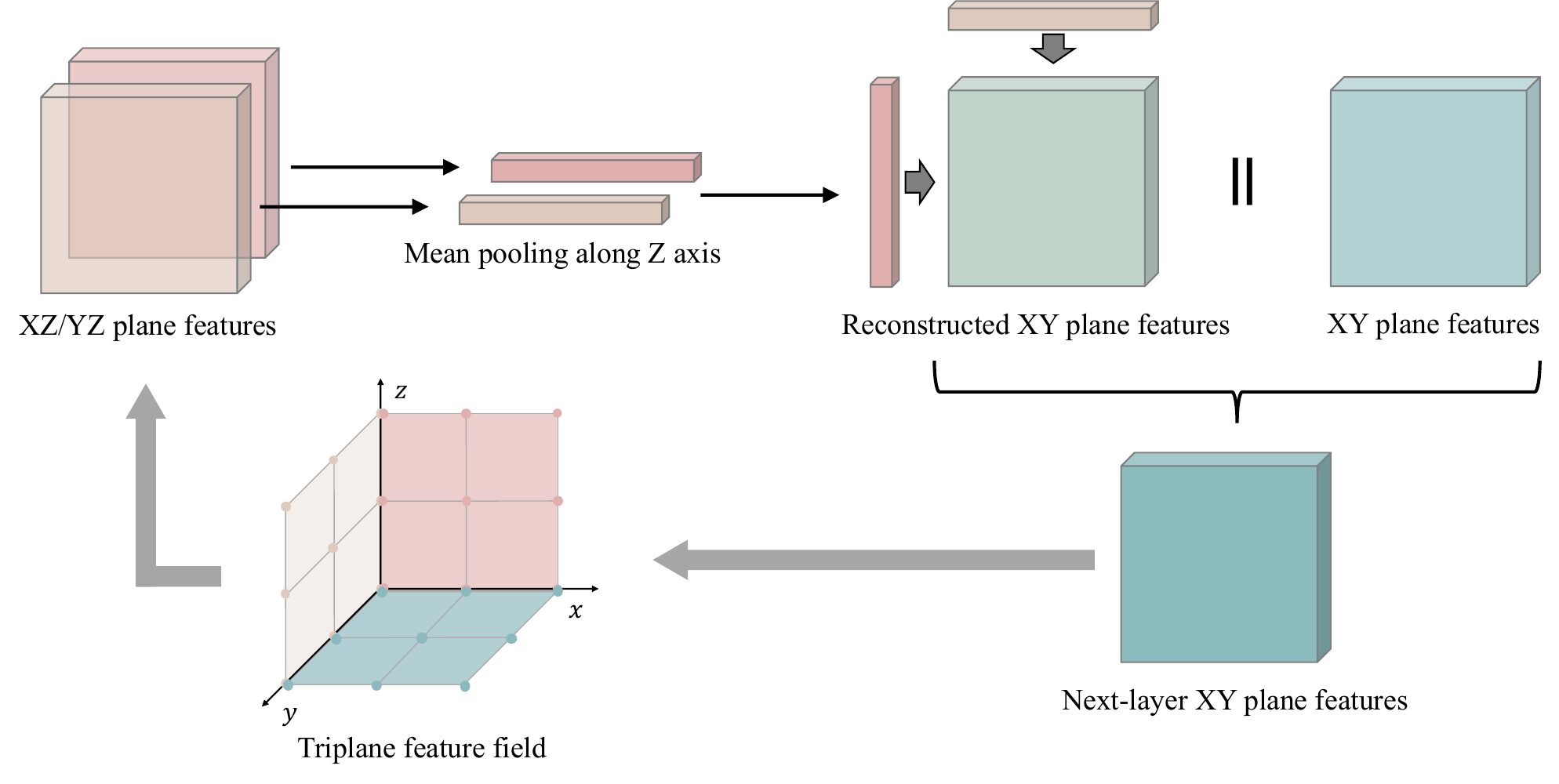}
  \caption{Illustration of the {vertical-to-tableplane} feature augmentation. Features from the vertical planes (XZ and YZ) are mean-pooled along the Z-axis to reconstruct auxiliary XY-plane features. These reconstructed features are then fused with the original XY-plane features to form the updated XY-plane representation for the features at the next UNet layer.}
  \label{fig: v2tp}
\end{figure}

\noindent \textbf{{Vertical-to-Tableplane Feature Augmentation}}.
Projecting volumetric features onto individual planes and processing each plane independently can limit the exchange of geometric information across the three views. To enrich the tableplane representation with information encoded by the two vertical planes, we introduce {vertical-to-tableplane} feature augmentation ({V2TP}), as illustrated in Fig.~\ref{fig: v2tp}. At layer $l$, let $f^l_{\mathrm{xz}}$ and $f^l_{\mathrm{yz}}$ denote the features on the two vertical planes. Each vertical-plane feature is composed of $\textnormal{C}_2$ regular fields, whose two components are denoted by $e$ and $s$, corresponding to the identity and reflection elements of $\textnormal{C}_2$. We first mean-pool both features along the $Z$-axis, obtaining $\bar f^l_{\mathrm{xz}}(x)$ and $\bar f^l_{\mathrm{yz}}(y)$
\begin{align}
    \bar f^l_{\mathrm{xz}}(x)
    &= \operatorname{MeanPool}_{z}
    \left(f^l_{\mathrm{xz}}(x,z)\right),\\
    \bar f^l_{\mathrm{yz}}(y)
    &= \operatorname{MeanPool}_{z}
    \left(f^l_{\mathrm{yz}}(y,z)\right).
\end{align}
The resulting one-dimensional features are then broadcast along the complementary spatial dimension to reconstruct features on the $XY$ plane. For each pair of corresponding $\textnormal{C}_2$ regular fields, we arrange the reconstructed features as
\begin{equation}
\bar f^l_{\mathrm{s}\rightarrow\mathrm{xy}}(x,y)
= [\bar f^l_{\mathrm{xz},e}(x),
    \bar f^l_{\mathrm{yz},e}(y),
    \bar f^l_{\mathrm{xz},s}(x),
    \bar f^l_{\mathrm{yz},s}(y)]
    \label{eq:v2tp_induced}
\end{equation}
which realizes the induced $\textnormal{C}_4$ feature ordering carried jointly by the $XZ$ and $YZ$ planes.

Finally, the reconstructed {vertical-plane} features are concatenated with the current $XY$-plane features and fused using a point-wise $\textnormal{C}_4$-equivariant convolution:
\begin{equation}
    f^{l+1}_{\mathrm{xy}}
    =
    F_l\left(
    f^l_{\mathrm{xy}}
    \,\Vert\,
    \bar f^l_{\mathrm{s}\rightarrow\mathrm{xy}}
    \right),
\label{eq:v2tp_fusion}
\end{equation}
where $\Vert$ denotes channel-wise concatenation and $F_l(\cdot)$ is a $1\times1$ $\textnormal{C}_4$-equivariant convolution. {V2TP} is applied at multiple encoder and decoder levels, allowing the $XY$ branch to incorporate complementary geometric information from both vertical planes while preserving the induced $\textnormal{C}_4$ transformation structure. Appendix 4 proves this equivariance property.


\section{Decoding and Evaluating Grasp Poses} \label{sec: decoding}
\begin{figure*} [t!]
\centering
    \resizebox{\linewidth}{!}{
    \subfloat[\label{fig:equi_giga}]{
    \includegraphics[scale=0.37]{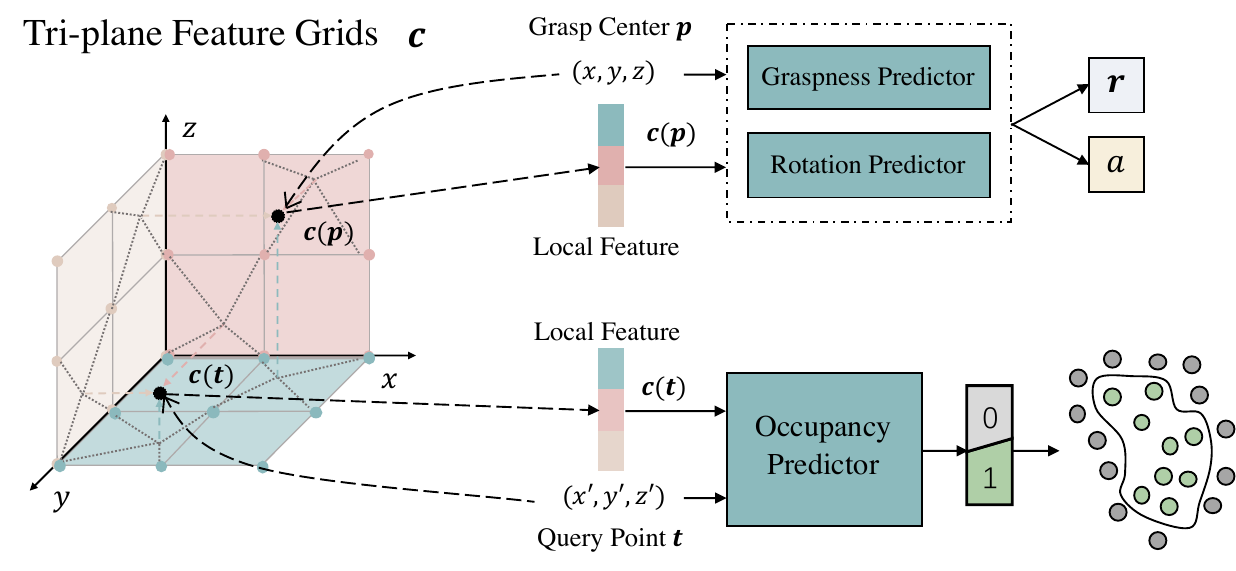}}
    \subfloat[\label{fig:equi_igd}]{
    \includegraphics[scale=0.37]{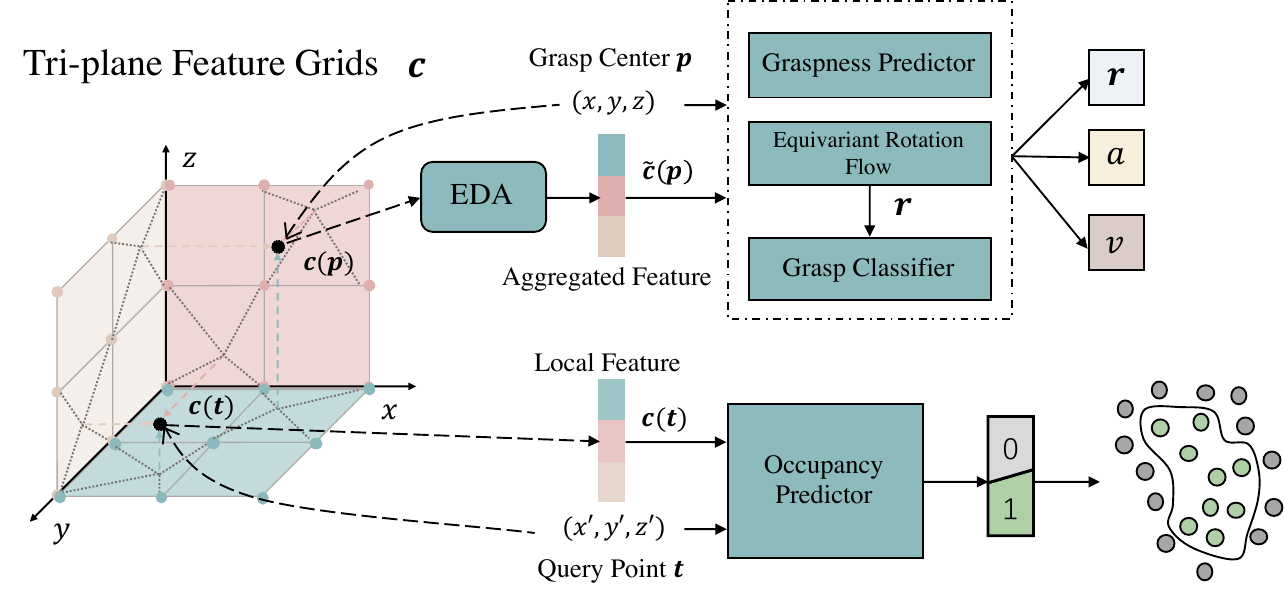}}
    }
    \caption{Illustration of the EquiGIGA and EquiIGD workflows. This figure is based on graphical elements from GIGA \citep{giga} and IGD \citep{igd}. Given a feature queried at position $\mathbf{p}$ from the tri-plane feature grid, (a) EquiGIGA directly regresses the rotation $\mathbf{r}$ and predicts the graspness $a$ at $\mathbf{p}$; (b) EquiIGD uses Equivariant Rotation Flow to sample $\mathbf{r}$ and additionally evaluates grasps with the grasp classifier $v$.}
\label{fig: decoder}
\end{figure*}

The proposed Equivariant Tri-plane UNet transforms the volumetric data into an equivariant implicit feature space $\mathbf{c}$, through which features at any position in the workspace can be queried. In this section, we leverage this feature space to design rotation-equivariant grasp models. We build upon GIGA \citep{giga} (direct grasp regression) and IGD \citep{igd} (diffusion-based rotation prediction), both of which are designed atop a non-rotation-equivariant tri-plane representation. We adjust these two models to use our new rotation-equivariant tri-plane representation and preserve $\textnormal{C}_4$ equivariance to vertical-axis rotations throughout grasp generation, yielding models that we refer to as EquiGIGA and EquiIGD. Fig.~\ref{fig: decoder}~(a,b) illustrates the workflows of these two models.

To realize this adaptation, we first identify the feature types (Sec.~\ref{sec: group act}) that adequately describe the transformation of the variables predicted by either model. In turn, we describe the use of equivariant linear layers to transform the queried equivariant features into the desired feature types.

\begin{figure*} [t!]
\centering
    \resizebox{\linewidth}{!}{
    \subfloat[\label{fig:EDA}]{
    \includegraphics[scale=0.38]{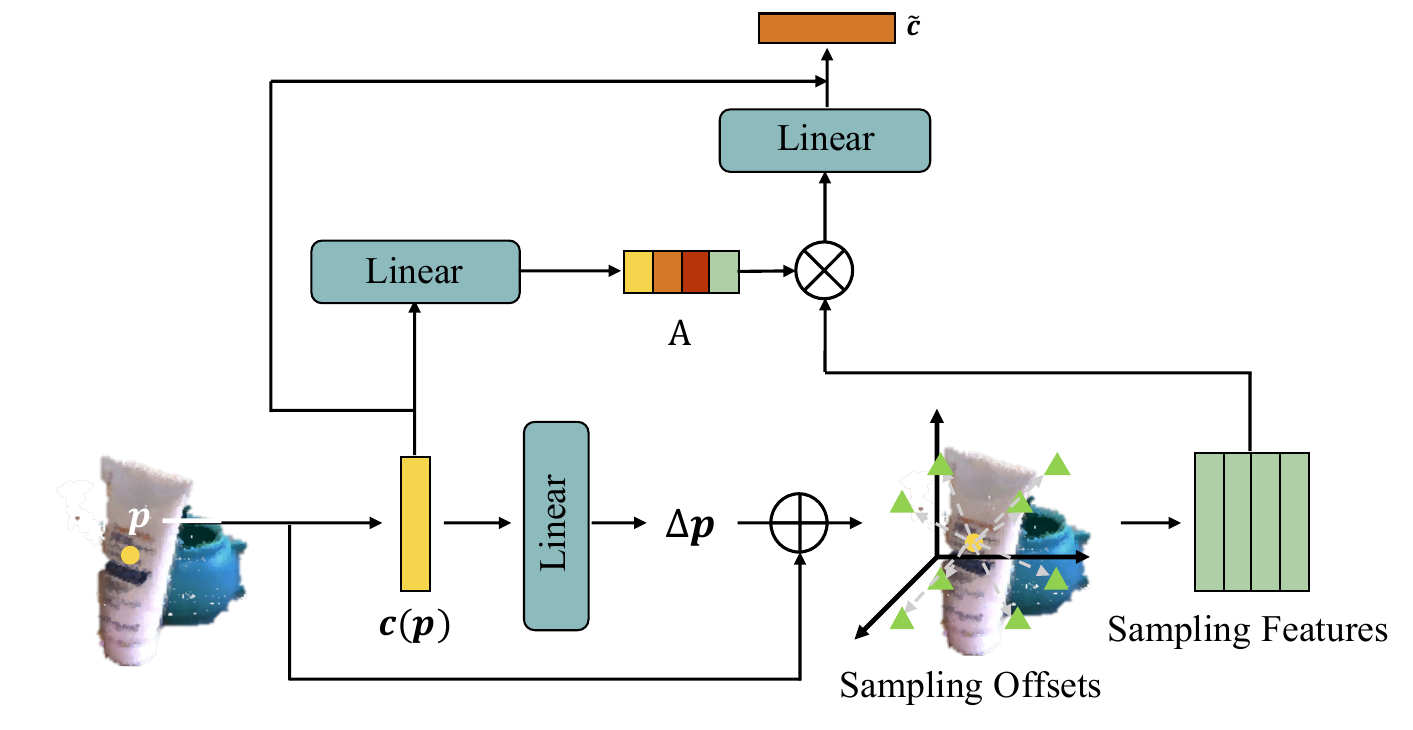}}
    \subfloat[\label{fig:graspDAM}]{
    \includegraphics[scale=0.38]{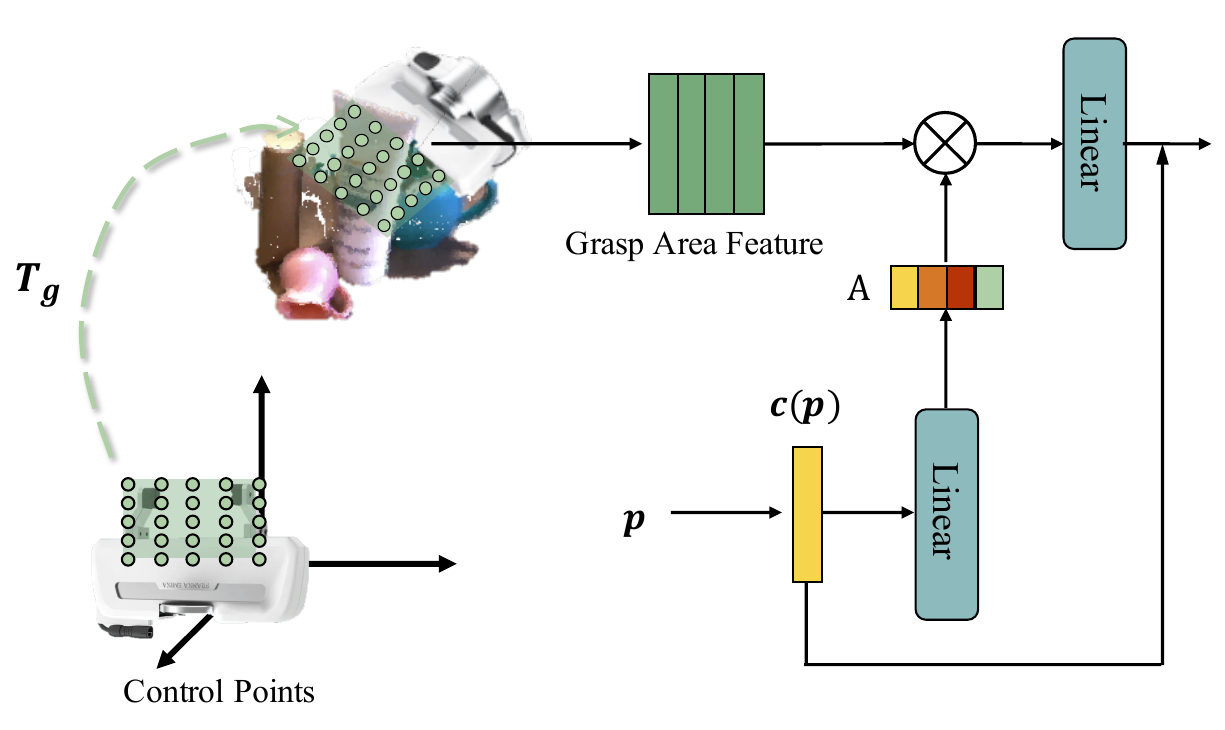}
    }
    }
    \caption{(a) The architecture of Equivariant Deformable Attention. (b) The architecture of Equivariant Grasp-conditioned Deformable Attention Module. This figure is based on graphical elements from DAM and GraspDAM in IGD \citep{igd}. See text for details.}
\label{fig:DAMs}
\end{figure*}

\subsection{Equivariant GIGA (EquiGIGA)}
Based on the features $\mathbf{c}(\mathbf{p})$ queried at position $\mathbf{p}$,  GIGA \citep{giga} regresses the graspness $a$, the occupancy $o$, and the rotation $\mathbf{r}$, where graspness denotes whether the position contains a valid grasp, occupancy denotes whether the position is occupied. The heads $h_{\text{a}}(\cdot)$, $h_{\text{o}}(\cdot)$, and $h_{\text{rot}}(\cdot)$ are equivariant linear layers used to decode graspness, occupancy, and rotation. Graspness $a$ and occupancy $o$ are invariant under $\textnormal{SO}(2)$ transformations, as $a(g\mathbf{p}, g\mathbf{c}) = a(\mathbf{p},\mathbf{c})$ and $o(g\mathbf{p},g\mathbf{c}) = o(\mathbf{p},\mathbf{c})$. We can define them as type-0 features. We encode the grasp rotation by the six-dimensional vector formed from the first two entries of each rotation-matrix column. This representation consists of three copies of the type-1 representation of $\textnormal{SO}(2)$ restricted to $\textnormal{C}_4$, so that, for $g\in\textnormal{C}_4$, $g \cdot \mathbf{r}=(\bigoplus_{i=1}^{3}\rho_1(g))\mathbf{r}$. The proof of this statement can be found in \emph{Proposition 2} in the Appendix.

The training loss of EquiGIGA is formulated as:
\begin{equation}
    \mathcal{L}_{\text{EquiGIGA}}= \mathcal{L}_{\text{rot}} + \mathcal{L}_{\text{graspness}} + \mathcal{L}_{\text{occ}},
\end{equation}
where the rotation loss $\mathcal{L}_{\text{rot}}$ is obtained by transforming the predicted rotation into a quaternion and calculating the negative cosine similarity between the predicted rotation and ground-truth rotations. The graspness loss $\mathcal{L}_{\text{graspness}}$ is the cross-entropy loss between the predicted graspness and the ground-truth graspness. The occupancy loss $\mathcal{L}_{\text{occ}}$ is the cross-entropy loss between the predicted occupancy and the ground-truth occupancy.

\subsection{Equivariant IGD (EquiIGD)} \label{sec:equiigd}
Based on GIGA, IGD \citep{igd} further proposes (i) a Deformable Attention Module for feature extraction, (ii) a grasp orientation diffuser for rotation prediction, and (iii) a grasp classifier with a Grasp-conditioned Deformable Attention Module. These three design choices make IGD a competitive alternative to GIGA: while its compute cost is higher than GIGA, it yields superior gripper placement, and captures multi-modal gripper orientation distributions---a key trait of grasping tasks. In the following text, we propose the equivariant counterparts of IGD's deformable attention, orientation diffusion and grasp classifier.

\subsubsection{Equivariant Deformable Attention}
The Deformable Attention Module in IGD aims to dynamically gather features from $\mathbf{p}$'s neighborhood by sampling nearby points and aggregating their features. We propose the Equivariant Deformable Attention (EDA; Fig.~\ref{fig:DAMs}~(a)), in which the sampled points and the aggregated features are equivariant to the group transformation. The proposed EDA is formulated as:
\begin{align}
    & \tilde{\mathbf{c}}(\mathbf{p}) = h_{\text{out}}\left(\sum_{k=1}^{K} A_k(\mathbf{p}) \cdot h_{\text{in}} ( \mathbf{c}(\mathbf{p}+\Delta_k(\mathbf{p})))\right)+\mathbf{c}(\mathbf{p}), \\
    & A_k(\mathbf{p}) = h^k_{\text{attn}}(\mathbf{c}(\mathbf{p})), \qquad \Delta_k(\mathbf{p}) = h^k_{\text{offset}}(\mathbf{c}(\mathbf{p})),
\end{align}
where $K$ is the number of sampling offsets, and $h_{\text{in}}$, $h_{\text{out}}$, $h^k_{\text{attn}}$, $h^k_{\text{offset}}$ are equivariant linear layers. $\Delta_k$ is the $k$-th sampling offset, which is represented as a concatenation of type-1 and type-0 features, as $\Delta_k(g\cdot\mathbf{p}) = (\rho_1(g) \oplus \rho_0(g)) \Delta_k(\mathbf{p})$. $A_k$ is the attention weight for the $k$-th offset $\Delta_k$, which is represented as a type-0 feature. In EquiIGD, the denoising of gripper rotations and decoding of grasp quality at location $\mathbf{p}$ is conditioned on $\tilde{\mathbf{c}}(\mathbf{p})$ instead of $\mathbf{c}(\mathbf{p})$.

\subsubsection{Equivariant Rotation Flow}

The generative model employed in IGD is diffusion: during training, noise was iteratively added to the grasp-rotation data \(\mathbf{r}\), and a denoiser was then trained to reverse the process. Recently, flow matching has demonstrated expressiveness that is competitive with diffusion at a lower cost \citep{liu2022flow}. Accordingly, EquiIGD employs flow matching instead of diffusion. 
In flow matching, a continuous time-dependent velocity field transports samples from a base distribution to the target distribution without explicitly defining a noise-injection process. Let $\mathbf{r}_0$ be a rotation representation sampled from the base distribution $q_0=\mathcal{N}(\mathbf{0},\mathbf{I})$, and let $\mathbf{r}_1$ be a target grasp rotation sampled from the data distribution $q_1$. For $t \in [0,1]$, we define the conditional probability path by linear interpolation:
\begin{align}
    &\mathbf{r}_t = (1-t)\mathbf{r}_0 + t\mathbf{r}_1, \\
    &\mathbf{u}_t(\mathbf{r}_t \mid \mathbf{r}_0,\mathbf{r}_1)
    = \frac{d\mathbf{r}_t}{dt} = \mathbf{r}_1-\mathbf{r}_0, \\
    & \mathbf{r}_0 \sim q_0=\mathcal{N}(\mathbf{0},\mathbf{I}), \qquad \mathbf{r}_1 \sim q_1, \qquad t \sim \mathcal{U}(0,1).
\end{align}
We learn a conditional velocity field $\mathbf{v}(\mathbf{r},\mathbf{c}(\mathbf{p}),t)$ by matching it to the conditional velocity field $\mathbf{u}_t$. The flow-matching objective is
\begin{equation}
    \mathcal{L}_{\mathrm{flow}}
    =
    \mathbb{E}_{\substack{\mathbf{r}_0 \sim q_0,\, \mathbf{r}_1 \sim q_1,\\ t \sim \mathcal{U}(0,1),\, \mathbf{p}}}
    \left[
        \left\|\mathbf{v}(\mathbf{r}_t,\mathbf{c}(\mathbf{p}),t) - (\mathbf{r}_1-\mathbf{r}_0)\right\|^2
    \right].
    \label{eq:flow-loss}
\end{equation}
After training, we generate a grasp rotation by sampling $\mathbf{r}_0\sim q_0$ and integrating the learned ordinary differential equation from $t=0$ to $t=1$:
\begin{equation}
    \mathbf{r}_1 = \mathbf{r}_0 + \int_0^1 \mathbf{v}\bigl(\mathbf{r}_t,\,\mathbf{c}(\mathbf{p}),t\bigr)\, dt.
\end{equation}
Because $\mathbf{v}(\cdot)$ is constructed from equivariant linear layers, it satisfies
\begin{equation}
    \mathbf{v}(g\cdot \mathbf{r}_t,\, \mathbf{c}(g\cdot \mathbf{p}),\,t) = g\cdot \mathbf{v}(\mathbf{r}_t,\,\mathbf{c}(\mathbf{p}),\,t), \label{eq:flow-equivariance}
\end{equation}
for any group transformation $g \in \textnormal{C}_4$. Consequently, the generated rotation distribution is $\mathrm{C}_4$-equivariant.

\subsubsection{Equivariant Grasp Classifier}
The Equivariant Grasp Classifier models the grasp score $v$ of a grasp, and a higher $v$ denotes a more feasible grasp. To extract a feature to identify a grasp, we propose an Equivariant Grasp-conditioned Deformable Attention Module (EquiGraspDAM, illustrated in Fig.~\ref{fig:graspDAM}). We first define a set of gripper-relative learnable control points $\mathbf{u}_1,\ldots,\mathbf{u}_L$. To express these points in the scene's base frame, we transform them with a transformation $T_{\mathbf{g}}$ defined by the grasp $\mathbf{g}=(\mathbf{p}, \mathbf{r})$.
The feature identifying the grasp can be obtained by applying deformable attention to the features of control points, as:
\begin{align}
     & \mathbf{Q}(\mathbf{p}) = h_{\text{Q}}(\mathbf{c}(\mathbf{p})), \\
     & \mathbf{K}(\mathbf{p}, \mathbf{r}) = \{h_{\text{K}}(\mathbf{c}(T_{\mathbf{g}} \mathbf{u}_l))\}_{l=1}^L, \\
     & \mathbf{V}(\mathbf{p}, \mathbf{r}) = \{h_{\text{V}}(\mathbf{c}(T_{\mathbf{g}} \mathbf{u}_l) )\}_{l=1}^L,\\
    & \bar{\mathbf{c}}(\mathbf{p}, \mathbf{r}) = h_{\text{out}}\left(\mathbf{V} \mathop{\text{Softmax}}\left(\frac{\mathbf{K}^T \mathbf{Q}}{\sqrt{d_k}}\right) \right) + \mathbf{c}(\mathbf{p}),
\end{align}
where $h_{\text{Q}}$, $h_{\text{K}}$,  $h_{\text{V}}$, $h_{\text{out}}$ are equivariant linear layers. Five equivariant linear layers are applied to $\bar{\mathbf{c}}$ to compute a grasp score $v$. The proof of equivariance can be found in the Appendix. Following IGD's design, we use a focal loss to train the model, which can alleviate the positive-negative imbalance:
\begin{equation}
  \mathcal{L}_{\text{grasp}}(\hat{v}_{\mathbf{g}})=\left\{
  \begin{aligned}
  &- (1-\hat{v}_{\mathbf{g}})^{\gamma} \log(\hat{v}_{\mathbf{g}}),  & v_{\mathbf{g}}=1,\\
  &-\hat{v}_{\mathbf{g}}^\gamma \log(1-\hat{v}_{\mathbf{g}}),  &v_{\mathbf{g}}=0, \label{eq: grasp score loss}
  \end{aligned}
  \right.
\end{equation}
where $\hat{v}_{\mathbf{g}}$ is the predicted grasp score, while $v_{\mathbf{g}}$ is the ground-truth grasp label. $\gamma$ is the focusing parameter. With the prediction of the graspness $a$, we can obtain the final grasp quality as $q = a \cdot v$.

The training loss of EquiIGD is formulated as:
\begin{equation}
    \mathcal{L}_{\text{EquiIGD}}= \mathcal{L}_{\text{flow}} + \mathcal{L}_{\text{graspness}} + \mathcal{L}_{\text{grasp}} + \mathcal{L}_{\text{occ}},
\end{equation}
where the flow loss $\mathcal{L}_{\text{flow}}$ is defined in Eq.~\ref{eq:flow-loss}, and the grasp score loss $\mathcal{L}_{\text{grasp}}$ is defined in Eq.~\ref{eq: grasp score loss}. The graspness loss $\mathcal{L}_{\text{graspness}}$ and the occupancy loss $\mathcal{L}_{\text{occ}}$ are defined in the same way as for EquiGIGA.

\begin{figure*}[tp]
    \centering
    \includegraphics[width=\linewidth]{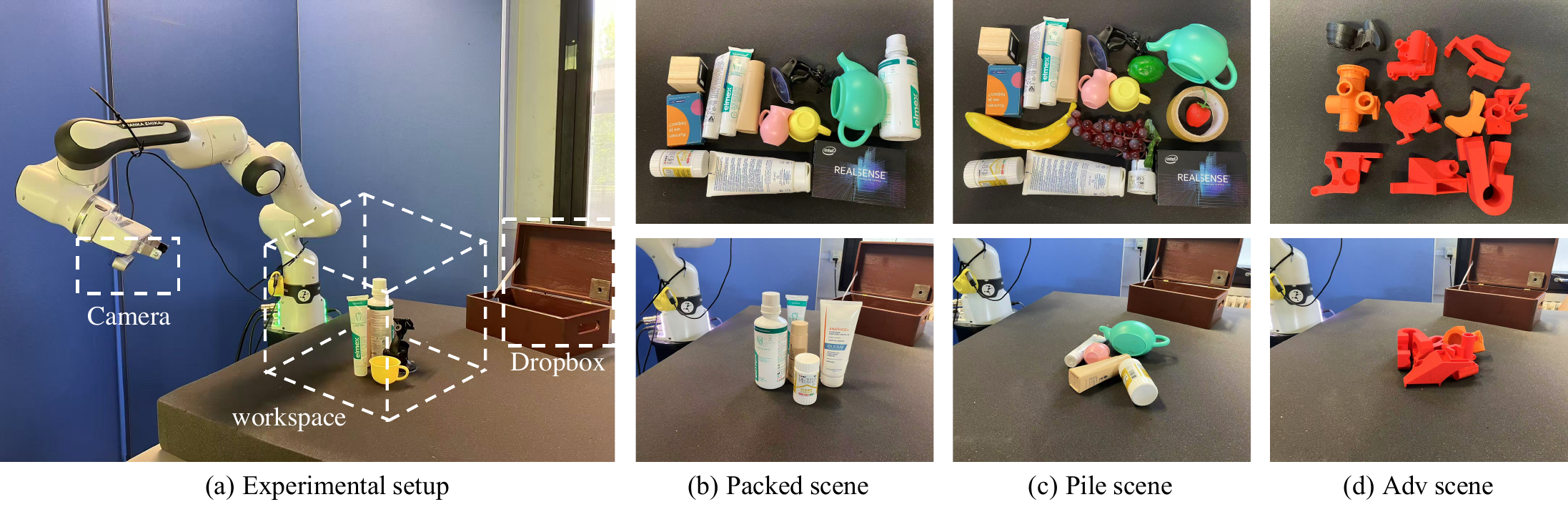}
    \caption{(a) Experimental setup for the real-world declutter experiment. (b) An illustration of the packed scene and the objects. (c) An illustration of the pile scene and the objects. (d) An illustration of the adversarial scene and the objects.}
    \label{fig:setup}
\end{figure*}

\section{Experiments}
In this section, we evaluate the proposed methods. We first compare EquiGIGA and EquiIGD with state-of-the-art methods in simulation and real-world experiments, then assess the individual components of the Equivariant Tri-plane UNet and EquiIGD's prediction heads through ablation studies. We further investigate sample efficiency, empirically verify output equivariance, and evaluate robustness to changes in camera viewpoint.

\subsection{Experimental Setup}

\noindent \textbf{Simulation Environment:}
Our simulated environment is built using PyBullet, featuring a free-floating gripper that samples grasps within a tabletop workspace measuring \(30 \times 30 \times 30 \, \text{cm}^3\). For a fair comparison, we employ the same object assets as VGN \citep{vgn} and GIGA \citep{giga}, including 303 training and 40 test objects from various datasets \citep{calli2015ycb, kappler2015leveraging, kasper2012kit, singh2014bigbird}. The model is validated in two types of simulated scenes: \emph{pile} and \emph{packed}. In the pile scene, objects are randomly dropped into a box of the same dimensions as the workspace, resulting in a cluttered pile once the box is removed. The packed scene features a subset of taller objects placed at random locations on the table in their canonical pose.
We validate our methods under two observation setups commonly used in robotic decluttering benchmarks: (i) the depth map is captured via a single fixed side-view camera pose. Gamma noise is injected into the depth map \citep{igd,giga}; (ii) the depth map is captured via a single random-view camera pose. Gaussian noise is injected into the depth map \citep{hu2024orbitgrasp,zurbrugg2024icgnet}. Because SE(2)-EquiGrasp requires top-down observations, it receives a fixed top-down view in both benchmarks.
The declutter process involves repeatedly predicting and executing a grasp, followed by removing the grasped object from the workspace until one of three conditions is met: all objects are cleared, two consecutive failures occur, or no grasp is detected. Results are averaged over 100 simulation rounds for each of five random seeds (500 rounds in total).

\noindent \textbf{Dataset Collection:}
We collect our grasping dataset following the setup used in ICGNet \citep{zurbrugg2024icgnet}. Using the described simulation environment, we select an observed surface point along with its corresponding surface normal and execute grasps from multiple approach angles. Specifically, for each contact point and final gripper width, we sample twelve different approach angles. Additionally, for each scene, we sample and store binary labels for 200,000 occupancy locations: 70\% are drawn uniformly across the volume and 30\% are concentrated near object surfaces to enable more accurate reconstructions. In total, we sample 1.5 million grasps across 5,000 scenes for the packed setting, and 4 million grasps across 10,000 scenes for the pile setting.

\noindent \textbf{Real-world Environment:} We conduct the real-world experiment on a Franka Research 3 robot arm with a RealSense D435i. In real-world experiments, we run 15 rounds for each of the packed, pile, and adversarial scenes. Everyday objects and ten Berkeley adversarial objects \citep{mahler2019learning} are used to conduct the experiments. In each round, 5 objects from a fixed pool (Fig.~\ref{fig:setup}) are randomly selected and placed on the table. In each grasp trial, we pass the TSDF or point cloud from a side-view depth camera to the model and execute the physically feasible grasp with the highest score.


\noindent \textbf{Metric:} (i) Grasp Success Rate (GSR = $\frac{\# \textnormal{successful grasps}}{\# \textnormal{total grasps}}$) which measures the ratio of successful grasps to total grasps; (ii) Declutter Rate (DR = $\frac{\# \textnormal{grasped objects}}{\# \textnormal{total objects}}$) which measures the ratio of objects removed successfully to the number of total objects presented.

\subsection{Training, Inference, and Comparison Models} \label{sec:training_inference}
We implement the proposed \emph{EquiGIGA} and \emph{EquiIGD} with \emph{PyTorch} and \emph{ESCNN} \citep{cesa2022program} and train the models with the \emph{Adam} optimizer for 12 epochs. An initial learning rate of $2 \times 10^{-4}$ is set. At inference time, we discretize the workspace volume into a $40^3$ voxel grid and use the centers of all voxel cells as grasp centers. We evaluate graspness (or grasp quality) at all grasp centers, and predict a grasp for each grasp center. Next, we mask out impractical grasps. The predicted grasp with the highest quality is selected if the graspness (or grasp quality) is beyond a threshold $T$. If no grasp has a graspness (or grasp quality) above $T$, we declare that there is no feasible grasp in the scene.


For the controlled comparisons in the following experiments, we construct non-equivariant variants of both GIGA and IGD. \emph{GIGA++} is an enhanced variant of GIGA that uses the dual-branch Tri-UNet backbone and {V2TP} introduced above, which corresponds to row 6 in Tab.~\ref{tab:ablation equi triunet}. \emph{GIGA++A} uses the same architecture as GIGA++ but is trained with data augmentation over the four $\textnormal{C}_4$ orientations ($0^\circ$, $90^\circ$, $180^\circ$, and $270^\circ$). Similarly, \emph{IGD++} is the non-equivariant counterpart of EquiIGD. It retains the same dual-branch Tri-plane UNet topology, V2TP module, flow-matching formulation, and remaining IGD components, but replaces all steerable convolutions with conventional convolutions. \emph{IGD++A} uses the same architecture as IGD++ and is trained with the same $\textnormal{C}_4$ augmentation. Unless otherwise specified, GIGA++, GIGA++A, IGD++, and IGD++A follow the same respective training and inference settings as EquiGIGA and EquiIGD.

\begin{table*}[t]
  \centering
  \caption{Simulated clutter-removal performance with a single fixed camera and gamma noise. We report mean $\pm$ standard deviation for GSR and DR. $N$ is the number of sampling rounds used by IGD and EquiIGD. Results marked ``*'' are from \citep{igd}, which uses the same experimental protocol and random seeds. $\dagger$ means that we use their open-source checkpoint to evaluate on this benchmark. Latency is measured from a depth map to predicted grasps on an RTX 4070 GPU, except for OrbitGrasp, which is measured on an RTX A6000 because it requires more than 40 GB of GPU memory. The best and second-best reported values in each column are shown in blue boldface and pink underlined text, respectively.}
  \resizebox{0.8\linewidth}{!}{
  \setlength{\tabcolsep}{3mm}{}
  \begin{tabular}{c|cc|cc|c}
    \toprule
    \multirow{2}*{Method}   & \multicolumn{2}{c}{Packed}  & \multicolumn{2}{c}{Pile}  & \multirow{2}*{Latency (ms)}\\
    & GSR (\%) & DR (\%) & GSR (\%) & DR (\%) & \\
    \midrule
    VGN* \citep{vgn}    & 72.5$\pm$2.6   & 76.7$\pm$1.7 & 59.3$\pm$2.9 & 43.5$\pm$2.9 & \boldcolor{First}{9} \\
    GIGA* \citep{giga} &  84.8$\pm$2.2  & 85.1$\pm$2.5   &  69.5$\pm$1.3 & 49.0$\pm$3.4  & 24\\
    GraspNet-1B Baselines* \citep{fang2020graspnet} &  {49.9$\pm$2.3}  & 40.1$\pm$2.2   &  50.2$\pm$4.2 & 30.0$\pm$2.3  & 77\\
    GSNet* \citep{wang2021graspness} &  67.8$\pm$2.5  & 60.1$\pm$3.2  &  58.3$\pm$3.8  & 51.3$\pm$4.6   & 156\\
    GPD* \citep{gpd} &  41.8$\pm$2.9  & 34.1$\pm$3.4  & 22.7$\pm$1.1  &  9.0$\pm$0.7 & 2138 \\
    6DoF-GraspNet* \citep{6dofgraspnet} &  17.9$\pm$0.8  &  11.9$\pm$0.9 & 15.5$\pm$2.9  &  6.9$\pm$1.1 & 2232 \\
    SE(3)-Dif* \citep{se3dif} &  7.2$\pm$1.5  &  4.3$\pm$1.0 & 7.6$\pm$1.8  & 3.0$\pm$0.8 & 5691 \\
    EdgeGraspNet$\dagger$ \citep{huang2023edge} &  54.1$\pm$2.1  &  54.0$\pm$2.7 & 50.5$\pm$3.7  & 43.0$\pm$4.8 & 843/685  \\
    VN-EdgeGraspNet$\dagger$ \citep{huang2023edge} &  60.6$\pm$2.2  &  60.1$\pm$3.8 & 55.0$\pm$2.1  & 50.1$\pm$4.0 & 1174/953 \\
    ICGNet$\dagger$ \citep{zurbrugg2024icgnet} & 60.3$\pm$4.1 & 64.5$\pm$5.9  & 57.3$\pm$1.5 & 51.7$\pm$3.3  & 806 \\
    DexGraspNet2$\dagger$ \citep{zhang2024dexgraspnet} &  51.6$\pm$2.5  &  53.9$\pm$4.3 & 39.7$\pm$1.3  & 30.9$\pm$2.2 & 2781 \\
    OrbitGrasp$\dagger$ \citep{hu2024orbitgrasp} & 71.1$\pm$1.8  & 72.8$\pm$1.6  & 69.3$\pm$2.1  &  64.7$\pm$3.3  & 3193 \\
    IGD* ($N$=1) \citep{igd} & 92.9$\pm$1.8  &  86.7$\pm$1.8  & 68.2$\pm$1.9 & 50.6$\pm$1.5  & 217\\
    IGD* ($N$=11) \citep{igd} & 91.2$\pm$0.9  &  88.8$\pm$1.5  & 71.8$\pm$2.2  &  55.7$\pm$2.6& 1823\\
    SE(2)-EquiGrasp$\dagger$ \citep{zhu2023robot} &  86.0$\pm$2.1  &  87.5$\pm$2.5 & 72.4$\pm$1.9  & 68.6$\pm$4.5 & \boldcolor{Second}{11} \\
    \midrule
    EquiGIGA & \boldcolor{Second}{94.5$\pm$1.3} & \boldcolor{Second}{91.4$\pm$1.2}  & \boldcolor{Second}{79.1$\pm$2.2}  & \boldcolor{First}{77.3$\pm$3.9} & 25 \\
    EquiIGD ($N$=1)& \boldcolor{First}{98.0$\pm$0.6}  & \boldcolor{First}{92.0$\pm$1.5}   & \boldcolor{First}{79.5$\pm$1.3}  & \boldcolor{Second}{77.0$\pm$1.1} & 52 \\
    \bottomrule
  \end{tabular}
  }
  \label{tab:sota}
\end{table*}

\begin{table*}[t]
  \centering
  \caption{Simulated clutter-removal performance with a single random-view camera and Gaussian noise. $N$ is the number of sampling rounds used by EquiIGD. We report mean $\pm$ standard deviation for GSR and DR. Results for all methods except OrbitGrasp, SE(2)-EquiGrasp, EquiGIGA, and EquiIGD are reproduced from \citep{zurbrugg2024icgnet} using its software and random seeds. Latency is measured from a depth map to predicted grasps. The best values are shown in blue boldface. Second-best values are shown in pink underlined text only when the best value is unique; tied best values receive the best-value style, and no second-best value is highlighted in that column.}
  \resizebox{0.8\linewidth}{!}{
  \setlength{\tabcolsep}{3mm}{}
  \begin{tabular}{c|cc|cc|c}
    \toprule
    \multirow{2}*{Method}   & \multicolumn{2}{c}{Packed}  & \multicolumn{2}{c}{Pile} & \multirow{2}*{Latency (ms)}  \\
    & GSR (\%) & DR (\%) & GSR (\%) & DR (\%)  \\
    \midrule
    PointNetGPD \citep{liang2019pointnetgpd} & 79.3$\pm$1.8 &  82.5$\pm$2.9  & 75.6$\pm$2.3  &  77.0$\pm$2.8 & 149 \\
    VGN \citep{vgn} & 80.2$\pm$1.6 &  86.2$\pm$2.0  & 64.9$\pm$2.2  &  69.1$\pm$3.2 & \boldcolor{First}{9} \\
    GIGA \citep{giga} & 89.9$\pm$1.7 &  87.6$\pm$2.0  & 76.3$\pm$2.4  &  80.9$\pm$4.1  & 24\\
    GIGA (HR) \citep{giga} & 91.4$\pm$1.5 &  88.5$\pm$1.4  & 86.5$\pm$1.2  & 80.8$\pm$1.9 & 53 \\
    EdgeGraspNet \citep{huang2023edge} & 92.5$\pm$0.9  & 94.3$\pm$1.1  &  91.5$\pm$1.3 &  92.5$\pm$1.3 &  843/685   \\
    VN-EdgeGraspNet \citep{huang2023edge} & 91.6$\pm$1.7   & 94.4$\pm$1.5  & \boldcolor{First}{92.0$\pm$1.8}  & 92.2$\pm$2.1 & 1174/953 \\
    ICGNet \citep{zurbrugg2024icgnet} & 97.7$\pm$0.9 &  \boldcolor{Second}{97.5$\pm$0.3} & \boldcolor{First}{92.0$\pm$2.6} & \boldcolor{Second}{94.1$\pm$1.4} &  806 \\
    OrbitGrasp \citep{hu2024orbitgrasp} & \boldcolor{Second}{98.1$\pm$0.7} &  \boldcolor{First}{98.2$\pm$0.8} & 91.6$\pm$1.7 & 93.2$\pm$1.3 & 3193 \\
    SE(2)-EquiGrasp \citep{zhu2023robot} &  \boldcolor{First}{98.6$\pm$0.2}  &  95.4$\pm$0.4 & 87.3$\pm$2.2  & 87.3$\pm$1.8 & \boldcolor{Second}{11} \\
    \midrule
    EquiGIGA & 95.4$\pm$0.7 & 97.1$\pm$1.2 & 91.3$\pm$0.6 & 90.5$\pm$1.4 & 25\\
    EquiIGD ($N$=1) & 96.0$\pm$1.5  & 97.1$\pm$0.8  & 91.4$\pm$0.8 &  \boldcolor{First}{94.2$\pm$0.6} & 52  \\
    \bottomrule
  \end{tabular}
  }
  \label{tab:sota_gaussian}
\end{table*}

\begin{table*}[t]
  \centering
  \caption{Real-world clutter-removal performance. Each cell reports the observed percentage and its two-sided 95\% Wilson confidence interval in brackets. GSR is followed by the numbers of successful grasps and total trials in parentheses; DR is followed by the number of removed objects out of 75 object instances. $N$ is the number of sampling rounds used by IGD and EquiIGD.
  }
  \resizebox{\linewidth}{!}{
  \setlength{\tabcolsep}{3mm}{}
  \begin{tabular}{c|cc|cc|cc}
    \toprule
    \multirow{2}*{Method}   & \multicolumn{2}{c}{Packed}  & \multicolumn{2}{c}{Pile} & \multicolumn{2}{c}{Adv}  \\
    & GSR (\%, 95\% CI) & DR (\%, 95\% CI) & GSR (\%, 95\% CI) & DR (\%, 95\% CI) & GSR (\%, 95\% CI) & DR (\%, 95\% CI) \\
    \midrule
    GIGA \citep{giga} & \realgsr{76.7}{66/86}{66.8}{84.4} & \realdr{88.0}{66}{78.7}{93.6} & \realgsr{61.1}{44/72}{49.6}{71.5} & \realdr{58.7}{44}{47.4}{69.1} & \realgsr{72.5}{66/91}{62.6}{80.6} & \realdr{88.0}{66}{78.7}{93.6} \\
    EdgeGraspNet \citep{huang2023edge} & \realgsr{73.4}{58/79}{62.8}{81.9} & \realdr{77.3}{58}{66.7}{85.3} & \realgsr{62.1}{41/66}{50.1}{72.9} & \realdr{54.7}{41}{43.4}{65.4} & \realgsr{72.2}{57/79}{61.4}{80.8} & \realdr{76.0}{57}{65.2}{84.2} \\
    VN-EdgeGraspNet \citep{huang2023edge} & \realgsr{71.3}{57/80}{60.5}{80.0} & \realdr{76.0}{57}{65.2}{84.2} & \realgsr{67.7}{44/65}{55.6}{77.8} & \realdr{58.7}{44}{47.4}{69.1} & \realgsr{79.5}{58/73}{68.8}{87.1} & \realdr{77.3}{58}{66.7}{85.3} \\
    IGD ($N$=1) \citep{igd} & \realgsr{78.0}{64/82}{67.9}{85.6} & \realdr{85.3}{64}{75.6}{91.6} & \realgsr{58.0}{51/88}{47.5}{67.7} & \realdr{68.0}{51}{56.8}{77.5} & \realgsr{78.2}{61/78}{67.8}{85.9} & \realdr{81.3}{61}{71.1}{88.5} \\
    ICGNet \citep{zurbrugg2024icgnet} & \realgsr{72.2}{57/79}{61.4}{80.8} & \realdr{76.0}{57}{65.2}{84.2} & \realgsr{71.1}{54/76}{60.0}{80.0} & \realdr{72.0}{54}{61.0}{80.9} & \realgsr{69.9}{51/73}{58.6}{79.2} & \realdr{68.0}{51}{56.8}{77.5} \\
    \midrule
    EquiGIGA & \realgsr{88.5}{69/78}{79.5}{93.8} & \realdr{92.0}{69}{83.6}{96.3} & \realgsr{85.3}{64/75}{75.6}{91.6} & \realdr{85.3}{64}{75.6}{91.6} & \realgsr{88.1}{74/84}{79.5}{93.4} & \realdr{98.7}{74}{92.8}{99.8} \\
    EquiIGD ($N$=1) & \realgsr{90.8}{69/76}{82.2}{95.5} & \realdr{92.0}{69}{83.6}{96.3} & \realgsr{82.5}{66/80}{72.7}{89.3} & \realdr{88.0}{66}{78.7}{93.6} & \realgsr{87.2}{75/86}{78.5}{92.7} & \realdr{100.0}{75}{95.1}{100.0} \\
    \bottomrule
  \end{tabular}
  }
  \label{tab:real}
\end{table*}

\begin{figure*}[t]
    \centering
    \includegraphics[width=\linewidth]{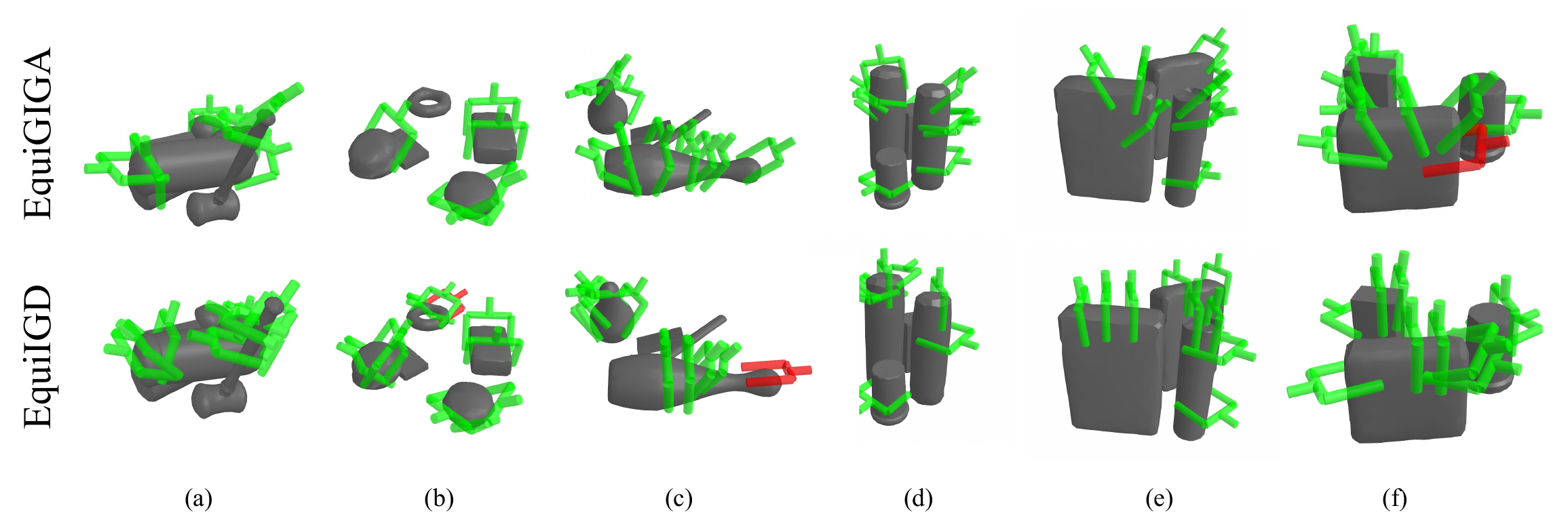}
    \caption{Visualization of top-10 grasps in challenging scenes. Grasps whose quality is below a fixed threshold $T$ are filtered out. The first row corresponds to EquiGIGA, and the second row corresponds to EquiIGD. Panels (a--c) show pile scenes, and panels (d--f) show packed scenes. Green grippers denote successful grasps, while red grippers denote failed grasps.}
    \label{fig:success cases}
\end{figure*}

\begin{figure*}[t]
    \centering
    \includegraphics[width=\linewidth]{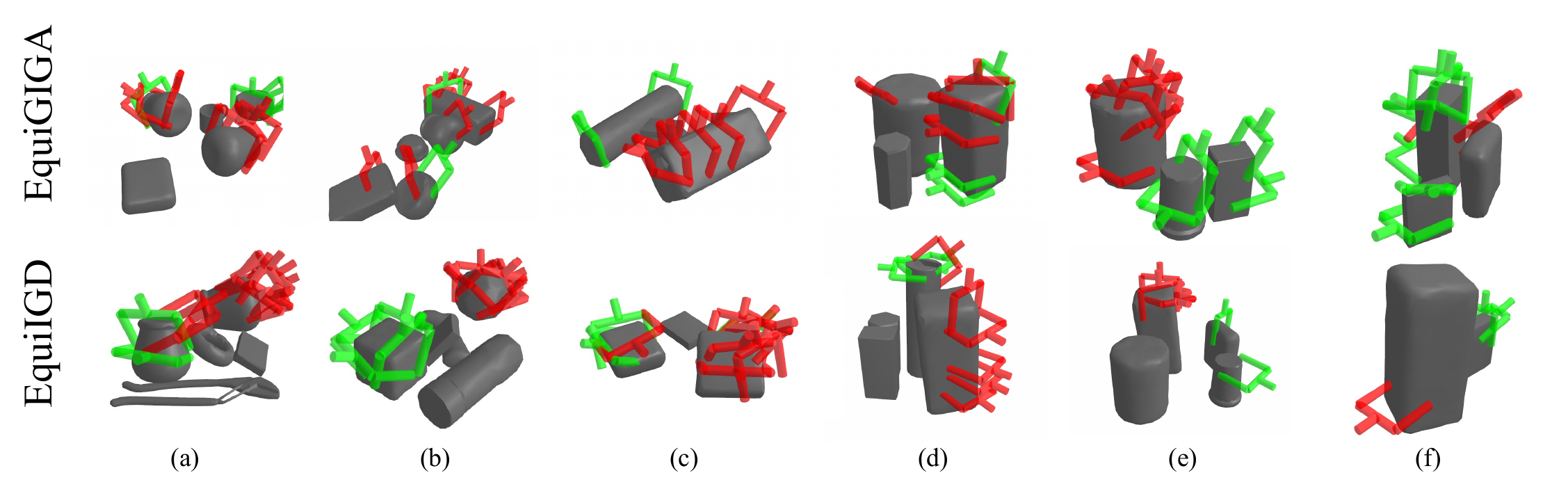}
    \caption{Different failure cases of EquiGIGA and EquiIGD. The first row corresponds to EquiGIGA, and the second row corresponds to EquiIGD. Panels (a--c) show pile scenes, and panels (d--f) show packed scenes. Green grippers denote successful grasps, while red grippers denote failed grasps.}
    \label{fig:failure cases}
\end{figure*}

\begin{table*}[!t]
  \centering
  \caption{Ablation of the Equivariant Tri-plane UNet (Eq.\ Tri-UNet) in the grasp-regression setting. We report mean $\pm$ standard deviation for GSR and DR. GFLOPs are measured for the backbone only with a $40 \times 40 \times 40$ input; training GPU memory uses a batch size of 64; and inference latency is measured on an RTX 4070 GPU.}
  \resizebox{\linewidth}{!}{
  \setlength{\tabcolsep}{3mm}{}
  \begin{tabular}{c|l|cccccc}
    \toprule
    No. & Ablated models &  GSR (\%) & DR (\%) & GPU Memory (MB) & GFLOPs & Params (M) & Latency (ms)\\
    \midrule
     1 & 3D UNet & 86.8$\pm$1.1 & 85.5$\pm$1.5 & 17226 & 44.30 & 2.24  & 22 \\
     2 & Eq.\ 3D UNet & 89.9$\pm$1.5 & 88.8$\pm$2.0 & 17208 & 44.30  & 0.22 & 22 \\
     3 & Tri-UNet  & 84.6$\pm$1.1   & 85.3$\pm$2.4  & 3485 & 1.90 & 0.84 & 24 \\
     4 & Tri-UNet ($XY$-separated) & 87.0$\pm$1.2 & 84.8$\pm$1.6 & 3292 & 1.90 & 1.61 & 24 \\
     5 & Eq.\ Tri-UNet (no {V2TP}) & 90.3$\pm$0.6 & 87.4$\pm$1.7 & 3860 & 1.90  & 0.40 & 24 \\
     6 & Tri-UNet (with {V2TP}) & 90.9$\pm$1.4 & 86.6$\pm$1.3 &3405 & 1.95 & 1.69  & 25 \\
     7 & Eq.\ Tri-UNet (with {V2TP}) & 93.7$\pm$1.0 & 88.6$\pm$1.7 &  3876 & 1.95 & 0.42 & 25 \\
    \bottomrule
  \end{tabular}
  }
  \label{tab:ablation equi triunet}
\end{table*}

\subsection{Grasp Detection Results}
We first test all models with a single fixed side-view camera. Tab.~\ref{tab:sota} reports GSR and DR in packed and pile scenes. In packed scenes, EquiIGD achieves the highest reported GSR and DR (98.0\% and 92.0\%), while EquiGIGA achieves the second-highest values (94.5\% and 91.4\%). In pile scenes, EquiIGD achieves the highest reported GSR (79.5\%), whereas EquiGIGA achieves the highest reported DR (77.3\%). Both methods improve on GIGA (69.5\% GSR and 49.0\% DR) and IGD (71.8\% GSR and 55.7\% DR), respectively. When GIGA does not identify a feasible grasp, the decluttering round terminates even when objects remain; such early termination can reduce DR without adding a failed attempt to the GSR denominator. EquiGIGA improves grasp detection and predicts more accurate orientation, resulting in a large improvement, especially on DR.
We note that IGD uses diffusion, whereas EquiIGD uses flow matching; the EquiIGD ablations in the Appendix separate the contributions of flow matching, equivariance, and the other proposed components.


Tab.~\ref{tab:sota} lists latency for each model. Latency is the time necessary to compute grasps from a depth map. Notably, for EdgeGraspNet and VN-EdgeGraspNet, most of the latency is due to scene preprocessing, and there is a significant latency difference between packed and pile scenes. Thus, we report the latency for packed/pile separately. EquiGIGA requires 25 ms per scene and EquiIGD 52 ms per scene. Although both are slower than VGN (9 ms), SE(2)-EquiGrasp (11 ms) and GIGA (24 ms), they are faster than GraspNet-1B Baselines (77 ms) \citep{fang2020graspnet}, GSNet (156 ms) \citep{wang2021graspness}, EdgeGraspNet (843 ms in packed scenes and 685 ms in pile scenes) \citep{huang2023edge}, VN-EdgeGraspNet (1174 ms in packed scenes and 953 ms in pile scenes) \citep{huang2023edge}, ICGNet (806 ms) \citep{zurbrugg2024icgnet}, and DexGraspNet2 (2781 ms) \citep{zhang2024dexgraspnet}. Due to the 10-step flow matching in EquiIGD, EquiIGD has lower latency compared to IGD, which uses 100-step DDPM.

We also observe a notable performance discrepancy between the results in Tab.~\ref{tab:sota} and those reported in prior works, including OrbitGrasp \citep{hu2024orbitgrasp} and ICGNet \citep{zurbrugg2024icgnet}. This difference primarily arises from variations in camera viewpoints and the type of noise added to the depth data. The experimental setup in Tab.~\ref{tab:sota} follows that of GIGA \citep{giga} and IGD \citep{igd}, where gamma noise is applied to the depth map. In contrast, methods like EdgeGraspNet \citep{huang2023edge} and ICGNet \citep{zurbrugg2024icgnet} use Gaussian noise ($\mathcal{N}(0, 0.0008)$), either on the depth map or the point cloud. Under gamma noise, surface normal estimation via PCA becomes less reliable, which degrades the performance of contact-based grasping models that rely heavily on accurate normal input. While gamma noise more closely mimics the output of low-cost, noisy depth sensors such as the RealSense D435i, Gaussian noise better approximates cleaner, near-noise-free sensors.

To validate robustness under both noise regimes, we conduct additional experiments using Gaussian noise. Tab.~\ref{tab:sota_gaussian} shows an accuracy--latency trade-off rather than uniform accuracy superiority. In packed scenes, EquiGIGA and EquiIGD achieve 95.4\%/97.1\% and 96.0\%/97.1\% GSR/DR, respectively; these values are below the highest packed-scene GSR of 98.6\% from SE(2)-EquiGrasp and the highest packed-scene DR of 98.2\% from OrbitGrasp. In pile scenes, EquiIGD achieves the highest reported DR (94.2\%), whereas the highest reported GSR is 92.0\% for both VN-EdgeGraspNet and ICGNet, compared with 91.3\% and 91.4\% for EquiGIGA and EquiIGD. The proposed methods instead provide substantially lower latency than the high-accuracy point-cloud alternatives: EquiGIGA and EquiIGD require 25~ms and 52~ms per scene, respectively, compared with 806~ms for ICGNet and 3193~ms for OrbitGrasp.

Tab.~\ref{tab:real} presents the real-world results, which are consistent with the fixed-view gamma-noise simulation results in Tab.~\ref{tab:sota}. We also show the two-sided 95\% Wilson confidence intervals for GSR and DR. In packed scenes, EquiIGD records 90.8\% GSR (95\% CI: [82.2, 95.5]), while both EquiGIGA and EquiIGD record 92.0\% DR (95\% CI: [83.6, 96.3]). In pile scenes, EquiGIGA records 85.3\% GSR (95\% CI: [75.6, 91.6]) and EquiIGD records 88.0\% DR (95\% CI: [78.7, 93.6]). In adversarial scenes, EquiGIGA records 88.1\% GSR (95\% CI: [79.5, 93.4]) and EquiIGD records 100.0\% DR (95\% CI: [95.1, 100.0]). Notably, we observe a sim-to-real gap when deploying contact-based methods such as EdgeGraspNet \citep{huang2023edge} and ICGNet \citep{zurbrugg2024icgnet}, primarily because the RealSense D435i camera used in our setup introduces more noise than the high-quality sensors used in their experiments. On noisy depth maps, the estimated surface normals become unreliable, leading to substantial performance degradation.
In contrast, volumetric grasping models such as GIGA, IGD, EquiGIGA, and EquiIGD use low-resolution TSDF volumes as input and do not rely on surface normals, making them inherently more robust to noise. As a result, these models exhibit smaller sim-to-real gaps and maintain high performance when transferred to real-world scenarios.

\subsection{Visualization of Grasp Detection} \label{sec. viz}
We visualize the top-10-scoring grasps in challenging cases in Fig.~\ref{fig:success cases}. Non-maximum suppression is applied to filter out similar grasps for better visualization. The first row denotes the grasp results of EquiGIGA, and the second row denotes the grasp results of EquiIGD. Fig.~\ref{fig:success cases}~(a--c) are pile scenes, while (d--f) illustrate packed scenes. Green grippers denote successful grasps, while red grippers denote failed grasps. As shown in Fig.~\ref{fig:success cases}, both EquiGIGA and EquiIGD generate accurate collision-free grasps.

We visualize failure cases of EquiGIGA and EquiIGD in Fig.~\ref{fig:failure cases}. Fig.~\ref{fig:failure cases}~(a--c) are pile scenes, while Fig.~\ref{fig:failure cases}~(d--f) illustrate packed scenes. Two main reasons explain the failures. (i) Volumetric grasping models have difficulty grasping objects with smooth curved surfaces (Fig.~\ref{fig:failure cases}~(a--b, d--e) of EquiGIGA and Fig.~\ref{fig:failure cases}~(a--b) of EquiIGD) or large objects (Fig.~\ref{fig:failure cases}~(c--d) of EquiGIGA and Fig.~\ref{fig:failure cases}~(c--f) of EquiIGD). To achieve valid grasps, the contact normals of the two fingers must closely align with the object's surface normals. Since a volumetric grasping model uses low-resolution volumetric data as input, capturing fine-grained surface normals on a curved surface is difficult. In contrast, contact-based models (e.g., EdgeGraspNet) with explicit surface normals as input are better positioned for such objects. (ii) Volumetric grasping models struggle to distinguish between different objects without RGB information. As shown in Fig.~\ref{fig:failure cases}~(f) of EquiGIGA, the two columns are not connected, but EquiGIGA, with a low-resolution input, regards them as connected and predicts a grasp in between them.

\begin{figure*}[t]
    \centering
    \includegraphics[width=\linewidth]{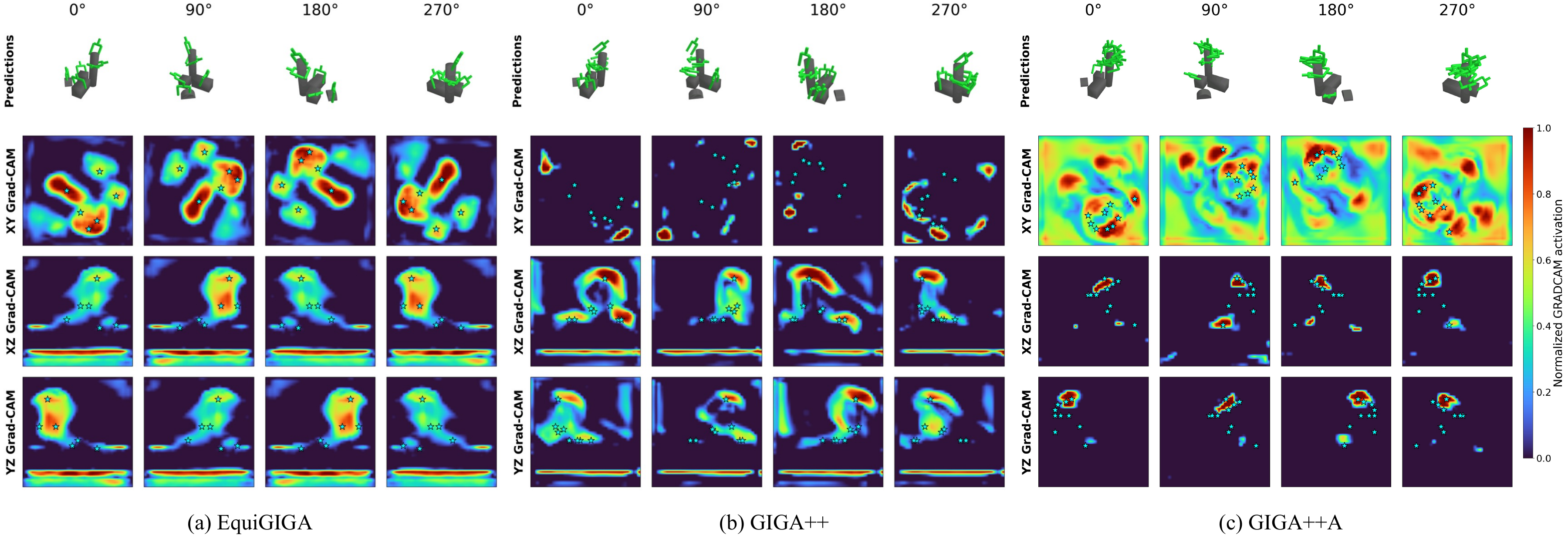}
    \caption{Qualitative comparison of EquiGIGA, GIGA++, and GIGA++A under $0^\circ$, $90^\circ$, $180^\circ$, and $270^\circ$ planar rotations of the same input scene. We select the top ten query locations whose graspness scores exceed the threshold $T$ at $0^\circ$ and rotate these same locations with the input for the other columns. The top row shows predicted grasps; the remaining rows show normalized Grad-CAM activations for the graspness predictions in the $XY$, $XZ$, and $YZ$ feature planes. Star markers indicate the projections of the selected 3D query locations onto the respective feature planes. Compared to GIGA++ and GIGA++A, EquiGIGA produces more consistently transformed grasp predictions and activation patterns.}
    \label{fig:equi cases}
\end{figure*}

Finally, Fig.~\ref{fig:equi cases} compares EquiGIGA with the non-equivariant GIGA++ and GIGA++A baselines defined in Sec.~\ref{sec:training_inference} under $0^\circ$, $90^\circ$, $180^\circ$, and $270^\circ$ rotations of the volumetric input. To construct the visualization, we first select the ten query locations with the highest predicted graspness in the $0^\circ$ input and compute Grad-CAM maps for their graspness predictions. For each rotated input, we use the corresponding ten locations after applying the same rotation, rather than selecting a new set of high-graspness locations. Each column therefore shows predictions and Grad-CAM activation maps for the same physical query locations across the $XY$, $XZ$, and $YZ$ feature planes. The star markers show the projections of these 3D query locations onto each feature plane. Across rotations, EquiGIGA produces more consistently transformed grasp predictions and activation patterns than GIGA++ and GIGA++A, illustrating the effect of the equivariant design. We restrict this visualization to the deterministic GIGA-family models because EquiIGD relies on stochastic orientation sampling and does not permit a directly comparable visualization.


\begin{figure*} [!t]
\centering
    \resizebox{\linewidth}{!}{
    \includegraphics[width=\linewidth]{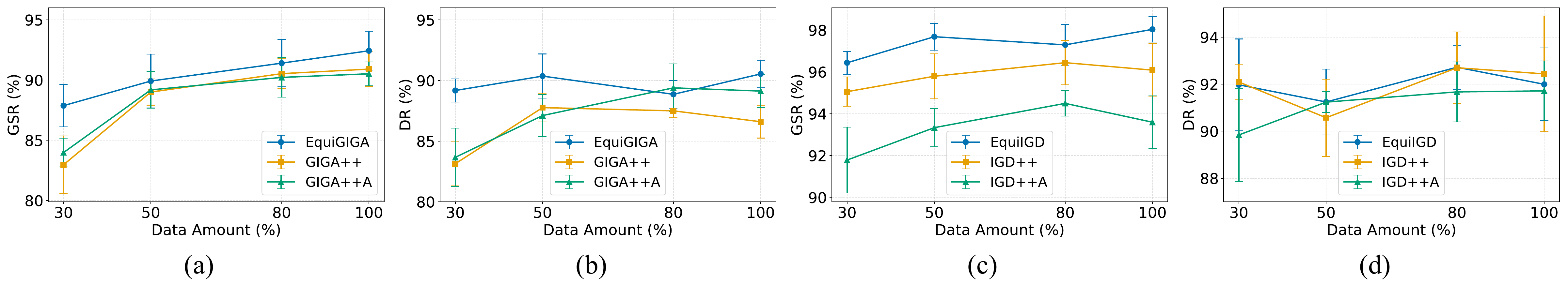}
    }
    \caption{Sample efficiency experiments on (a--b) EquiGIGA and (c--d) EquiIGD. We train the models with reduced fractions of the available training data and evaluate their performance. Results are averaged across five random seeds, and error bars indicate one standard deviation above and below the mean. For each given training set size, equivariant models (blue) generally outperform non-equivariant ones (green and orange). See text for detailed discussion.}
    \vspace{-0.3cm}
\label{fig:sample_effi}
\end{figure*}

\begin{figure*}[t]
    \centering
    \includegraphics[width=\linewidth]{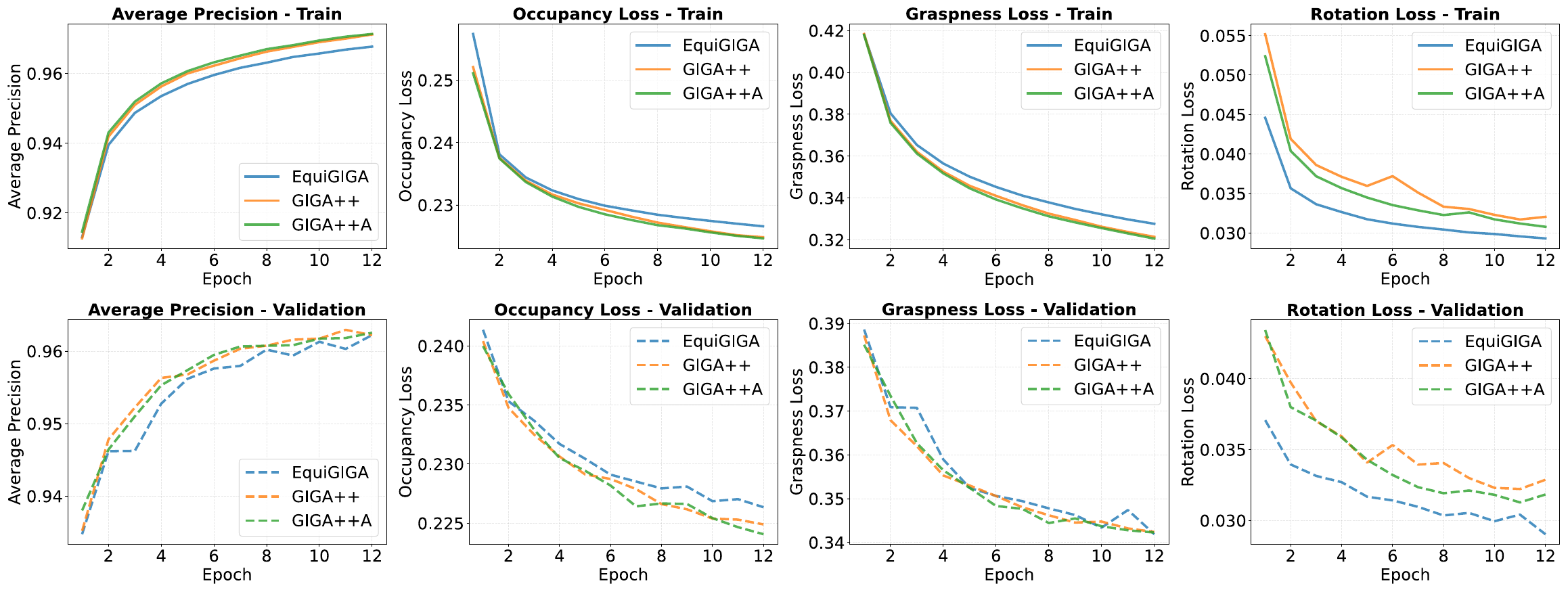}
    \caption{We visualize the curves of four metrics during training in both the training and validation sets: average precision, occupancy loss, graspness loss, and rotation loss. The superior performance of EquiGIGA (blue) compared to GIGA++ and GIGA++A (green and orange) mainly stems from the lower rotation loss, which benefits from the model's encoded equivariance.}
    \label{fig:equigiga_loss}
\end{figure*}

\subsection{Ablation Studies} \label{sec:ablation}
All the ablation studies are conducted with a subset of the packed scene dataset. We follow the experimental settings used for Tab.~\ref{tab:sota}: a fixed camera, and gamma noise added to the simulated depth map. To compare equivariant models with non-equivariant ones, we use ESCNN \citep{cesa2022program} to export steerable CNNs to conventional CNNs, which guarantees that the GFLOPs are the same.

Tab.~\ref{tab:ablation equi triunet} reports the ablation of the proposed Equivariant Tri-plane UNet under the grasp-regression setting used by GIGA and EquiGIGA. Rows 1 and 2 compare conventional and $\textnormal{C}_4$ $Z$-axis rotation-equivariant 3D UNets. The equivariant 3D UNet attains higher GSR and DR (89.9\% and 88.8\%) than the conventional 3D UNet (86.8\% and 85.5\%), which demonstrates the effectiveness of enforcing equivariance. Row 3 is the Tri-plane UNet used by GIGA. Compared with the conventional 3D UNet, it substantially reduces training memory (3,485 MB versus 17,226 MB) and computation (1.90 versus 44.30 GFLOPs), at the cost of a modest reduction in GSR and DR. Its latency is slightly higher, which we attribute to more efficient parallelization in the 3D UNet implementation.

Rows 4--7 isolate the effects of the proposed components. Row 4 (\emph{$XY$-separated Tri-UNet}) uses a dual-branch UNet architecture (one for the $XY$ plane and one for two vertical planes) built with conventional convolutions and learns the remaining symmetries from data. Adding architectural equivariance without {V2TP} (row 5) increases GSR from 87.0\% to 90.3\%. Adding {V2TP} to the non-equivariant tri-plane model (row 6, GIGA++) increases GSR from 87.0\% to 90.9\%. Combining both components (row 7) yields the highest GSR among all ablated models (93.7\%) and the highest DR among the tri-plane variants (88.6\%), while retaining the compact tri-plane backbone. Although the equivariant 3D UNet in row 2 attains a marginally higher DR (88.8\%), it requires substantially greater memory, computation, and parameter count. These results support the complementary value of architectural equivariance and explicit information exchange between the three planes.

\begin{figure*}[t!]
    \centering
    \includegraphics[width=\linewidth]{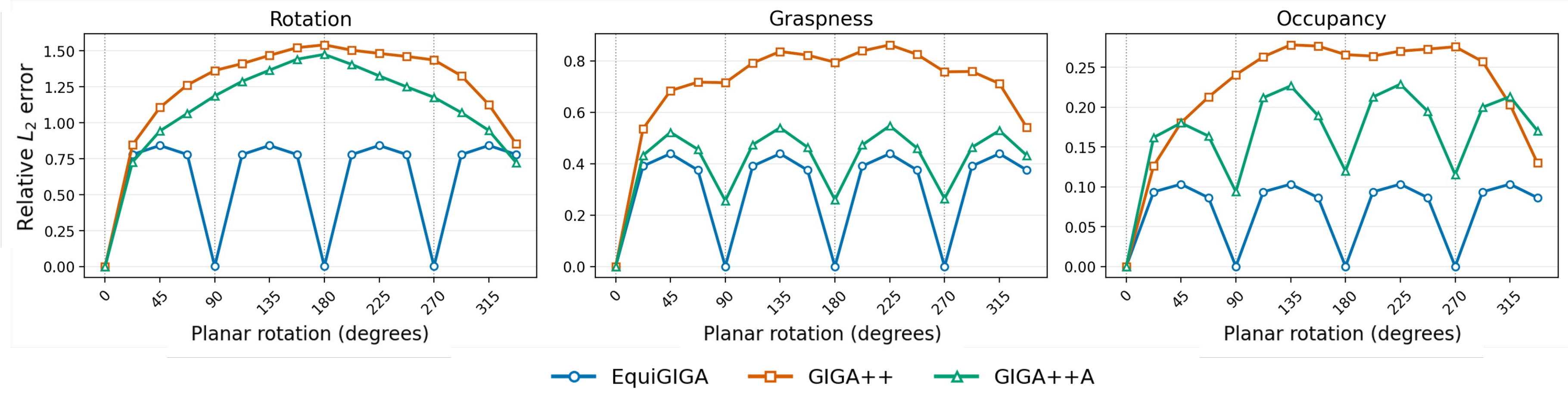}
    \caption{Empirical equivariance consistency error under planar input rotations, measured as the relative $\ell_2$ discrepancy between predictions from the rotated input and appropriately transformed predictions from the original input. EquiGIGA attains errors at floating-point precision for the non-identity $\textnormal{C}_4$ rotations in all three heads. GIGA++A exhibits recurring error minima at these rotations for graspness and occupancy, but no corresponding minima for rotation prediction. Rotations outside $\textnormal{C}_4$ involve TSDF resampling and therefore also introduce interpolation effects.}
    \label{fig:equivariance_error}
\end{figure*}

\begin{figure*}[t!]
    \centering
    \includegraphics[width=\linewidth]{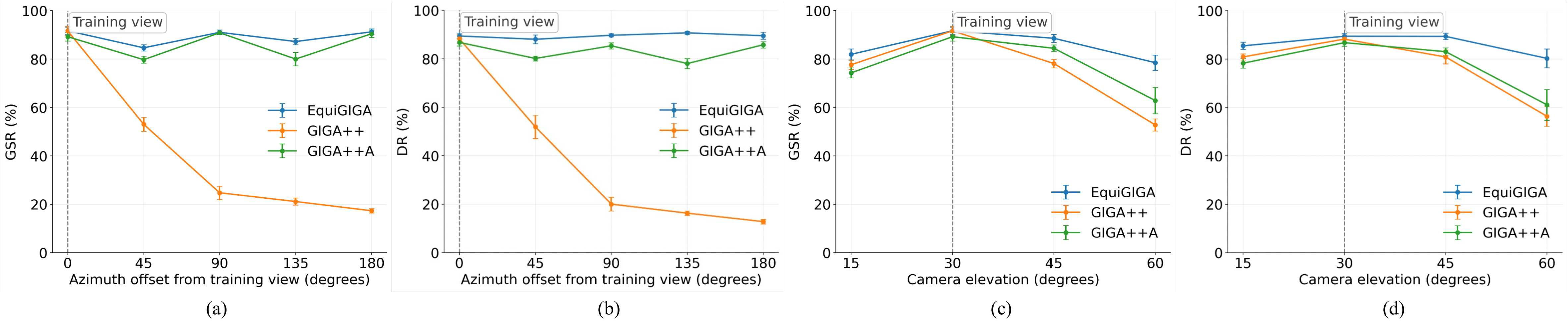}
    \caption{Robustness to camera viewpoint changes for models trained with a fixed acquisition viewpoint. The camera is moved while the tabletop coordinate frame and initial object arrangement are held fixed, and a new depth observation is integrated into the TSDF. (a,b) GSR and DR under camera azimuth offsets of $0^\circ$, $45^\circ$, $90^\circ$, $135^\circ$, and $180^\circ$, at a fixed elevation of $30^\circ$. (c,d) GSR and DR at camera elevations of $15^\circ$, $30^\circ$, $45^\circ$, and $60^\circ$, with the training azimuth fixed. Dashed lines indicate the training viewpoint; points and error bars denote the mean and standard deviation across five random seeds. EquiGIGA achieves the highest mean GSR and DR at every tested viewpoint, although its GSR decreases at intermediate azimuth offsets and both metrics decrease at an elevation of $60^\circ$. GIGA++A mitigates the azimuth sensitivity of GIGA++, with GSR recovering at offsets of $90^\circ$ and $180^\circ$.}
    \label{fig:cameraview_robustness}
\end{figure*}

\subsection{Sample Efficiency}
Results shown in Fig.~\ref{fig:sample_effi} support our claim that equivariant models improve sample efficiency. In this experiment, we compare EquiGIGA with GIGA++ and GIGA++A under different amounts of training data. EquiGIGA shows the strongest overall GSR and DR trends across the evaluated training-data fractions, although GIGA++A attains a slightly higher DR at the 80\% data fraction. We also compare EquiIGD to IGD++, a variant that does not use steerable filters, and IGD++ with rotation data augmentation (IGD++A). It should be noted that, compared to IGD, IGD++ utilizes an $XY$-separated Tri-plane UNet backbone, {V2TP}, and flow matching. We observe a smaller performance gap between EquiIGD and IGD++ than between EquiGIGA and GIGA++. We hypothesize that the smaller performance gap may be partly attributable to the partial equivariance provided by GraspDAM in IGD++ \citep{igd}. EquiIGD shows the strongest overall performance compared to its variants, while IGD++A is generally weaker, particularly in GSR.

This degradation does not imply that $\textnormal{C}_4$ rotations are intrinsically harmful. The augmentations add no new scene geometry; instead, they require IGD++, which has no architectural rotation equivariance, to learn four orientation-specific versions of the same input--output relationship. Under a fixed model capacity and training budget, these redundant views compete for capacity rather than sharing it. This is particularly challenging for IGD++ because its flow-matching head models a multimodal conditional distribution of grasp rotations: the model must learn the corresponding modes independently for each rotated view. Consequently, rotational augmentation can increase optimization difficulty and reduce generalization, even though it increases the number of training samples. EquiIGD shares these mappings through its equivariant design, so rotated observations reinforce rather than duplicate the learned grasp-rotation relationship.



To further validate how the equivariance constraint enhances grasping performance, we analyze learning curves across four key metrics---average precision (AP), occupancy loss, graspness loss, and rotation loss---on both the training and validation sets (Fig.~\ref{fig:equigiga_loss}). We observe a critical trade-off favoring generalization: while EquiGIGA exhibits a higher graspness loss and lower AP on the training set, it either matches or marginally surpasses its non-equivariant counterparts in validation AP. This suggests that the non-equivariant models suffer from slight overfitting despite the use of data augmentation. The most substantial benefit of the equivariant approach is evident in the rotation loss, where EquiGIGA outperforms the non-equivariant models on both the training and validation data. This reduction in rotation error is directly attributable to leveraging the inherent rotational symmetries of the task, confirming that grasp rotation prediction highly respects the equivariance constraint. Conversely, EquiGIGA underperforms its non-equivariant counterparts in occupancy prediction. This inferior performance might be caused by the non-symmetric nature of the task, where all models reconstruct the whole scene from a partial observation. Enforcing equivariance might limit models' ability to imagine the unobserved part. As an auxiliary task, however, this does not directly impede the grasping performance. The clear superiority compared to GIGA++ and GIGA++A in rotation prediction is consistent with, and likely contributes to, EquiGIGA's improved performance in Tabs.~\ref{tab:sota} and~\ref{tab:sota_gaussian}.

The results above align with empirical findings in other equivariant neural networks \citep{bekkers2018roto,weiler2024equivariant}: while vanilla CNNs combined with data augmentation can, in principle, learn approximate invariance or equivariance to geometric transformations such as rotations or reflections, they often perform less well than steerable CNNs in our experiments and prior empirical studies. Two factors may contribute to this difference. First, the equivariance learned from the data is unstructured, which means that vanilla CNNs may memorize input-output pairs instead of learning their underlying relationship. In contrast, steerable CNNs incorporate symmetry constraints directly into the model architecture. By mathematically encoding the relationship between input and output transformations, they guarantee consistent feature behavior under group actions without the need for exhaustive augmentation. Second, data augmentation can only be applied globally---transforming the whole input and all outputs with the same transformation. The diversity increase from data augmentation is limited by the original data. In comparison, steerable CNNs detect local patterns and output equivariantly when local patterns transform, providing local feature equivariance. This local feature equivariance may improve generalization to new scenes.

\subsection{Empirical Verification of Output Equivariance Consistency} \label{sec:output_equivariance}

We empirically evaluate whether the predicted outputs transform consistently under planar rotations. Given a TSDF volume $\mathcal{V}$, query points $\mathbf{p}$, a planar rotation $g$, and a volumetric grasp model $F$, we compare $F(g \cdot \mathcal{V}, g \cdot \mathbf{p})$ with $\rho(g)F(\mathcal{V},\mathbf{p})$ using the relative $\ell_2$ error, separately for the rotation, graspness, and occupancy heads. For graspness and occupancy, $\rho(g)$ is the identity: the scalar predictions at corresponding locations should remain unchanged. For the rotation head, $\rho(g)$ applies the same planar rotation to the predicted grasp orientation. This metric measures consistency under input transformations, rather than prediction accuracy relative to ground truth.

We generate 20 independently randomized packed scenes, each containing five objects, and acquire one TSDF observation per scene. The TSDF resolution is $40^3$. For each scene, we uniformly sample 4,096 query locations within the workspace, resulting in 81,920 query locations per model at each tested rotation. We evaluate the $\textnormal{C}_{16}$ test group, i.e., 16 rotations about the vertical axis with an angular spacing of $22.5^\circ$. For every rotation, we rotate both the TSDF and its corresponding query locations, evaluate the output heads, and aggregate the relative $\ell_2$ error over all scenes and query points.

Fig.~\ref{fig:equivariance_error} compares EquiGIGA, the non-equivariant GIGA++, and GIGA++A, which is trained with $\textnormal{C}_4$ rotation augmentation. Rotations by multiples of $90^\circ$ are applied exactly by permuting the TSDF grid; other angles require trilinear TSDF resampling. EquiGIGA exhibits a repeating pattern with errors at floating-point precision for all three heads at $90^\circ$, $180^\circ$, and $270^\circ$, empirically confirming the intended $\textnormal{C}_4$ equivariance of the complete model. Its nonzero errors between these angles reflect the absence of an architectural guarantee for rotations outside $\textnormal{C}_4$, together with the effects of TSDF resampling. All models have zero error at $0^\circ$ by construction, so the identity rotation does not provide evidence of learned equivariance.

GIGA++ retains substantial error at the non-identity $\textnormal{C}_4$ rotations and shows no comparable recurring minima. GIGA++A, however, behaves differently across output heads. Its graspness and occupancy errors exhibit recurring local minima at $90^\circ$, $180^\circ$, and $270^\circ$, with higher errors between these angles. These approximately $90^\circ$-periodic reductions align with the augmentation group, suggesting that augmentation encourages approximate invariance of the scalar outputs. The residual errors at these minima show that this invariance remains incomplete. Although the minima also coincide with rotations that avoid TSDF interpolation, GIGA++ is evaluated with the same transformations without exhibiting the same pattern; thus, the absence of resampling alone does not explain the scalar-head behavior of GIGA++A.

The rotation head of GIGA++A does not exhibit these recurring reductions. Instead, its error broadly increases towards $180^\circ$ and then decreases, remaining high at the non-identity $\textnormal{C}_4$ rotations despite some improvement over GIGA++. This distinction is consistent with the different transformation requirements of the outputs. Scalar invariance requires the model to assign the same value to corresponding query locations across rotated inputs, whereas rotation equivariance requires the predicted orientation to change according to $\rho(g)$. Augmentation exposes the model to transformed training pairs, but does not explicitly enforce this relationship between its predictions. The scalar-head minima therefore cannot be taken as evidence that the orientation predictions have acquired a corresponding equivariant transformation rule.

Together, these results show that the effect of rotation augmentation is output-dependent: GIGA++A learns a clear signature of the augmentation group in its scalar predictions, while its orientation predictions lack the corresponding periodic consistency. EquiGIGA preserves the required transformation relationship for all three heads at the $\textnormal{C}_4$ rotations. This comparison supports the value of enforcing equivariance throughout both feature extraction and output decoding, particularly for grasp orientations, whose transformation behavior is not captured by scalar invariance alone.

\subsection{Robustness to Camera Viewpoint Changes} \label{sec:camera_viewpoint}

We next evaluate whether models trained with a fixed acquisition viewpoint remain effective when the camera is moved at test time. This experiment complements Sec.~\ref{sec:output_equivariance}: instead of rotating an existing TSDF, we acquire a new depth observation from each camera pose and reconstruct the TSDF in the same tabletop coordinate frame. The initial object arrangement is not transformed when changing the camera pose. Consequently, the input changes through visibility and occlusion as well as viewing direction, and this experiment measures grasping robustness rather than exact output equivariance.

We evaluate EquiGIGA, GIGA++, and GIGA++A in packed scenes without retraining or adapting their parameters to the test viewpoints. The reference camera uses the fixed side-view acquisition setup, with an elevation of $30^\circ$. We vary its azimuth relative to the training view over $\{0^\circ,45^\circ,90^\circ,135^\circ,180^\circ\}$ while keeping the elevation fixed. In a separate sweep, we keep the training azimuth fixed and vary the elevation over $\{15^\circ,30^\circ,45^\circ,60^\circ\}$. Each grasp attempt uses a single depth image with the gamma noise to construct a $40^3$ TSDF. For each model and viewpoint, we evaluate five random seeds, with a total of 1,712 objects across seeds. We use the same seed set across models and viewpoints. We compute GSR and DR separately for each seed and report their means and standard deviations across seeds in Fig.~\ref{fig:cameraview_robustness}.


The results of the azimuth sweep robustness experiment are shown in Fig.~\ref{fig:cameraview_robustness}~(a,b). GIGA++ deteriorates sharply as the azimuth offset increases. GIGA++A mitigates this sensitivity, recovering its training-view GSR at offsets aligned with the augmentation group but showing clear dips at intermediate offsets. EquiGIGA also exhibits intermediate-offset dips in GSR, while its DR remains close to the training-view level throughout the sweep. It achieves the highest mean GSR and DR at every tested offset and the smallest worst-case decrease in both metrics, indicating more reliable transfer to unseen azimuths.

The elevation sweep in Fig.~\ref{fig:cameraview_robustness}~(c,d) exposes a shared limitation: all three models perform best at the training elevation and degrade most at the highest tested elevation. EquiGIGA nevertheless achieves higher mean GSR and DR than both baselines, with smaller decreases from its training-view performance at every non-training elevation. Its advantage therefore extends beyond azimuth changes, although elevation remains a greater source of degradation in these experiments. Because elevation changes alter the visible top and side surfaces and cannot be represented by vertical-axis rotations, this advantage is an empirical generalization benefit rather than a guarantee of $\textnormal{C}_4$ equivariance.

The task-level recovery of GIGA++A is consistent with its augmentation structure, but does not imply exact output equivariance: different views can yield different successful grasps from different partial observations. It therefore does not contradict the rotation-head errors in Sec.~\ref{sec:output_equivariance}. Together, the two experiments distinguish output consistency under prescribed input rotations from robustness to newly acquired observations, and show that EquiGIGA supports both within the tested settings.


\section{Discussion}
In this paper, we introduce the Equivariant Tri-plane UNet, a volumetric backbone for robotic grasping. By combining this encoder with the state-of-the-art volumetric grasping models GIGA \citep{giga} and IGD \citep{igd}, we demonstrate substantial performance gains across both simulation and real-world settings. While these results validate the efficacy of equivariant representations, we identify two limitations that offer grounds for future investigation.

First, the proposed Equivariant Tri-plane UNet preserves equivariance only with respect to $\textnormal{C}_4$ $Z$-axis rotations. This choice is tailored to tabletop decluttering in a fixed tabletop coordinate frame, where vertical-axis rotations are the dominant source of pose variation. Accordingly, the formal guarantee does not extend to arbitrary $\textnormal{SO}(2)$ angles or general $\textnormal{SO}(3)$ transformations. The representation may therefore provide less benefit for objects that are not supported by a horizontal surface, scenes with substantial out-of-plane rotations or overturning, or manipulation tasks with an unconstrained object or camera coordinate frame. In such settings, the in-plane $XY$ action and the induced action on the $XZ$/$YZ$ pair no longer capture the principal geometric variation. While our experiments show this compromise to be valuable, preserving $\textnormal{O}(3)$ symmetry is likely to offer additional benefits, particularly in the early stages of feature extraction. In future work, we will consider hybrid approaches that apply equivariant convolutions in higher-dimensional groups (e.g., $\textnormal{O}(3)$) in early layers, and project the resulting features onto $\textnormal{O}(2)$ via restricted representations. Such extensions could accommodate settings in which the tabletop vertical-axis assumption no longer holds while retaining the efficiency and flexibility of the proposed representation.

Second, while contact normals provide a valuable geometric prior, volumetric grasping models typically fail to exploit this information. As discussed in Sec.~\ref{sec. viz}, this omission can lead to performance degradation when handling large or curved objects---scenarios where contact-based methods often excel \citep{huang2023edge, hu2024orbitgrasp, zurbrugg2024icgnet}. Currently, volumetric approaches offer an advantage in their resilience to noisy depth data, which often precludes reliable normal estimation; however, this edge may diminish as sensor quality improves. A promising future direction lies in unifying these paradigms to leverage both robust volumetric features and precise surface normals. This trend is already emerging in recent literature \citep{liu2024efficient}, and we anticipate that our work will further enhance the efficacy of such hybrid frameworks.

\section{Conclusion}

This paper introduces an equivariant volumetric grasping model that achieves strong sample efficiency by leveraging the rotational structure inherent in tabletop manipulation tasks. Rather than pursuing full \(\textnormal{SO}(3)\) equivariance at prohibitive computational cost, we demonstrate that restricting equivariance to \(\textnormal{C}_4\) vertical-axis rotations captures the dominant symmetry in tabletop scenes while remaining tractable in terms of memory and runtime.

The central technical contribution is a novel equivariant tri-plane representation in which horizontal-plane features transform equivariantly under \(\textnormal{C}_4\), while the paired vertical-plane features transform under the corresponding induced action. This structure arises naturally from the geometry of plane projections and is preserved throughout processing by a dual-branch UNet whose branches are respectively governed by $\textnormal{C}_4$-equivariant and reflection-equivariant operations. The {vertical-to-tableplane} feature augmentation further recovers 3D information that would otherwise be lost during projection, improving the expressiveness of the resulting feature fields without sacrificing the equivariance guarantees established analytically.

Building on these equivariant features, we derive equivariant adaptations of two state-of-the-art planners, GIGA and IGD, including an equivariant reformulation of deformable attention and a flow-matching-based generative model for grasp orientation. Both adaptations outperform their non-equivariant counterparts in simulation and real-world experiments, achieving higher success rates with a tractable memory and runtime cost compared to models relying solely on data-driven rotation generalization.


\section*{Acknowledgment}
This work is supported by Interne Fondsen KU Leuven/Internal Funds KU Leuven (C14/22/134) and funded by the European Union. Views and opinions expressed are however those of the author(s) only and do not necessarily reflect those of the European Union or HADEA. Neither the European Union nor the granting authority can be held responsible for them.

\begin{figure}[h]
    \centering
    \includegraphics[width=0.8\linewidth]{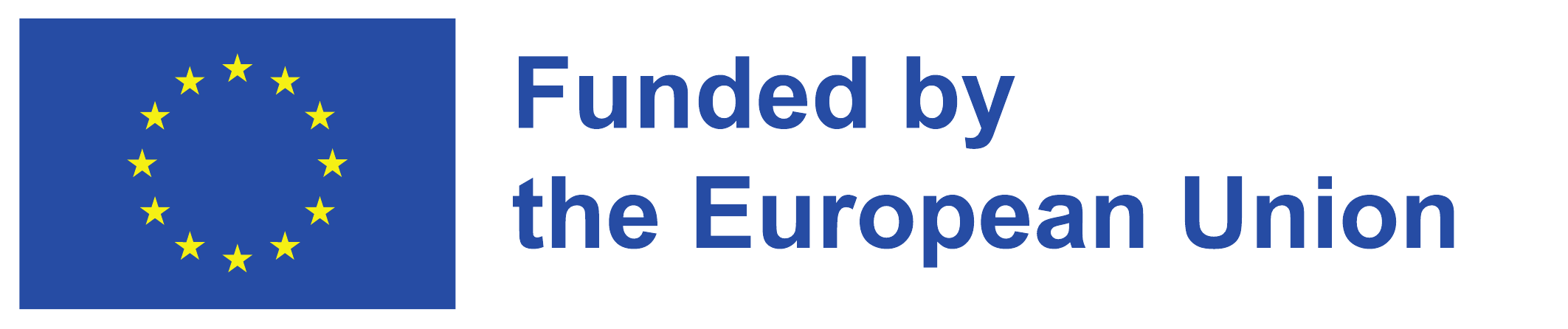}
\end{figure}

\section*{Appendix}

\appendix

\subsection*{Appendix 1: A Review of Steerable Convolution}
\label{appendix:sc}

Let $x:\mathbb{R}^2\rightarrow\mathbb{R}^{c_{\mathrm{in}}}$ denote an
input feature field and
$y:\mathbb{R}^2\rightarrow\mathbb{R}^{c_{\mathrm{out}}}$ the output
feature field. Let $G\subseteq\textnormal{SO}(2)$ be a linear rotation group acting on spatial coordinates, with input and output
representations $\rho_{\mathrm{in}}$ and $\rho_{\mathrm{out}}$,
respectively. Their feature-field actions are
\begin{equation}
    [g\triangleright_{\rho_{\mathrm{in}}}x](\mathbf p)
    =
    \rho_{\mathrm{in}}(g)x(g^{-1}\cdot\mathbf p),
\end{equation}
and
\begin{equation}
    [g\triangleright_{\rho_{\mathrm{out}}}y](\mathbf p)
    =
    \rho_{\mathrm{out}}(g)y(g^{-1}\cdot\mathbf p).
\end{equation}

Let $S\subset\mathbb{R}^2$ be the set of offsets in a discrete
convolutional kernel. We assume that $S$ is invariant under the action
of $G$:
\begin{equation}
    g^{-1}S=S,\qquad \forall g\in G.
    \label{eq:stencil-invariance}
\end{equation}
Equivalently, each $g\in G$ only permutes the offsets in $S$. For
example, a square convolutional stencil centered at the origin is
invariant under the discrete group $\textnormal{C}_4$, but not under
arbitrary rotations in $\textnormal{SO}(2)$.

Let
\[
w:\mathbb{R}^2\rightarrow
\mathbb{R}^{c_{\mathrm{out}}\times c_{\mathrm{in}}}
\]
be a steerable convolution kernel. It satisfies the kernel constraint
\begin{equation}
    w(g\cdot\mathbf p)
    =
    \rho_{\mathrm{out}}(g)
    w(\mathbf p)
    \rho_{\mathrm{in}}(g)^{-1},
    \qquad
    \forall g\in G,\;\mathbf p\in\mathbb{R}^2.
    \label{eq:steerable-kernel-constraint}
\end{equation}
The discrete convolution is defined by
\begin{equation}
    [w\star x](\mathbf p)
    =
    \sum_{\mathbf q\in S}
    w(\mathbf q)x(\mathbf p+\mathbf q).
    \label{eq:discrete-convolution}
\end{equation}

We now prove equivariance. By substituting
$\mathbf p\leftarrow g^{-1}\cdot\mathbf q$ into
Eq.~\ref{eq:steerable-kernel-constraint}, we obtain
\begin{equation}
    w(\mathbf q)\rho_{\mathrm{in}}(g)
    =
    \rho_{\mathrm{out}}(g)w(g^{-1}\cdot\mathbf q).
    \label{eq:kernel-rearrangement}
\end{equation}
Therefore,
\begin{equation}
\begin{aligned}
&[w\star(g\triangleright_{\rho_{\mathrm{in}}}x)](\mathbf p) \\
&=
\sum_{\mathbf q\in S}
w(\mathbf q)
\rho_{\mathrm{in}}(g)
x\bigl(g^{-1}\cdot(\mathbf p+\mathbf q)\bigr) \\
&=
\rho_{\mathrm{out}}(g)
\sum_{\mathbf q\in S}
w(g^{-1}\cdot\mathbf q)
x\bigl(g^{-1}\cdot\mathbf p+g^{-1}\cdot\mathbf q\bigr).
\end{aligned}
\label{eq:steerable-convolution-step-1}
\end{equation}

Let $\mathbf z=g^{-1}\cdot\mathbf q$. By the stencil-invariance
assumption in Eq.~\ref{eq:stencil-invariance}, this substitution is a
bijection from $S$ to itself. We can therefore reindex the sum:
\begin{equation}
\begin{aligned}
&\rho_{\mathrm{out}}(g)
\sum_{\mathbf q\in S}
w(g^{-1}\cdot\mathbf q)
x\bigl(g^{-1}\cdot\mathbf p+g^{-1}\cdot\mathbf q\bigr) \\
&=
\rho_{\mathrm{out}}(g)
\sum_{\mathbf z\in S}
w(\mathbf z)
x\bigl(g^{-1}\cdot\mathbf p+\mathbf z\bigr) \\
&=
\rho_{\mathrm{out}}(g)
[w\star x](g^{-1}\cdot\mathbf p) \\
&=
[g\triangleright_{\rho_{\mathrm{out}}}(w\star x)](\mathbf p).
\end{aligned}
\label{eq:steerable-convolution-step-2}
\end{equation}

Hence,
\begin{equation}
    w\star(g\triangleright_{\rho_{\mathrm{in}}}x)
    =
    g\triangleright_{\rho_{\mathrm{out}}}(w\star x),
\end{equation}
which proves that the steerable convolution is equivariant for every
$g\in G$ whose action preserves the offset set $S$.

\subsection*{Appendix 2: Proof of Tri-plane Equivariance}
\label{proof-tri-equi}

\noindent
\textbf{Proposition 1: Tri-plane equivariance.}
\begin{itshape}
Let $f:\mathbb{R}^{3}\rightarrow\mathbb{R}^{|\textnormal{C}_4|}$ be a regular volumetric feature field under the planar $\textnormal{C}_4$ group and let $f_{\mathrm{xy}}$ denote its mean-pool projection onto the $XY$ plane. Let
\[
H=\{e,r^2\}\cong\textnormal{C}_2
\]
be the subgroup of $\textnormal{C}_4$ preserving each vertical plane, where $r$ denotes a $90^\circ$ rotation about the $Z$-axis. For each $\textnormal{C}_4$ regular feature $\mathbf v=(v_0,v_1,v_2,v_3)^\top$, define
\begin{equation}
\begin{aligned}
    P_{\mathrm{xz}}\mathbf v
    &=
    (v_0,v_2)^\top,\\
    P_{\mathrm{yz}}\mathbf v
    &=
    (v_1,v_3)^\top.
\end{aligned}
\label{eq: side feature decomposition}
\end{equation}
The two vertical-plane feature fields are defined as
\begin{equation}
\begin{aligned}
    f_{\mathrm{xz}}
    &=
    P_{\mathrm{xz}}
    \mathrm{Proj}_{\mathrm{xz}}(f),\\
    f_{\mathrm{yz}}
    &=
    P_{\mathrm{yz}}
    \mathrm{Proj}_{\mathrm{yz}}(f).
\end{aligned}
\label{eq: side feature definition}
\end{equation}
Let $h_{\mathrm{tp}}$ be a steerable CNN equivariant to $g\in\textnormal{C}_4$ with regular representation $\rho_{\mathrm{reg}}$, and let $h_{\mathrm{s}}$ be a steerable CNN equivariant to $H\cong\textnormal{C}_2$, shared between the $XZ$ and $YZ$ planes. We define
\begin{equation}
\begin{aligned}
    \hat f_{\mathrm{xy}}
    &=h_{\mathrm{tp}}(f_{\mathrm{xy}}),\\
    \hat f_{\mathrm{xz}}
    &=h_{\mathrm{s}}(f_{\mathrm{xz}}),\\
    \hat f_{\mathrm{yz}}
    &=h_{\mathrm{s}}(f_{\mathrm{yz}}).
\end{aligned}
\label{eq: processed tri-plane features}
\end{equation}
Let $\sigma$ and $\tau$ denote the $H$-representations carried by the input
and output features of $h_{\mathrm{s}}$, respectively, and let
\[
\rho_{\mathrm{ind}}
=
\mathrm{Ind}_{H}^{\textnormal{C}_4}\tau.
\]
Then, for every $g\in\textnormal{C}_4$ and every coordinate
$\mathbf p$, we have
\begin{equation}
\begin{aligned}
    &
    \left[
    h_{\mathrm{tp}}\cdot
    \mathrm{Proj}_{\mathrm{xy}}
    (g\triangleright f)
    \right](\mathbf p)
     =
    \rho_{\mathrm{reg}}(g)
    \hat f_{\mathrm{xy}}
    (g^{-1}\cdot\mathbf p),
\end{aligned}
\label{eq: prop1 xy}
\end{equation}
and
\begin{equation}
\begin{aligned}
&
\begin{bmatrix}
\left[
h_{\mathrm{s}}\cdot
P_{\mathrm{xz}}\cdot
\mathrm{Proj}_{\mathrm{xz}}
(g\triangleright f)
\right](\mathbf p)
\\[3pt]
\left[
h_{\mathrm{s}}\cdot
P_{\mathrm{yz}}\cdot
\mathrm{Proj}_{\mathrm{yz}}
(g\triangleright f)
\right](\mathbf p)
\end{bmatrix}
 =
\rho_{\mathrm{ind}}(g)
\begin{bmatrix}
\hat f_{\mathrm{xz}}
(g^{-1}\cdot\mathbf p)
\\[3pt]
\hat f_{\mathrm{yz}}
(g^{-1}\cdot\mathbf p)
\end{bmatrix}.
\end{aligned}
\label{eq: prop1 side}
\end{equation}
\end{itshape}

\vspace{6pt}
To prove \emph{Proposition 1}, we state four short definitions
and two lemmas, then combine them.

\noindent
\textbf{Definition 1.1: Projection.}
For $f:\mathbb{R}^3\rightarrow\mathbb{R}^{C}$, the projection to the $XY$ plane is the mean along $z$:
\begin{equation}
\begin{aligned}
    \mathrm{Proj}_{\mathrm{xy}}[f](\mathbf p)
    =
    \frac{1}{S}
    \int_{-\frac{S}{2}}^{\frac{S}{2}}
    f(\mathbf p_x,\mathbf p_y,z)\mathrm{d}z.
\end{aligned}
\end{equation}
The projections $\mathrm{Proj}_{\mathrm{xz}}$ and $\mathrm{Proj}_{\mathrm{yz}}$ are defined analogously.

\noindent
\textbf{Definition 1.2: Evaluation convention.}
A projected field $f_{\mathrm{xy}}$ is evaluated at a 3D point
$\mathbf p=(x,y,z)$ by ignoring the dropped coordinate:
\[
f_{\mathrm{xy}}(\mathbf p)=f_{\mathrm{xy}}(x,y).
\]
Similarly,
\[
f_{\mathrm{xz}}(\mathbf p)=f_{\mathrm{xz}}(x,z),
\qquad
f_{\mathrm{yz}}(\mathbf p)=f_{\mathrm{yz}}(y,z).
\]

\noindent
\textbf{Definition 1.3: Group action on 3D coordinates.}
The generator $r\in\textnormal{C}_4$ acts on coordinates by
\begin{equation}
    r\cdot(x,y,z)=(-y,x,z),
\end{equation}
and therefore
\begin{equation}
    r^{-1}\cdot(x,y,z)=(y,-x,z).
\end{equation}
The action on a regular volumetric feature field is
\begin{equation}
    [r\triangleright f](\mathbf p)
    =
    \rho_{\mathrm{reg}}(r)
    f(r^{-1}\cdot\mathbf p).
\label{eq: regular field action}
\end{equation}

\noindent
\textbf{Definition 1.4: Restriction of the regular representation.}
For the regular representation of $\textnormal{C}_4$, we adopt the channel ordering
\begin{equation}
    \rho_{\mathrm{reg}}(r)
    \begin{bmatrix}
        v_0\\v_1\\v_2\\v_3
    \end{bmatrix}
    =
    \begin{bmatrix}
        v_3\\v_0\\v_1\\v_2
    \end{bmatrix}.
\label{eq: c4 regular permutation}
\end{equation}
Let $\sigma$ denote the regular representation of $H\cong\textnormal{C}_2$, whose non-identity element $s=r^2$ acts as
\begin{equation}
    \sigma(s)
    =
    \begin{bmatrix}
        0&1\\
        1&0
    \end{bmatrix}.
\label{eq: c2 regular representation}
\end{equation}
Using the projections $P_{\mathrm{xz}}$ and $P_{\mathrm{yz}}$ defined in Eq.~\ref{eq: side feature decomposition}, we have
\begin{equation}
\begin{aligned}
    P_{\mathrm{xz}}
    \rho_{\mathrm{reg}}(r)
    &=
    \sigma(s)P_{\mathrm{yz}},\\
    P_{\mathrm{yz}}
    \rho_{\mathrm{reg}}(r)
    &=
    P_{\mathrm{xz}}.
\end{aligned}
\label{eq: c4 to c2 decomposition}
\end{equation}
Thus, the restriction of one $\textnormal{C}_4$ regular feature to $H$ decomposes into two $\textnormal{C}_2$ regular features:
\begin{equation}
    \mathrm{Res}^{\textnormal{C}_4}_{H}
    \rho_{\mathrm{reg}}^{\textnormal{C}_4}
    \cong
    \rho_{\mathrm{reg}}^{\textnormal{C}_2}
    \oplus
    \rho_{\mathrm{reg}}^{\textnormal{C}_2}.
\label{eq: restriction decomposition}
\end{equation}

\noindent
\textbf{Lemma 1.1: $XY$-plane projection equivariance.}
\emph{
For any regular volumetric feature field
$f:\mathbb{R}^3\rightarrow
\mathbb{R}^{|\textnormal{C}_4|}$
under the planar $\textnormal{C}_4$ group,
\begin{equation}
    r\triangleright
    \mathrm{Proj}_{\mathrm{xy}}[f]
    =
    \mathrm{Proj}_{\mathrm{xy}}
    [r\triangleright f].
\label{eq: lemma xy}
\end{equation}
}

\noindent
\emph{Proof:}
Let $\mathbf p=(x,y,z)$. From
Eq.~\ref{eq: regular field action},
\begin{equation}
\begin{aligned}
    &
    \mathrm{Proj}_{\mathrm{xy}}
    [r\triangleright f](\mathbf p)
    \\
    &=
    \frac{1}{S}
    \int_{-\frac{S}{2}}^{\frac{S}{2}}
    \rho_{\mathrm{reg}}(r)
    f(y,-x,z')\mathrm{d}z'
    \\
    &=
    \rho_{\mathrm{reg}}(r)
    \left(
    \frac{1}{S}
    \int_{-\frac{S}{2}}^{\frac{S}{2}}
    f(y,-x,z')\mathrm{d}z'
    \right)
    \\
    &=
    \rho_{\mathrm{reg}}(r)
    f_{\mathrm{xy}}(y,-x)
    \\
    &=
    \rho_{\mathrm{reg}}(r)
    f_{\mathrm{xy}}(r^{-1}\cdot\mathbf p)
    \\
    &=
    [r\triangleright f_{\mathrm{xy}}](\mathbf p).
\end{aligned}
\end{equation}
\hfill$\square$

\noindent
\textbf{Lemma 1.2: Vertical-plane projection equivariance.}
\emph{
Let $f_{\mathrm{xz}}$ and $f_{\mathrm{yz}}$ be the vertical-plane features defined in Eq.~\ref{eq: side feature definition}. Under the generator $r\in\textnormal{C}_4$, they transform as
\begin{equation}
\begin{aligned}
    &
    P_{\mathrm{xz}}
    \mathrm{Proj}_{\mathrm{xz}}
    [r\triangleright f](\mathbf p)
    =
    \sigma(s)
    f_{\mathrm{yz}}
    (r^{-1}\cdot\mathbf p),
    \\
    &
    P_{\mathrm{yz}}
    \mathrm{Proj}_{\mathrm{yz}}
    [r\triangleright f](\mathbf p)
    =
    f_{\mathrm{xz}}
    (r^{-1}\cdot\mathbf p).
\end{aligned}
\label{eq: lemma side}
\end{equation}
}

\noindent
\emph{Proof:}
For the $XZ$-plane projection,
\begin{equation}
\begin{aligned}
    &
    P_{\mathrm{xz}}
    \mathrm{Proj}_{\mathrm{xz}}
    [r\triangleright f](\mathbf p)
    \\
    &=
    P_{\mathrm{xz}}
    \frac{1}{S}
    \int_{-\frac{S}{2}}^{\frac{S}{2}}
    [r\triangleright f](x,y',z)\mathrm{d}y'
    \\
    &=
    P_{\mathrm{xz}}
    \rho_{\mathrm{reg}}(r)
    \frac{1}{S}
    \int_{-\frac{S}{2}}^{\frac{S}{2}}
    f(y',-x,z)\mathrm{d}y'
    \\
    &=
    P_{\mathrm{xz}}
    \rho_{\mathrm{reg}}(r)
    \mathrm{Proj}_{\mathrm{yz}}[f](-x,z).
\end{aligned}
\end{equation}
From Eq.~\ref{eq: c4 to c2 decomposition},
\begin{equation}
\begin{aligned}
    &
    P_{\mathrm{xz}}
    \mathrm{Proj}_{\mathrm{xz}}
    [r\triangleright f](\mathbf p)
    \\
    &=
    \sigma(s)
    P_{\mathrm{yz}}
    \mathrm{Proj}_{\mathrm{yz}}[f](-x,z)
    \\
    &=
    \sigma(s)
    f_{\mathrm{yz}}(-x,z)
    \\
    &=
    \sigma(s)
    f_{\mathrm{yz}}
    (r^{-1}\cdot\mathbf p).
\end{aligned}
\label{eq: xz side transform}
\end{equation}
The last equality follows from the evaluation convention in Definition 1.2. In particular, $f_{\mathrm{yz}}$ ignores the first coordinate of $r^{-1}\cdot\mathbf p=(y,-x,z)$, and therefore
\[
f_{\mathrm{yz}}
(r^{-1}\cdot\mathbf p)
=
f_{\mathrm{yz}}(-x,z).
\]

For the $YZ$-plane projection,
\begin{equation}
\begin{aligned}
    &
    P_{\mathrm{yz}}
    \mathrm{Proj}_{\mathrm{yz}}
    [r\triangleright f](\mathbf p)
    \\
    &=
    P_{\mathrm{yz}}
    \frac{1}{S}
    \int_{-\frac{S}{2}}^{\frac{S}{2}}
    [r\triangleright f](x',y,z)\mathrm{d}x'
    \\
    &=
    P_{\mathrm{yz}}
    \rho_{\mathrm{reg}}(r)
    \frac{1}{S}
    \int_{-\frac{S}{2}}^{\frac{S}{2}}
    f(y,-x',z)\mathrm{d}x'
    \\
    &=
    P_{\mathrm{yz}}
    \rho_{\mathrm{reg}}(r)
    \mathrm{Proj}_{\mathrm{xz}}[f](y,z)
    \\
    &=
    P_{\mathrm{xz}}
    \mathrm{Proj}_{\mathrm{xz}}[f](y,z)
    \\
    &=
    f_{\mathrm{xz}}(y,z)
    =
    f_{\mathrm{xz}}
    (r^{-1}\cdot\mathbf p).
\end{aligned}
\label{eq: yz side transform}
\end{equation}
Combining Eqs.~\ref{eq: xz side transform} and
\ref{eq: yz side transform} proves
Eq.~\ref{eq: lemma side}.
\hfill$\square$

\noindent
\textbf{Proof of Proposition 1:}

\emph{Lemma 1.1} shows that the projection onto the $XY$ plane is an equivariant function under the planar $\textnormal{C}_4$ group. Since $h_{\mathrm{tp}}$ is also $\textnormal{C}_4$-equivariant, $h_{\mathrm{tp}}\cdot\mathrm{Proj}_{\mathrm{xy}}$ is equivariant, which proves Eq.~\ref{eq: prop1 xy}.

We now consider the two vertical planes. Since $\textnormal{C}_4$ is generated by $r$, it is sufficient to prove equivariance to $r$. According to \emph{Lemma 1.2}, the {vertical-plane} features after rotating the volumetric field satisfy
\begin{equation}
\begin{aligned}
&
\begin{bmatrix}
P_{\mathrm{xz}}
\mathrm{Proj}_{\mathrm{xz}}
[r\triangleright f](\mathbf p)
\\[3pt]
P_{\mathrm{yz}}
\mathrm{Proj}_{\mathrm{yz}}
[r\triangleright f](\mathbf p)
\end{bmatrix}
 =
\begin{bmatrix}
\sigma(s)
f_{\mathrm{yz}}
(r^{-1}\cdot\mathbf p)
\\[3pt]
f_{\mathrm{xz}}
(r^{-1}\cdot\mathbf p)
\end{bmatrix}.
\end{aligned}
\label{eq: side before hs}
\end{equation}

Since $h_{\mathrm{s}}$ is equivariant to the $H\cong\textnormal{C}_2$ action, it satisfies
\begin{equation}
    h_{\mathrm{s}}
    \left(
    \sigma(s)
    q(-u,z)
    \right)
    =
    \tau(s)
    \left[
    h_{\mathrm{s}}(q)
    \right](-u,z).
\label{eq: hs c2 equivariance}
\end{equation}
Moreover, the same $h_{\mathrm{s}}$ is shared between the $XZ$ and $YZ$ planes. Applying $h_{\mathrm{s}}$ to both components of Eq.~\ref{eq: side before hs} therefore gives
\begin{equation}
\begin{aligned}
&
\begin{bmatrix}
\left[
h_{\mathrm{s}}\cdot
P_{\mathrm{xz}}\cdot
\mathrm{Proj}_{\mathrm{xz}}
(r\triangleright f)
\right](\mathbf p)
\\[3pt]
\left[
h_{\mathrm{s}}\cdot
P_{\mathrm{yz}}\cdot
\mathrm{Proj}_{\mathrm{yz}}
(r\triangleright f)
\right](\mathbf p)
\end{bmatrix}
\\
&\qquad =
\begin{bmatrix}
\tau(s)
\hat f_{\mathrm{yz}}
(r^{-1}\cdot\mathbf p)
\\[3pt]
\hat f_{\mathrm{xz}}
(r^{-1}\cdot\mathbf p)
\end{bmatrix}.
\end{aligned}
\label{eq: side after hs}
\end{equation}

Using the coset representatives $\{H,rH\}$ of $\textnormal{C}_4/H$, the induced representation of the generator $r$ is
\begin{equation}
    \rho_{\mathrm{ind}}(r)
    =
    \begin{bmatrix}
        0 & \tau(s)\\
        I & 0
    \end{bmatrix}.
\label{eq: induced generator}
\end{equation}
Hence,
\begin{equation}
\begin{aligned}
&
\begin{bmatrix}
\tau(s)
\hat f_{\mathrm{yz}}
(r^{-1}\cdot\mathbf p)
\\[3pt]
\hat f_{\mathrm{xz}}
(r^{-1}\cdot\mathbf p)
\end{bmatrix}
\\
&\qquad =
\rho_{\mathrm{ind}}(r)
\begin{bmatrix}
\hat f_{\mathrm{xz}}
(r^{-1}\cdot\mathbf p)
\\[3pt]
\hat f_{\mathrm{yz}}
(r^{-1}\cdot\mathbf p)
\end{bmatrix}.
\end{aligned}
\label{eq: induced side action}
\end{equation}

In particular,
\begin{equation}
\begin{aligned}
    \rho_{\mathrm{ind}}(r^2)
    &=
    \begin{bmatrix}
        \tau(s)&0\\
        0&\tau(s)
    \end{bmatrix},
    \\
    \rho_{\mathrm{ind}}(r^3)
    &=
    \begin{bmatrix}
        0&I\\
        \tau(s)&0
    \end{bmatrix},
    \\
    \rho_{\mathrm{ind}}(r^4)
    &=I.
\end{aligned}
\label{eq: induced powers}
\end{equation}
Therefore, $\rho_{\mathrm{ind}}$ defines a representation of
$\textnormal{C}_4$. Since every
$g\in\textnormal{C}_4$ can be written as $g=r^k$,
Eq.~\ref{eq: prop1 side} holds for every
$g\in\textnormal{C}_4$.

\hfill Q.E.D.

\noindent \textbf{Remark: Practical implementation and expressiveness.} The equivariance proof is stated using explicit projections and representation-theoretic decompositions, but an implementation need not discard any feature dimension in order to realize this construction. In particular, the $XZ$ and $YZ$ branches may retain all available spatial and channel dimensions, while $P_{\mathrm{xz}}$ and $P_{\mathrm{yz}}$ can be implemented as index selections, reshaping operations, or learned equivariant linear maps that preserve the corresponding $H$-representation. Likewise, the shared network $h_{\mathrm{s}}$ can operate on the complete feature tensors of both planes, with the induced $\textnormal{C}_4$ action implemented by a permutation of plane indices together with the within-plane $H$ action. Thus, the theoretical decomposition does not require dropping arbitrary dimensions in practical engineering implementations, allowing the model to preserve its full expressive capacity while maintaining the stated equivariance.

\subsection*{Appendix 3: $\textnormal{SO}(3)$ Rotation Described with $\textnormal{SO}(2)$ Representations}

\noindent \textbf{Proposition 2.} \textit{Let $R\in\textnormal{SO}(3)$ be a gripper orientation. Its row-based six-dimensional encoding consists of three copies of the type-1 representation of $\textnormal{SO}(2)$ restricted to $\textnormal{C}_4$:}
\begin{equation}
g\cdot\mathbf{r}=(\bigoplus_{i=1}^{3}\rho_1(g))\mathbf{r},\qquad g\in\textnormal{C}_4.
\end{equation}

\emph{Proof:} Let $T_g=\left(\begin{smallmatrix}\rho_1(g)&0\\0&1\end{smallmatrix}\right)$ denote the $\textnormal{SO}(2)$ rotation about the world $Z$-axis. Write $R=[R_1,R_2,R_3]$, where $R_i$ is its $i$-th column, and let $a_i=(R_{1i},R_{2i})^\top\in\mathbb{R}^2$ be the first two entries of $R_i$. We define
\begin{equation}
\mathbf{r}=\left(a_1^\top,a_2^\top,a_3^\top\right)^\top\in\mathbb{R}^6.
\end{equation}
The first two entries of the $i$-th column of $T_gR$ are $\rho_1(g)a_i$. Therefore,
\begin{equation}
\mathbf{r}(T_gR)=(\rho_1(g)\oplus\rho_1(g)\oplus\rho_1(g))\mathbf{r}(R),
\end{equation}
which proves the claim. This row-based encoding differs from the commonly used six-dimensional rotation representation, which retains the first two columns of $R$ (equivalently, drops the last column). For $R\in\textnormal{SO}(3)$, the retained first two rows uniquely determine the third row through their cross product.

\subsection*{Appendix 4: Equivariance of Vertical-to-Tableplane Feature Augmentation}
\label{proof-v2tp}

\noindent \textbf{Proposition 3: V2TP equivariance.}
\begin{itshape}
At every V2TP connection, suppose that $f^l_{\mathrm{xz}}$ and $f^l_{\mathrm{yz}}$ are paired $\textnormal{C}_2$ regular feature fields whose joint $\textnormal{C}_4$ action is the induced action established in Proposition 1. Let $f^l_{\mathrm{xy}}$ be a direct sum of $\textnormal{C}_4$ regular fields. Then the V2TP update in Eqs.~\ref{eq:v2tp_induced} and \ref{eq:v2tp_fusion} is $\textnormal{C}_4$-equivariant.
\end{itshape}

\noindent \emph{Proof:}
It suffices to verify the action of the generator $r\in\textnormal{C}_4$. Consider one pair of $\textnormal{C}_2$ regular fields and write
\begin{align}
& a(x)=\operatorname{MeanPool}_{z}\left(f^l_{\mathrm{xz}}(x,z)\right) \\
& b(y)=\operatorname{MeanPool}_{z}\left(f^l_{\mathrm{yz}}(y,z)\right).
\end{align}
Let $*'$ denote the corresponding feature fields after applying the generator $r$. By Proposition 1, the vertical fields transform as
\[
f^{l\prime}_{\mathrm{xz}}(x,z)=\tau(s)f^l_{\mathrm{yz}}(-x,z),
\qquad
f^{l\prime}_{\mathrm{yz}}(y,z)=f^l_{\mathrm{xz}}(y,z),
\]
where $s=r^2$ and $\tau(s)$ swaps the $e$ and $s$ components of a $\textnormal{C}_2$ regular field. Since the $\textnormal{C}_4$ action leaves the $z$ coordinate unchanged, mean pooling commutes with this action. Therefore,
\[
a'(x)=\tau(s)b(-x),
\qquad
b'(y)=a(y).
\]

Define the V2TP-reconstructed field for this pair by
\[
B(x,y)=[a_e(x),b_e(y),a_s(x),b_s(y)]^\top.
\]
Using the preceding identities gives
\[
B'(x,y)=[b_s(-x),a_e(y),b_e(-x),a_s(y)]^\top.
\]
On the other hand, $r^{-1}\cdot(x,y)=(y,-x)$, and the regular representation in Eq.~\ref{eq: c4 regular permutation} yields
\begin{equation}
\begin{aligned}
& \rho_{\mathrm{reg}}(r)B(r^{-1}\cdot(x,y)) \\
& =\rho_{\mathrm{reg}}(r)
[a_e(y),b_e(-x),a_s(y),b_s(-x)]^\top
=B'(x,y).
\end{aligned}
\end{equation}
Thus, mean pooling, broadcasting, and the channel ordering in Eq.~\ref{eq:v2tp_induced} map every paired $\textnormal{C}_2$ regular field to a $\textnormal{C}_4$ regular field. Applying this construction independently to all field pairs gives a direct sum of $\textnormal{C}_4$ regular fields.

Finally, both $f^l_{\mathrm{xy}}$ and $\bar f^l_{\mathrm{s}\rightarrow\mathrm{xy}}$ transform under direct sums of $\rho_{\mathrm{reg}}$. Their channel-wise concatenation consequently has the same $\textnormal{C}_4$ representation. Because $F_l$ is a point-wise $\textnormal{C}_4$-equivariant convolution,
\[
F_l\left([r\triangleright f^l_{\mathrm{xy}}]\,\Vert\,[r\triangleright\bar f^l_{\mathrm{s}\rightarrow\mathrm{xy}}]\right)
=r\triangleright F_l\left(f^l_{\mathrm{xy}}\,\Vert\,\bar f^l_{\mathrm{s}\rightarrow\mathrm{xy}}\right).
\]
Hence $f^{l+1}_{\mathrm{xy}}$ is $\textnormal{C}_4$-equivariant under $r$. Since $r$ generates $\textnormal{C}_4$, the result holds for every group element.
\hfill$\square$

\subsection*{Appendix 5: Proof of Equivariance for Equivariant Deformable Attention}
The proposed EDA is formulated as:
\begin{align}
    & \tilde{\mathbf{c}}(\mathbf{p}) = h_{\text{out}}\left(\sum_{k=1}^{K} A_k \cdot h_{\text{in}} ( \mathbf{c}(\mathbf{p}+\Delta_k))\right)+\mathbf{c}(\mathbf{p}), \\
    & A_k(\mathbf{p}) = h^k_{\text{attn}}(\mathbf{c}(\mathbf{p})), \quad \Delta_k(\mathbf{p}) = h^k_{\text{offset}}(\mathbf{c}(\mathbf{p})),
\end{align}
where $K$ is the number of sampling offsets, and $h_{\text{in}}$, $h_{\text{out}}$, $h^k_{\text{attn}}$, $h^k_{\text{offset}}$ are equivariant linear layers. $\Delta_k$ is the $k$-th sampling offset, which is represented as a concatenation of type-1 and type-0 features, as $\Delta_k(g\cdot\mathbf{p}) = (\rho_1(g) \oplus \rho_0(g)) \Delta_k(\mathbf{p})$. $A_k$ is the attention weight for the $k$-th offset $\Delta_k$, which is represented as a type-0 feature, as $A_k(g\cdot\mathbf{p}) = \rho_0(g)A_k(\mathbf{p}) = A_k(\mathbf{p})$.
Thus, we have:
\begin{equation}
\begin{aligned}
    & \tilde{\mathbf{c}}(g\cdot\mathbf{p}) = h_{\text{out}}(\sum_{k=1}^{K} A_k \cdot h_{\text{in}} ( \mathbf{c}(g\cdot\mathbf{p}+g\cdot\Delta_k)))+\mathbf{c}(g\cdot\mathbf{p}) \\
    & =  h_{\text{out}}(\sum_{k=1}^{K} A_k  \cdot \rho_{\text{in}}(g) h_{\text{in}} ( \mathbf{c}(\mathbf{p}+\Delta_k)))+\rho_{\text{out}}(g) \mathbf{c}(\mathbf{p})\\
    & = \rho_{\text{out}}(g) h_{\text{out}}(\sum_{k=1}^{K} A_k   h_{\text{in}} ( \mathbf{c}(\mathbf{p}+\Delta_k)))+\rho_{\text{out}}(g) \mathbf{c}(\mathbf{p})\\
    & = \rho_{\text{out}}(g)\tilde{\mathbf{c}}(\mathbf{p}),
\end{aligned}
\end{equation}
where $\rho_\textnormal{in}$ and $\rho_\textnormal{out}$ are the representations of the outputs of $h_\textnormal{in}(\cdot)$ and $h_\textnormal{out}(\cdot)$.

\begin{figure*} [!t]
\centering
    \resizebox{0.9\linewidth}{!}{
    \subfloat[\label{fig:packed_sr}]{
    \includegraphics[scale=0.4]{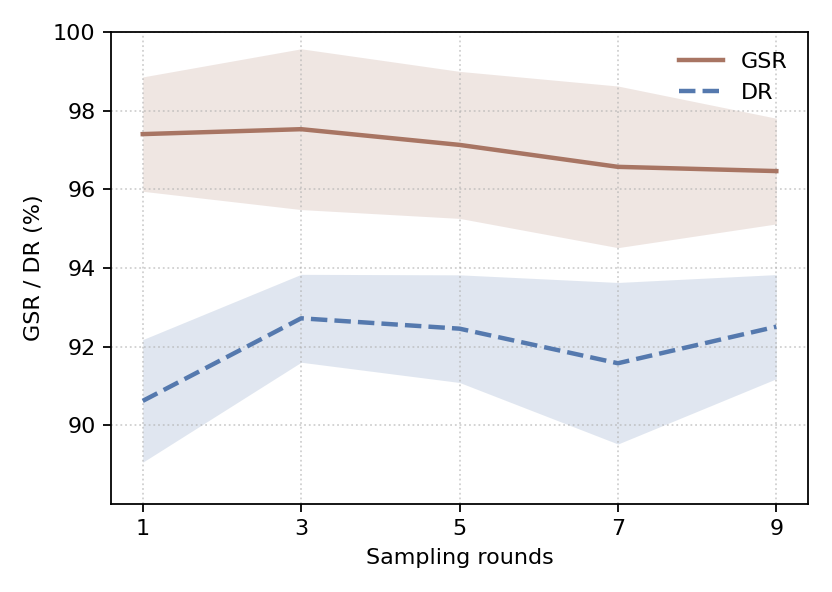}}
    \subfloat[\label{fig:pile_sr}]{
    \includegraphics[scale=0.4]{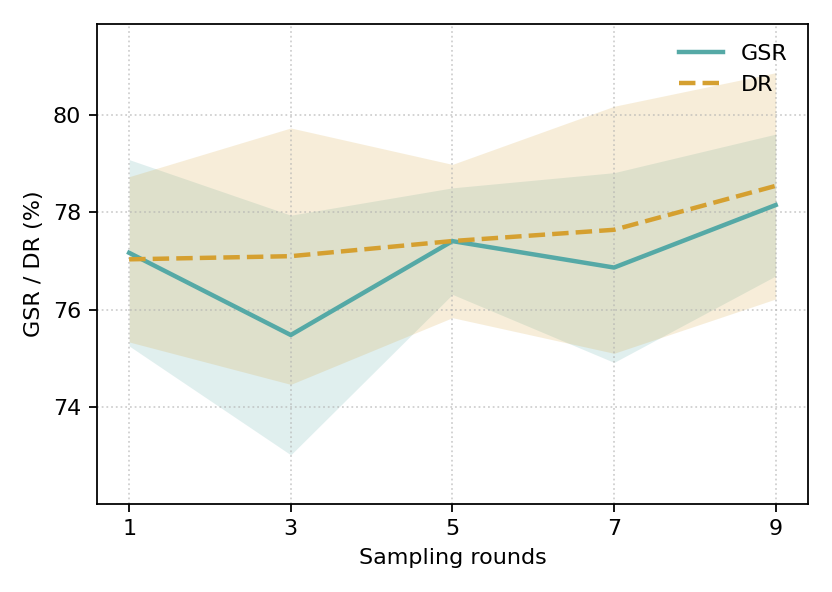}}
    }
    \caption{Ablation study of the number of sampling rounds in EquiIGD. (a) Packed scene. (b) Pile scene. Solid lines denote GSR, and dashed lines denote DR.}
\label{fig:sample_rounds}
\end{figure*}

\subsection*{Appendix 6: Proof of Equivariance for the Equivariant Grasp-conditioned Deformable Attention Module}

We begin by formulating the proposed EquiGraspDAM as follows:
\begin{align}
    \mathbf{Q}(\mathbf{p}) &= h_{\text{Q}}(\mathbf{c}(\mathbf{p})), \\
    \mathbf{K}(\mathbf{p}, \mathbf{r}) &= h_{\text{K}}\Bigl(\bigl\{\mathbf{c}(T_{\mathbf{g}} \mathbf{u}_l)\bigr\}_{l=1}^L\Bigr), \\
    \mathbf{V}(\mathbf{p}, \mathbf{r}) &= h_{\text{V}}\Bigl(\bigl\{\mathbf{c}(T_{\mathbf{g}} \mathbf{u}_l)\bigr\}_{l=1}^L\Bigr), \\
    \bar{\mathbf{c}}(\mathbf{p}, \mathbf{r}) &= h_{\text{out}}\Biggl(\mathbf{V}(\mathbf{p}, \mathbf{r})\,\text{Softmax}\Bigl(\frac{\mathbf{K}(\mathbf{p}, \mathbf{r})^T\mathbf{Q}(\mathbf{p})}{\sqrt{d_k}}\Bigr)\Biggr) + \mathbf{c}(\mathbf{p}),
\end{align}
where $\mathbf{p}$ denotes the spatial position and $\mathbf{r}$ represents grasp orientation. Note that the query $\mathbf{Q}(\mathbf{p})$ is equivariant by virtue of the equivariant properties of both $h_{\text{Q}}(\cdot)$ and $\mathbf{c}(\cdot)$.

Let $\mathbf{g} = (\mathbf{p},\mathbf{r})$, and consider a group action such that $g\cdot \mathbf{g} = (g\cdot\mathbf{p}, g\cdot\mathbf{r})$. By the assumed property of the transformation operator, we have:
\[
T_{g\cdot \mathbf{g}} = g\cdot T_{\mathbf{g}}.
\]
Thus, the key representation can be derived as:
\begin{equation}
    \begin{aligned}
        \mathbf{K}(g\cdot\mathbf{p}, g\cdot\mathbf{r})
        &= \left\{h_{\text{K}}\left(\mathbf{c}(g\cdot T_{\mathbf{g}} \mathbf{u}_l)\right)\right\}_{l=1}^L \\
        &= \left\{h_{\text{K}}\left(\rho(g)\,\mathbf{c}(T_{\mathbf{g}} \mathbf{u}_l)\right)\right\}_{l=1}^L \\
        &= \rho_{\text{hid}}(g)\left\{h_{\text{K}}\left(\mathbf{c}(T_{\mathbf{g}} \mathbf{u}_l)\right)\right\}_{l=1}^L \\
        &= \rho_{\text{hid}}(g)\,\mathbf{K}(\mathbf{p}, \mathbf{r}),
    \end{aligned}
\end{equation}
where $\rho(g)$ and $\rho_{\text{hid}}(g)$ denote the representations of the group element $g$ in the input and hidden feature spaces, respectively. A similar derivation holds for $\mathbf{V}(\mathbf{p}, \mathbf{r})$.

We now establish the equivariance of the output feature $\bar{\mathbf{c}}(\mathbf{p}, \mathbf{r})$. By definition, we have:
\begin{equation}
    \scriptsize
    \begin{aligned}
       &  \bar{\mathbf{c}}(g\cdot\mathbf{p}, g\cdot\mathbf{r})\\
        &= h_{\text{out}}\Biggl(\mathbf{V}(g\cdot\mathbf{p}, g\cdot\mathbf{r})\,\text{Softmax}\Bigl(\frac{\mathbf{K}(g\cdot\mathbf{p}, g\cdot\mathbf{r})^T\mathbf{Q}(g\cdot\mathbf{p})}{\sqrt{d_k}}\Bigr)\Biggr) + \mathbf{c}(g\cdot\mathbf{p}) \\
        &= h_{\text{out}}\Biggl(\rho_{\text{hid}}(g)\,\mathbf{V}(\mathbf{p}, \mathbf{r})\,\text{Softmax}\Bigl(\frac{(\rho_{\text{hid}}(g)\,\mathbf{K}(\mathbf{p}, \mathbf{r}))^T \, \rho_{\text{hid}}(g)\,\mathbf{Q}(\mathbf{p})}{\sqrt{d_k}}\Bigr)\Biggr) \\
        & \quad \quad \quad + \rho(g)\,\mathbf{c}(\mathbf{p}) \\
        &= h_{\text{out}}\Biggl(\rho_{\text{hid}}(g)\,\mathbf{V}(\mathbf{p}, \mathbf{r})\,\text{Softmax}\Bigl(\frac{\mathbf{K}(\mathbf{p}, \mathbf{r})^T \mathbf{Q}(\mathbf{p})}{\sqrt{d_k}}\Bigr)\Biggr) + \rho(g)\,\mathbf{c}(\mathbf{p}) \\
        &= \rho(g)\,\Biggl(h_{\text{out}}\Bigl(\mathbf{V}(\mathbf{p}, \mathbf{r})\,\text{Softmax}\Bigl(\frac{\mathbf{K}(\mathbf{p}, \mathbf{r})^T \mathbf{Q}(\mathbf{p})}{\sqrt{d_k}}\Bigr)\Bigr) + \mathbf{c}(\mathbf{p})\Biggr) \\
        &= \rho(g)\,\bar{\mathbf{c}}(\mathbf{p}, \mathbf{r}),
    \end{aligned}
\end{equation}
where $\rho_\textnormal{hid}(g)^T\rho_\textnormal{hid}(g)=I$, and the output of $h_{\text{out}}(\cdot)$ shares the same representation as $\mathbf{c}(\mathbf{p})$.
Thus, we have shown that the Equivariant Grasp-conditioned Deformable Attention Module is equivariant with respect to the considered group action.

\begin{table}[!t]
  \centering
  \caption{Ablation of EquiIGD's grasp-orientation model and sampling method. FM denotes flow matching; the \emph{Graspness}, \emph{EDA}, and \emph{GC} columns indicate the use of graspness prediction, Equivariant Deformable Attention, and the Grasp Classifier, respectively. We report mean $\pm$ standard deviation for GSR and DR.}
  \setlength{\tabcolsep}{3mm}{}
  \resizebox{\linewidth}{!}{
  \begin{tabular}{cccc|cc}
    \toprule
     Orientation generation method & Graspness  & EDA & GC  &  GSR (\%) & DR (\%) \\
    \midrule
     FM (10-step) & \Checkmark &   &  & 91.1$\pm$2.5 & 89.1$\pm$2.1 \\
     FM (10-step) & \Checkmark &  \Checkmark & & 93.5$\pm$1.0 & 91.1$\pm$0.5 \\
     FM (10-step) & \Checkmark &  \Checkmark & \Checkmark  & 97.3$\pm$1.1 & 92.1$\pm$1.2 \\
     DDPM (10-step) & \Checkmark &  \Checkmark & \Checkmark  & 92.6$\pm$0.9 & 88.1$\pm$1.1 \\
     DDPM (100-step) & \Checkmark &  \Checkmark & \Checkmark  & 92.6$\pm$1.4 & 85.0$\pm$2.2 \\
    \bottomrule
  \end{tabular}
  }
  \label{tab:ablation igd}
\end{table}

\subsection*{Appendix 7: Ablation Study for EquiIGD}
To validate the components proposed in Sec.~\ref{sec:equiigd}, we conduct the EquiIGD ablation in Tab.~\ref{tab:ablation igd}. Starting from the FM baseline, adding EDA improves GSR from 91.1\% to 93.5\% and DR from 89.1\% to 91.1\%. Adding the grasp classifier (GC), which is supervised with grasp orientations rather than only graspness labels, further improves GSR to 97.3\% and DR to 92.1\%. Under the same 10-step budget, FM outperforms DDPM in both GSR (97.3\% versus 92.6\%) and DR (92.1\% versus 88.1\%). Increasing DDPM sampling from 10 to 100 steps does not improve GSR (both 92.6\%) and reduces DR from 88.1\% to 85.0\%.

\subsection*{Appendix 8: Ablation Study of Sampling Rounds in EquiIGD}
As in IGD \citep{igd}, EquiIGD can perform multiple sampling rounds to generate diverse grasp orientations at the same position. We evaluate the influence of this hyperparameter by varying the number of sampling rounds from 1 to 9 (Fig.~\ref{fig:sample_rounds}). Overall, increasing the number of rounds does not yield a consistent improvement. In packed scenes, both GSR and DR remain relatively stable, with GSR showing a slight decrease as more rounds are used. In pile scenes, the metrics fluctuate across sampling rounds, with only a modest improvement at nine rounds. These variations largely fall within the corresponding uncertainty intervals. The results suggest that EquiIGD already produces sufficiently effective rotation candidates with a small number of sampling rounds, while additional samples often receive similar scores from the grasp classifier and therefore provide diminishing returns. Thus, using more sampling rounds increases computational cost without reliably improving grasp performance.

\bibliographystyle{SageH}

\bibliography{ref}

@article{giga,
  title={Synergies between affordance and geometry: 6-dof grasp detection via implicit representations},
  author={Jiang, Zhenyu and Zhu, Yifeng and Svetlik, Maxwell and Fang, Kuan and Zhu, Yuke},
  journal={arXiv preprint arXiv:2104.01542},
  year={2021}
}

@inproceedings{se3dif,
  title={Se (3)-diffusionfields: Learning smooth cost functions for joint grasp and motion optimization through diffusion},
  author={Urain, Julen and Funk, Niklas and Peters, Jan and Chalvatzaki, Georgia},
  booktitle={2023 IEEE International Conference on Robotics and Automation (ICRA)},
  pages={5923--5930},
  year={2023},
  organization={IEEE}
}

@article{diffusion-edfs,
  title={Diffusion-edfs: Bi-equivariant denoising generative modeling on se (3) for visual robotic manipulation},
  author={Ryu, Hyunwoo and Kim, Jiwoo and Chang, Junwoo and Ahn, Hyun Seok and Seo, Joohwan and Kim, Taehan and Choi, Jongeun and Horowitz, Roberto},
  journal={arXiv preprint arXiv:2309.02685},
  year={2023}
}

@inproceedings{vgn,
  title={Volumetric grasping network: Real-time 6 dof grasp detection in clutter},
  author={Breyer, Michel and Chung, Jen Jen and Ott, Lionel and Siegwart, Roland and Nieto, Juan},
  booktitle={Conference on Robot Learning},
  pages={1602--1611},
  year={2021},
  organization={PMLR}
}

@inproceedings{6dofgraspnet,
  title={6-dof graspnet: Variational grasp generation for object manipulation},
  author={Mousavian, Arsalan and Eppner, Clemens and Fox, Dieter},
  booktitle={Proceedings of the IEEE/CVF International Conference on Computer Vision},
  pages={2901--2910},
  year={2019}
}

@inproceedings{s4g,
  title={S4g: Amodal single-view single-shot se (3) grasp detection in cluttered scenes},
  author={Qin, Yuzhe and Chen, Rui and Zhu, Hao and Song, Meng and Xu, Jing and Su, Hao},
  booktitle={Conference on robot learning},
  pages={53--65},
  year={2020},
  organization={PMLR}
}

@inproceedings{calli2015ycb,
  title={The ycb object and model set: Towards common benchmarks for manipulation research},
  author={Calli, Berk and Singh, Arjun and Walsman, Aaron and Srinivasa, Siddhartha and Abbeel, Pieter and Dollar, Aaron M},
  booktitle={2015 international conference on advanced robotics (ICAR)},
  pages={510--517},
  year={2015},
  organization={IEEE}
}

@inproceedings{kappler2015leveraging,
  title={Leveraging big data for grasp planning},
  author={Kappler, Daniel and Bohg, Jeannette and Schaal, Stefan},
  booktitle={2015 IEEE international conference on robotics and automation (ICRA)},
  pages={4304--4311},
  year={2015},
  organization={IEEE}
}

@article{kasper2012kit,
  title={The kit object models database: An object model database for object recognition, localization and manipulation in service robotics},
  author={Kasper, Alexander and Xue, Zhixing and Dillmann, R{\"u}diger},
  journal={The International Journal of Robotics Research},
  volume={31},
  number={8},
  pages={927--934},
  year={2012},
  publisher={SAGE Publications Sage UK: London, England}
}

@inproceedings{singh2014bigbird,
  title={Bigbird: A large-scale 3d database of object instances},
  author={Singh, Arjun and Sha, James and Narayan, Karthik S and Achim, Tudor and Abbeel, Pieter},
  booktitle={2014 IEEE international conference on robotics and automation (ICRA)},
  pages={509--516},
  year={2014},
  organization={IEEE}
}

@article{gpd,
  title={Grasp pose detection in point clouds},
  author={Ten Pas, Andreas and Gualtieri, Marcus and Saenko, Kate and Platt, Robert},
  journal={The International Journal of Robotics Research},
  volume={36},
  number={13-14},
  pages={1455--1473},
  year={2017},
  publisher={SAGE Publications Sage UK: London, England}
}

@inproceedings{peng2020convolutional,
  title={Convolutional occupancy networks},
  author={Peng, Songyou and Niemeyer, Michael and Mescheder, Lars and Pollefeys, Marc and Geiger, Andreas},
  booktitle={Computer Vision--ECCV 2020: 16th European Conference, Glasgow, UK, August 23--28, 2020, Proceedings, Part III 16},
  pages={523--540},
  year={2020},
  organization={Springer}
}

@inproceedings{fang2020graspnet,
  title={Graspnet-1billion: A large-scale benchmark for general object grasping},
  author={Fang, Hao-Shu and Wang, Chenxi and Gou, Minghao and Lu, Cewu},
  booktitle={Proceedings of the IEEE/CVF conference on computer vision and pattern recognition},
  pages={11444--11453},
  year={2020}
}

@inproceedings{wang2021graspness,
  title={Graspness discovery in clutters for fast and accurate grasp detection},
  author={Wang, Chenxi and Fang, Hao-Shu and Gou, Minghao and Fang, Hongjie and Gao, Jin and Lu, Cewu},
  booktitle={Proceedings of the IEEE/CVF International Conference on Computer Vision},
  pages={15964--15973},
  year={2021}
}

@inproceedings{cesa2022program,
  title={A program to build E (N)-equivariant steerable CNNs},
  author={Cesa, Gabriele and Lang, Leon and Weiler, Maurice},
  booktitle={International conference on learning representations},
  year={2022}
}

@inproceedings{igd,
  title={Implicit Grasp Diffusion: Bridging the Gap between Dense Prediction and Sampling-based Grasping},
  author={Song, Pinhao and Li, Pengteng and Detry, Renaud},
  booktitle={8th Annual Conference on Robot Learning},
  year={2024}
}

@article{freiberg2025diffusion,
  title={Diffusion for Multi-Embodiment Grasping},
  author={Freiberg, Roman and Qualmann, Alexander and Vien, Ngo Anh and Neumann, Gerhard},
  journal={IEEE Robotics and Automation Letters},
  year={2025},
  publisher={IEEE}
}

@article{hu2024orbitgrasp,
  title={OrbitGrasp: $ SE (3) $-Equivariant Grasp Learning},
  author={Hu, Boce and Zhu, Xupeng and Wang, Dian and Dong, Zihao and Huang, Haojie and Wang, Chenghao and Walters, Robin and Platt, Robert},
  journal={arXiv preprint arXiv:2407.03531},
  year={2024}
}

@article{edf,
  title={Equivariant descriptor fields: Se (3)-equivariant energy-based models for end-to-end visual robotic manipulation learning},
  author={Ryu, Hyunwoo and Lee, Hong-in and Lee, Jeong-Hoon and Choi, Jongeun},
  journal={arXiv preprint arXiv:2206.08321},
  year={2022}
}

@inproceedings{huang2023edge,
  title={Edge grasp network: A graph-based se (3)-invariant approach to grasp detection},
  author={Huang, Haojie and Wang, Dian and Zhu, Xupeng and Walters, Robin and Platt, Robert},
  booktitle={2023 IEEE International Conference on Robotics and Automation (ICRA)},
  pages={3882--3888},
  year={2023},
  organization={IEEE}
}

@article{neugraspnet,
  title={Learning any-view 6dof robotic grasping in cluttered scenes via neural surface rendering},
  author={Jauhri, Snehal and Lunawat, Ishikaa and Chalvatzaki, Georgia},
  journal={arXiv preprint arXiv:2306.07392},
  year={2023}
}

@inproceedings{qi2017pointnet,
  title={Pointnet: Deep learning on point sets for 3d classification and segmentation},
  author={Qi, Charles R and Su, Hao and Mo, Kaichun and Guibas, Leonidas J},
  booktitle={Proceedings of the IEEE conference on computer vision and pattern recognition},
  pages={652--660},
  year={2017}
}

@inproceedings{vn,
  title={Vector neurons: A general framework for so (3)-equivariant networks},
  author={Deng, Congyue and Litany, Or and Duan, Yueqi and Poulenard, Adrien and Tagliasacchi, Andrea and Guibas, Leonidas J},
  booktitle={Proceedings of the IEEE/CVF International Conference on Computer Vision},
  pages={12200--12209},
  year={2021}
}

@inproceedings{minkunet,
  title={4d spatio-temporal convnets: Minkowski convolutional neural networks},
  author={Choy, Christopher and Gwak, JunYoung and Savarese, Silvio},
  booktitle={Proceedings of the IEEE/CVF conference on computer vision and pattern recognition},
  pages={3075--3084},
  year={2019}
}

@inproceedings{zhang2024dexgraspnet,
  title={DexGraspNet 2.0: Learning Generative Dexterous Grasping in Large-scale Synthetic Cluttered Scenes},
  author={Zhang, Jialiang and Liu, Haoran and Li, Danshi and Yu, XinQiang and Geng, Haoran and Ding, Yufei and Chen, Jiayi and Wang, He},
  booktitle={8th Annual Conference on Robot Learning},
  year={2024}
}

@inproceedings{wu2024point,
  title={Point transformer v3: Simpler faster stronger},
  author={Wu, Xiaoyang and Jiang, Li and Wang, Peng-Shuai and Liu, Zhijian and Liu, Xihui and Qiao, Yu and Ouyang, Wanli and He, Tong and Zhao, Hengshuang},
  booktitle={Proceedings of the IEEE/CVF Conference on Computer Vision and Pattern Recognition},
  pages={4840--4851},
  year={2024}
}

@inproceedings{zurbrugg2024icgnet,
  title={ICGNet: a unified approach for instance-centric grasping},
  author={Zurbr{\"u}gg, Ren{\'e} and Liu, Yifan and Engelmann, Francis and Kumar, Suryansh and Hutter, Marco and Patil, Vaishakh and Yu, Fisher},
  booktitle={2024 IEEE International Conference on Robotics and Automation (ICRA)},
  pages={4140--4146},
  year={2024},
  organization={IEEE}
}

@article{liu2019point,
  title={Point-voxel cnn for efficient 3d deep learning},
  author={Liu, Zhijian and Tang, Haotian and Lin, Yujun and Han, Song},
  journal={Advances in neural information processing systems},
  volume={32},
  year={2019}
}

@article{cohen2016steerable,
  title={Steerable cnns},
  author={Cohen, Taco S and Welling, Max},
  journal={arXiv preprint arXiv:1612.08498},
  year={2016}
}

@article{zhu2023robot,
  title={On robot grasp learning using equivariant models},
  author={Zhu, Xupeng and Wang, Dian and Su, Guanang and Biza, Ondrej and Walters, Robin and Platt, Robert},
  journal={Autonomous Robots},
  volume={47},
  number={8},
  pages={1175--1193},
  year={2023},
  publisher={Springer}
}

@inproceedings{wang2022equivariant,
  title={Equivariant $ q $ learning in spatial action spaces},
  author={Wang, Dian and Walters, Robin and Zhu, Xupeng and Platt, Robert},
  booktitle={Conference on Robot Learning},
  pages={1713--1723},
  year={2022},
  organization={PMLR}
}

@inproceedings{lim2024equigraspflow,
  title={Equigraspflow: Se (3)-equivariant 6-dof grasp pose generative flows},
  author={Lim, Byeongdo and Kim, Jongmin and Kim, Jihwan and Lee, Yonghyeon and Park, Frank C},
  booktitle={8th Annual Conference on Robot Learning},
  year={2024}
}

@article{fuchs2020se,
  title={Se (3)-transformers: 3d roto-translation equivariant attention networks},
  author={Fuchs, Fabian and Worrall, Daniel and Fischer, Volker and Welling, Max},
  journal={Advances in neural information processing systems},
  volume={33},
  pages={1970--1981},
  year={2020}
}

@article{liao2022equiformer,
  title={Equiformer: Equivariant graph attention transformer for 3d atomistic graphs},
  author={Liao, Yi-Lun and Smidt, Tess},
  journal={arXiv preprint arXiv:2206.11990},
  year={2022}
}

@article{liao2023equiformerv2,
  title={Equiformerv2: Improved equivariant transformer for scaling to higher-degree representations},
  author={Liao, Yi-Lun and Wood, Brandon and Das, Abhishek and Smidt, Tess},
  journal={arXiv preprint arXiv:2306.12059},
  year={2023}
}

@article{liu2022flow,
  title={Flow straight and fast: Learning to generate and transfer data with rectified flow},
  author={Liu, Xingchao and Gong, Chengyue and Liu, Qiang},
  journal={arXiv preprint arXiv:2209.03003},
  year={2022}
}

@article{weiler2024equivariant,
  title={Equivariant and coordinate independent convolutional networks: A gauge field theory of neural networks},
  author={Weiler, Maurice and others},
  year={2024}
}

@article{ko2025simultaneous,
  title={Simultaneous Pick and Place Detection by Combining SE (3) Diffusion Models with Differential Kinematics},
  author={Ko, Tianyi and Ikeda, Takuya and Nishiwaki, Koichi},
  journal={arXiv preprint arXiv:2504.19502},
  year={2025}
}

@inproceedings{liang2019pointnetgpd,
  title={Pointnetgpd: Detecting grasp configurations from point sets},
  author={Liang, Hongzhuo and Ma, Xiaojian and Li, Shuang and G{\"o}rner, Michael and Tang, Song and Fang, Bin and Sun, Fuchun and Zhang, Jianwei},
  booktitle={2019 International Conference on Robotics and Automation (ICRA)},
  pages={3629--3635},
  year={2019},
  organization={IEEE}
}

@article{mahler2019learning,
  title={Learning ambidextrous robot grasping policies},
  author={Mahler, Jeffrey and Matl, Matthew and Satish, Vishal and Danielczuk, Michael and DeRose, Bill and McKinley, Stephen and Goldberg, Ken},
  journal={Science Robotics},
  volume={4},
  number={26},
  pages={eaau4984},
  year={2019},
  publisher={American Association for the Advancement of Science}
}

@inproceedings{liu2024efficient,
  title={Efficient end-to-end detection of 6-dof grasps for robotic bin picking},
  author={Liu, Yushi and Qualmann, Alexander and Yu, Zehao and Gabriel, Miroslav and Schillinger, Philipp and Spies, Markus and Vien, Ngo Anh and Geiger, Andreas},
  booktitle={2024 IEEE International Conference on Robotics and Automation (ICRA)},
  pages={5427--5433},
  year={2024},
  organization={IEEE}
}

@inproceedings{bekkers2018roto,
  title={Roto-translation covariant convolutional networks for medical image analysis},
  author={Bekkers, Erik J and Lafarge, Maxime W and Veta, Mitko and Eppenhof, Koen AJ and Pluim, Josien PW and Duits, Remco},
  booktitle={International conference on medical image computing and computer-assisted intervention},
  pages={440--448},
  year={2018},
  organization={Springer}
}

\end{document}